%% file: computing_with_spikes.tex
\documentclass[11pt]{article}
\usepackage{amsmath}
\usepackage{amssymb}

\usepackage{times}
\usepackage{graphicx}
\usepackage{color}
\usepackage{multirow}
\usepackage[authoryear]{natbib}
\usepackage{rotating}
\usepackage{bbm}
\usepackage{latexsym}
\graphicspath{{figures/}}
\DeclareGraphicsExtensions{.pdf,.jpeg,.png,.jpg}

\usepackage[tight,footnotesize]{subfigure}
\usepackage{dblfloatfix}
\usepackage{algorithmic}
\usepackage{algorithm}
\usepackage{rotating}
\usepackage{dsfont}
\usepackage{placeins}

\newcommand{\captionfonts}{\normalsize}

\makeatletter
\long\def\@makecaption#1#2{%
  \vskip\abovecaptionskip
  \sbox\@tempboxa{{\captionfonts #1: #2}}%
  \ifdim \wd\@tempboxa >\hsize
    {\captionfonts #1: #2\par}
  \else
    \hbox to\hsize{\hfil\box\@tempboxa\hfil}%
  \fi
  \vskip\belowcaptionskip}
\makeatother
\begin{document}
\hspace{13.9cm}1

\ \vspace{20mm}\\

{\LARGE \centering STICK: Spike Time Interval Computational Kernel,\\A Framework for General Purpose Computation using \\Neurons, Precise Timing, Delays, and Synchrony} 

\ \\
{\bf \large Xavier Lagorce, Ryad Benosman}\\
{Vision and Natural Computation Group, Institut National de la Sant\'{e}
  et de la Recherche M\'{e}dicale, Paris F-75012, France, Sorbonne Universit\'{e}s,
  Institut de la Vision, Universit\'{e} Paris 06, Paris F-75012, France, and also with
  Centre National de la Recherche Scientifique, Paris F-75012, France.}\\
%{$^{\displaystyle 2}$Your second affiliation.}\\
%

%\ \\[-2mm]

\thispagestyle{empty}
\markboth{Lagorce \emph{et al.}}{Spike-Interval Computational Kernel}
\ \vspace{-0mm}\\
%
%Abstract
\begin{center} {\bf Abstract} \end{center}
\input{abstract}

%%%%%%%%%%%

\input{introduction}

\input{methods}

\input{results}

\input{discussion}

\input{conclusion}

\section*{Acknowledgments}

This work was performed in the frame of the LABEX LIFESENSES [ANR-10-LABX-65] and was
supported by French state funds managed by the ANR within the Investissements d'Avenir
programme [ANR-11-IDEX-0004-02]. XL has been supported by the European Union Seventh
Framework Programme (FP7/2007-2013) under grant agreement n$^\circ$ 604102 (HBP).
%The authors would also like to acknowledge discussion at the NSF Telluride Neuromorphic Cognition Engineering Workshop.

\appendix

\input{appendix}

%%\section*{Appendix}
%% You should put the details that are not required in the main body into this Appendix.

%\bibliographystyle{apacite}
\bibliographystyle{authordate1}
\bibliography{biblio}

\end{document}

%% file: abstract.tex
There has been significant research over the past two decades in developing new platforms for spiking neural computation. 
Current neural computers are primarily developed to mimick biology. They use  neural networks which can be trained to perform specific tasks to mainly solve pattern recognition problems. These machines can do more than simulate biology, they allow us to re-think our current paradigm of computation. The ultimate goal is to develop brain inspired general purpose computation architectures that can breach the current bottleneck introduced by the \emph{Von Neumann} architecture.
This work proposes a new framework for such a machine. We show that the use of neuron like units with precise timing representation, synaptic diversity, and temporal delays allows us to set a complete, scalable compact computation framework.
The presented framework provides both linear and non linear operations, allowing us to represent and solve any function. We show usability in solving real use cases from simple differential equations to sets of non-linear differential equations leading to chaotic attractors. %This goes beyond the existing Neural Engineering Framework (\citep{eliasmith2012}) that offers such a possibility using a spike rate representation of neural information thus requiring a large number of neurons and a heavy learning mechanism to implement simple functions.

%% file: introduction.tex
\section{Introduction}
\label{sec:introduction}
More than 50 years after the first \emph{Von Neumann} single processor, it is becoming more and more evident that this sequential power greedy architecture scales poorly to multiprocessors.
Despite the increase of the size of on-chip cache to stay away from RAM and to put the data closer to the processor, major processor manufacturers have run out of solutions to increase performances. The current solutions to use multicore devices and hyperthreading tries to overcome the problem by allowing programs to run in parallel. This parallelism is however limited as hyper-threaded CPUs even if they include extra registers still have only one essential element of most basic CPU features \citep{Sutter005}. \\
The quest for a more power efficient alternative has seen a major breakthrough these last years, specially in asynchronous brain like dataflow architectures. Recent endeavours, such as the SyNAPSE DARPA program, led to the development of silicon neuromorphic neural chip technology that allows to build a new kind of computer with similar function, and architecture to the brain. The advantage of these systems is their power efficiency and the scaling of performance with the number of neurons and synapses used.
There are currently several available platforms, to cite the most sucessful: IBM TrueNorth \citep{merolla2014Science}, Neurogrid \citep{neurogrid}, SpiNNaker \citep{DBLP:journals/pieee/FurberGTP14}, FACETS \citep{4633828}.
These machines seem to be primarly intended to simulate biology, their main application being currently in the field of machine learning and more specifically running deep learning architectures such as deep neural networks, convolutional deep neural networks, deep belief networks and recurrent neural networks. These techniques have shown to be efficient in several fields such as machine vision, speech recognition and natural language processing.
Other options exists such as the Neural Engineering Framework (NEF), that has shown to be able to simulate brain functionalities and provides networks that can accomplish visual, cognitive, and motor tasks \citep{eliasmith2012}. NEF intrinsically uses a spike rate-encoded information paradigm and a representation of functions using weighted spiking basis funtions; it thus requires a very large number of neurons to compute simple functions.
Other methods synthetize spiking neural networks for computation using Winner-Take-All (WTA) networks \citep{Indiveri01}, or more vision dedicated spiking structures such as convolutional neural networks \citep{journals/tbcas/Zamarreno-RamosLSL13}.
Linear Solutions of Higher Dimensional Interlayers networks \citep{DBLP:journals/corr/abs-1207-3368} are another class of approaches that are currently used to derive what is called Extreme Learning Machine (ELM) \citep{Huang:2008:ERS:1411851.1412017}. Recently the Synaptic Kernel Inverse Method (SKIM) framework \citep{10.3389/fnins.2013.00153} has been introduced, it uses multiple synapses to create the required higher dimensionality for learning time sequences.
These methods use random nonlinear projections into higher dimensional spaces \citep{NIPS2008_3495,ICML2011Saxe_551}. They create randomly initialized static weights to connect the input layer to the hidden layer, and then use nonlinear neurons in the hidden layer (which in the case of NEF are usually leaky integrate-and-fire neurons, with a high degree of variability in their population). The linear output layer allows for easy solution of the hidden-to-output layer weights; in NEF this is computed in a single step by pseudoinversion, using singular value decomposition.\\

Our method goes beyond the classical point of view that neurons transmit information exclusively via modulations of their mean firing rates \citep{Shadlen98,Mazurek02,Litvak03}. There seems to be growing evidence that neurons can generate spike-timing patterns with millisecond temporal precision in \citep{SEJ95,Mainen95,Lindsey97,Prut98,Villa99,Chang00,Tetko01}. Converging evidence suggests also that neurons in early stages of sensory processing in primary cortical areas (including vision and other modalities) use the millisecond precise time of neural responses to carry information \citep{Berry97,Reinagel00,Buracas98,Mazer02,BLANCHE08}.\\
Our approach will also make use of precisely timed transmission delays. The propagation delay between any individual pair of neurons is known to be precise and reproducible with a submillisecond precision \citep{Swadlow85,Swadlow94}. Axonal conduction delays in the mammalian neocortex \citep{Swadlow85,Swadlow88,Swadlow92} are known to range from 0.1 ms to 44 ms. Finally we will also use the property of biological neurons that states the same presynaptic axon can give rise to synapses with different properties, depending on the type of the postsynaptic target neuron \citep{Thomson93,Reyes98,citeulike:2933097}.
%There has been several studies on precise timing, transmission delays, and computation. Izhikevich (2006) showed that time-locked asynchronous spiking activity can have more synfire braids than the number of neurons or even the number of synapses in the network (Izhikevich 2006).

In this paper we are interested in deriving a new paradigm for computation using neuron like units and precise timing.
The goal is to design micro neural circuits operating in the precise timing domain to perform mathematic operations.
We show that when using computation units that have common properties with biological neurons such as precise timing, transmission delays, and synaptic diversity, it becomes possible to derive a Turing complete framework that can compute every known mathematical function using a non Von Neumann architecture.
 The presented framework allows to derive all mathematical operators whether they are linear or non linear. It also allows relational operations that are essential to develop algorithms.
The precise timing framework has a compact representation and it uses a low number of neurons to solve complex equations.
Examples will be shown on first and second order differential equations.\\
The developed methodology is in adequation with scalable neuromorphic architectures that make no distinction between memory and computation. Every synapse of each computational unit of the model simultaneously stores information and uses this information for computation. This contrasts with conventional computers that separate memory and processing thus causing the \emph{von Neumann bottleneck} where most of the computation time is spent in moving information between storage and the central processing unit rather than operating on it \citep{Backus78}.
The developed approach is easily scalable and is designed to naturally operate using an event-driven massive parallel communication similar to biological neural networks.\\

The next section describes the neural model used in this work and the encoding scheme chosen to
represent values in the exact timing of spikes. It also describes neural networks implementing
elementary operations that can be assembled to implement arbitrary calculus.
We then present results of applications of such networks, followed by a discussion on the methods
proposed in this work and a conclusion.

%% file: methods.tex
\section{Methods}
\label{sec:methods}

\subsection{Neural model}

These neuron-like computational units use the following neural model:
\begin{equation}
  \left\{\begin{matrix}
  \tau_m.\frac{\mathrm{d}V}{dt} & = & g_e + gate.g_f \\
  \frac{\mathrm{d}g_e}{dt} & = & 0 \\
  \tau_f.\frac{\mathrm{d}g_f}{dt} & = & -g_f \\
  \end{matrix}\right.
  \label{eq:neural_model}
\end{equation}

$V$ is the membrane potential of the neuron. We consider here that there is no leakage
of the membrane potential (or that the time constant of this leakage is much slower than all the
other time constants considered in this work, in which case it can be neglected).
$g_e$ represents a constant input current which can only be changed by synaptic events.
$g_f$ represents input synapses with exponential dynamics. These synapses are gated
by the $gate$ signal which is triggered by synaptic events.
For the experiments presented in this work, we use $\tau_m = 100\mathrm{s}$ and $\tau_f = 20\mathrm{ms}.$

We thus distinguish $4$ type of synapses, where $w$ is the weight of the synapse:
\begin{itemize}
  \item $V-\mathrm{synapses}$ directly modify the membrane potential value: $V \leftarrow V+w,$
  \item $g_e-\mathrm{synapses}$ directly modify the constant input current: $g_e \leftarrow g_e+w,$
  \item $g_f-\mathrm{synapses}$ directly modify exponential input current: $g_f \leftarrow g_f+w,$
  \item $gate-\mathrm{synapses}:$ $w=1$ activate the exponential synapses by setting $gate \leftarrow 1;$
                       $w=-1$ deactivate the exponential synapses by setting $gate \leftarrow 0.$
\end{itemize}
All synaptic connections are also defined by a propagation delay between the source and target neurons.

A neuron spikes when its membrane potential reaches a threshold, i.e.:
\begin{equation}
  V \geq V_t
\end{equation}
it then emits a spike and is reset by putting back its state variables to:
\begin{eqnarray}
  V &\leftarrow& V_\mathrm{reset} \\
  g_e &\leftarrow& 0 \\
  g_f &\leftarrow& 0 \\
  gate &\leftarrow& 0
\end{eqnarray}
without loss of generality, we will consider $V_t = 10\mathrm{mV}$ and $V_\mathrm{reset}=0$ to
simplify the following equations.

In the following subsections, we use the following notations. $T_\mathrm{syn}$ is the propagation
delay between two neurons for standard synapses. $T_\mathrm{neu}$ is the time needed by a neuron to emit
a spike when triggered by an input synaptic event; it can model, for instance, timesteps of a neural
simulator. In the experiments presented in this work, we use $T_\mathrm{syn}=1\mathrm{ms}$ and
$T_\mathrm{neu}=10\mathrm{us}.$
We define $w_e$ as the minimum excitatory weight for $V-\mathrm{synapses}$ required to trigger a
neuron in its reset state, and $w_i$ the inhibitory weight of counteracting effect:
\begin{eqnarray}
  w_e &=& V_t \\
  w_i &=& -w_e
\end{eqnarray}
Standard weights for $g_e-\mathrm{synapses}$ will be defined in the next subsection.

\subsection{Signal representation}

The main idea of the method proposed in this work is to represent values as the precise
time interval in between two spikes. 
% We then use the interspike interval of a pair of spikes emitted by a particular neuron to encode a value.

If the series $e_n(i)$ is the list of times at which neuron $n$ emitted spikes, with $i$
the index of the spike in the series, neuron $n$ encodes the signal $u(t)$ by:
\begin{equation}
  u(e_n(i)) = f^{-1}(e_n(i+1)-e_n(i)), \forall i=2.p,\,p\in\mathds{N},
\end{equation}
with $i$ an even number and $f^{-1}$ the inverse of the encoding function $f$ of our choice.

The encoding function $f:\mathds{R} \rightarrow \mathds{R}$ can be chosen
depending on the considered signals in a particular system and adapted to the required precision.
$f$ computes the interspike time $\Delta t$ associated with a particular value.
In the following work, we chose to represent values using the following linear encoding function:
\begin{equation}
  \Delta t = f(x) = T_\mathrm{min} + x.T_\mathrm{cod},
\end{equation}
with $x\in[0,1]$ and $T_\mathrm{cod}$ the elementary time step.

This representation allows us to encode any value between a minimum and a maximum interspike (of
$T_\mathrm{min}$ and $T_\mathrm{max}=(T_\mathrm{min}+T_\mathrm{cod})$). We chose to use a minimum interspike
to encode zero for several reasons. If the two spikes encoding a value originate from one unique
neuron or are received by a single neuron, this minimum interspike gives time to recover from the first
spike before spiking again. $T_\mathrm{min}$ allows networks to react to the first input spike and
propagate a state change before the second encoding spike is received.
In the experiments presented in this work, we use $T_\mathrm{min}=10\mathrm{ms}$ and
$T_\mathrm{cod}=100\mathrm{ms}.$

We could also choose a logarithmic function to allow encoding a large range of values with dynamic precision
(precision would be smaller for large values).

\bigskip

To represent signed values, we use two different pathways for the two different signs. Positive values
will be encoded by causing a neuron to spike and negative values by eliciting another neuron to spike. We arbitrary
chose to represent zero as a positive value.

To ease the understanding and the routing of several networks, each implementing simple operations, we add some interface neurons to these
networks. For instance, the input of the networks are materialized by special \emph{input} neurons. Their
output by some \emph{output}, \emph{output~+} or \emph{output~-} neurons. Other special neurons used for
interaction between networks are also marked. In all the following figures describing networks, neurons
in blue will be input neurons to the networks whereas neurons in red will be output neurons of the circuit.

\begin{figure}[h!]
  \centering
  \includegraphics[width=0.4\columnwidth]{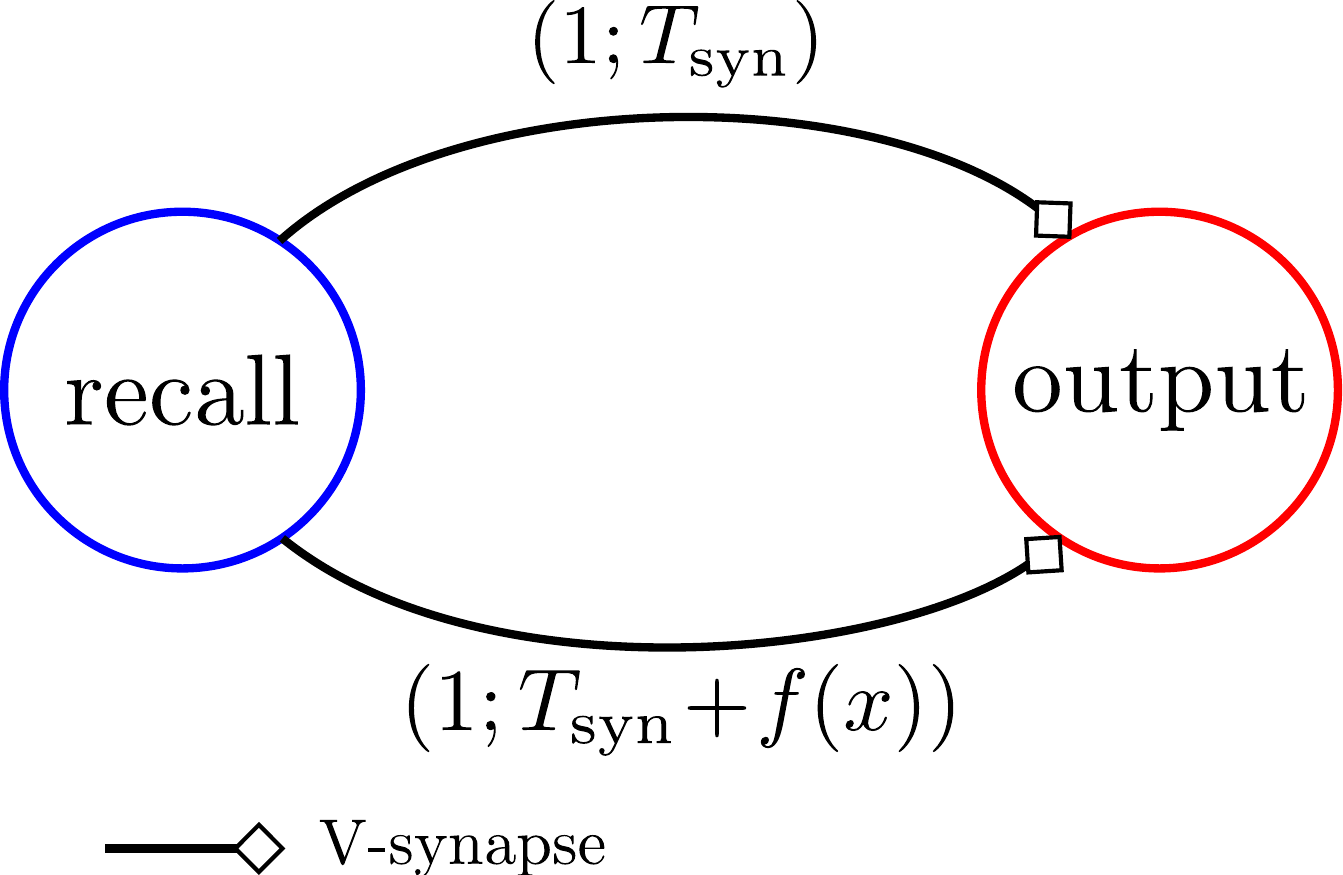}
  \caption{A network generating a constant value when required. When the \emph{recall} neuron is
           activated, the \emph{output} neuron will generate a pair of spikes coding for value $x$.}
  \label{fig:net_constant}
\end{figure}

The simple network presented in Fig.~\ref{fig:net_constant} encodes a constant value. It shows
the design principles which will be used in the rest of this section.
In the network shown in Fig.~\ref{fig:net_constant}, the \emph{recall} neuron is an input. When a
spike is received by \emph{recall}, the constant value encoded in the network is output to the
\emph{output} neuron. In this example, the output is generated by two different
synaptic connections from \emph{recall} to \emph{output}. They generate two output spikes with the
interspike corresponding to the encoded value.

\bigskip

We define standard weights for $g_e-synapses.$
Let $w_\mathrm{acc}$ be the weight value for $g_e$ synapses to cause a neuron to spike from its reset
state after time $T_\mathrm{max}=T_\mathrm{min}$+$T_\mathrm{cod}.$ According to Eq.~\ref{eq:neural_model},
we have:
\begin{equation}
  V_t = \frac{w_\mathrm{acc}}{\tau_m}.T_\mathrm{max}
\end{equation}
such that:
\begin{equation}
  w_\mathrm{acc} = V_t.\frac{\tau_m}{T_\mathrm{max}}
\end{equation}
We also define weight $\bar{w}_\mathrm{acc}$ as the $g_e$ value necessary to have a neuron spiking from its
reset state after time $T_\mathrm{cod}.$ The same equation gives us:
\begin{equation}
  \bar{w}_\mathrm{acc} = V_t.\frac{\tau_m}{T_\mathrm{cod}}
\end{equation}

\bigskip

We can now describe different neural networks implementing elementary operations
such as : memories, synchronizer, linear combination or non-linearities such as multiplications, directly operating on inter spike intervals.

\subsection{Storing data: memory}

\paragraph{Inverting Memory}

\begin{figure}[h!]
  \centering
  \includegraphics[width=0.75\columnwidth]{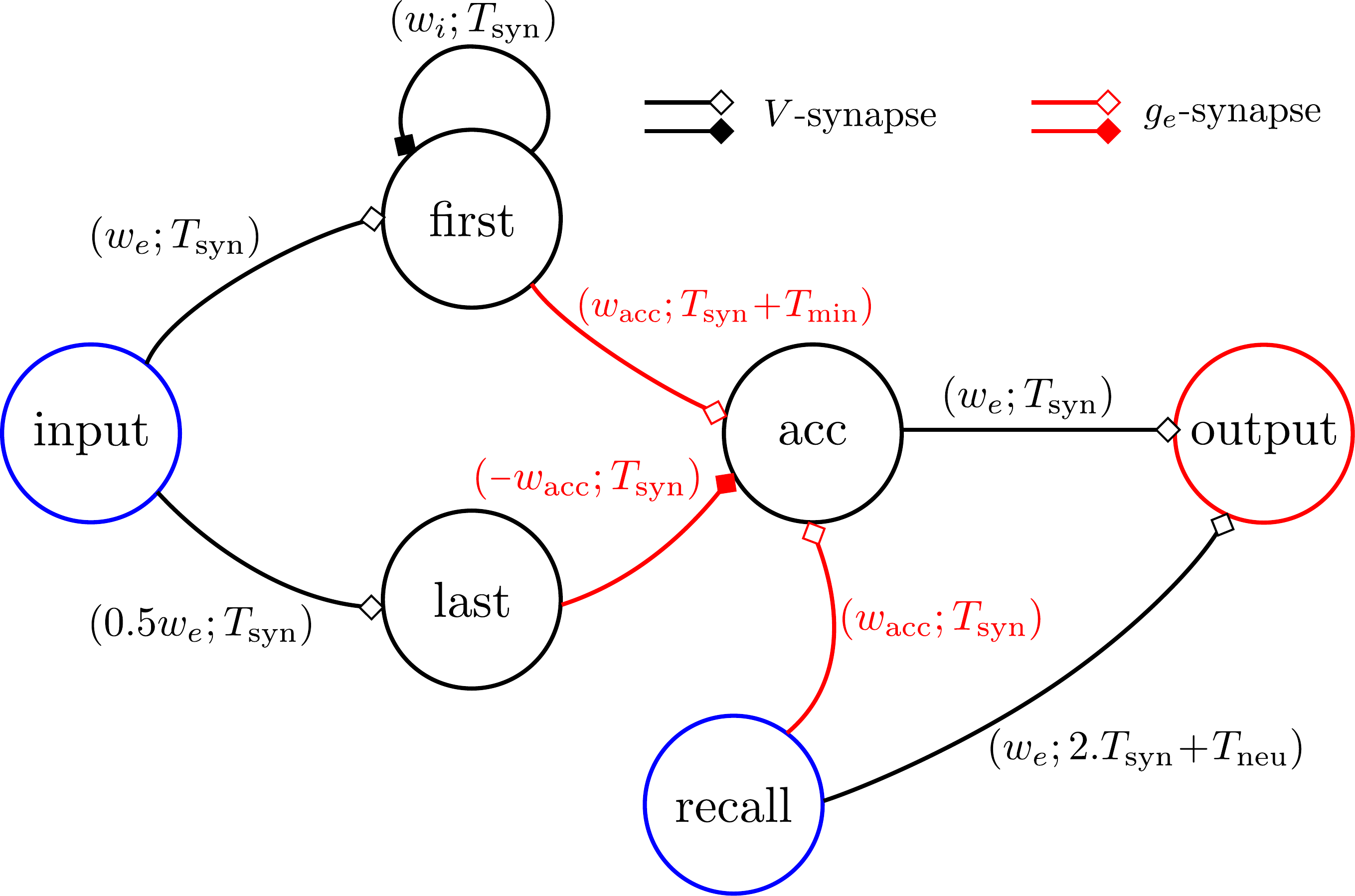}
  \caption{\textbf{Inverting Memory:} this network stores a value encoded in a pair of input
           spikes $x$ (to the \emph{input} neuron) by integrating current on the $g_e$ dynamics of neuron
           \emph{acc} during the input interspike. The value, stored in the membrane potential of the
           \emph{acc} neuron is read out when the \emph{recall} neuron is triggered. Releasing a pair
           of spikes with neuron \emph{output} corresponding to $1-x$.
           Blue, red and black neurons are, respectively, input, output and internal neurons.}
  \label{fig:net_inverting_memory}
\end{figure}

\begin{figure}[h!]
  \centering
  \includegraphics[width=0.6\columnwidth]{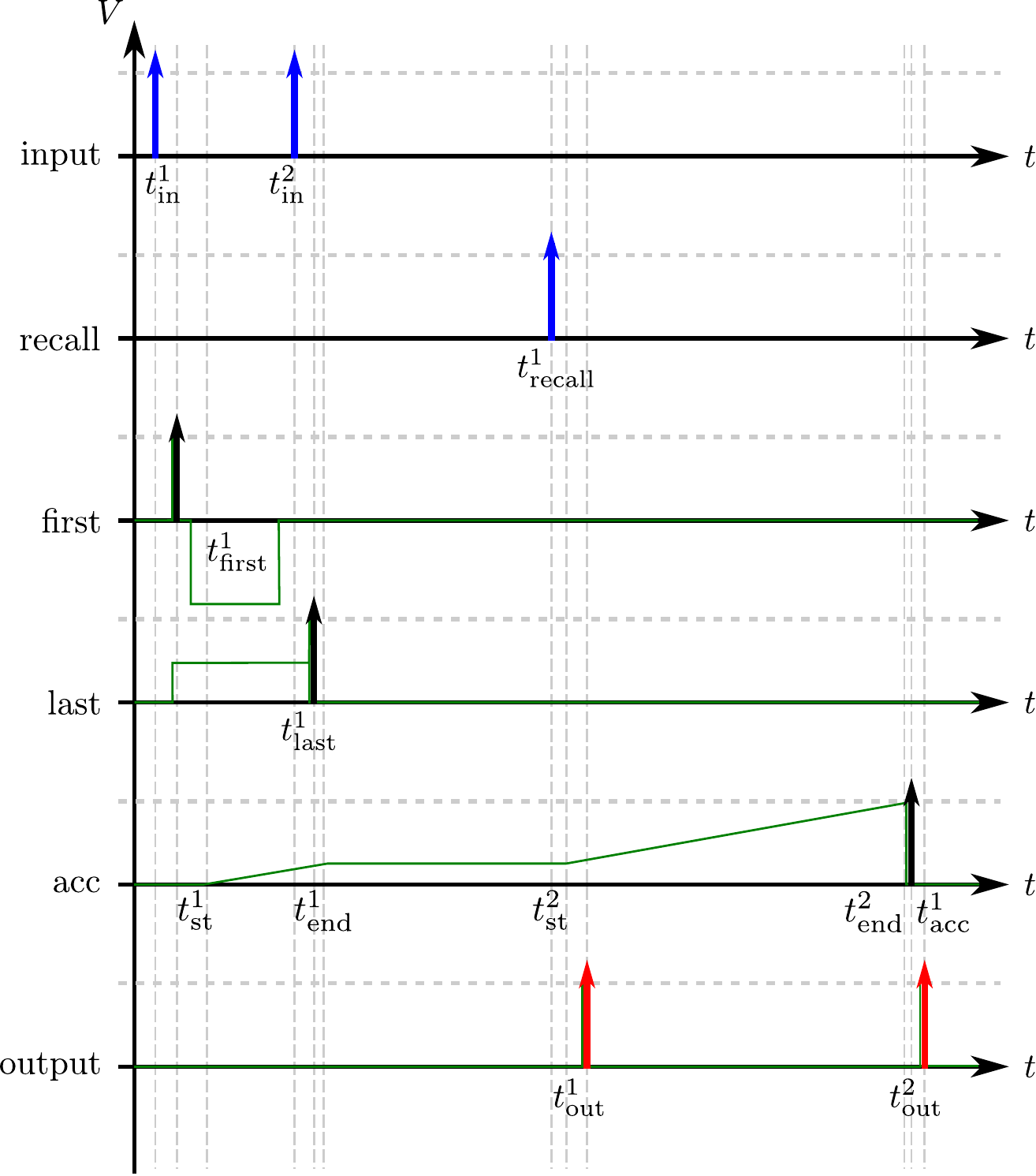}
  \caption{\textbf{Inverting Memory:} chronogram of network operation for an input at times
           $t_\mathrm{in}^1$ and $t_\mathrm{in}^2$ and a recall at time $t_\mathrm{recall}^1.$
           (Input spikes are drawn in blue, output spikes in red. Green plots show the membrane
           potential of interesting neurons.)}
  \label{fig:chrono_inverting_memory}
\end{figure}

Fig.~\ref{fig:net_inverting_memory} presents an Inverting Memory network. This network is
constituted of two input neurons (in blue): \emph{input} and \emph{recall}, one output neuron (in red):
\emph{output} and $3$ internal neurons (in black): \emph{first}, \emph{last} and \emph{acc}.
Details and proof of the inner working of the network can be found in Appendix \ref{sec:app:inverting_memory}.
The chronogram of spikes during operation of this network is presented in Fig.~\ref{fig:chrono_inverting_memory}.
When a pair of spikes arrives at the \emph{input} neuron, they are sorted by the \emph{first} and
\emph{last} neurons. Their synaptic connections are such that \emph{first} will only spike in response
to the first encoding spike of the pair (at time $t_\mathrm{in}^1$) and \emph{last} will only spike in
response to the second encoding spike of the pair (at time $t_\mathrm{in}^2$), thus seperating the two
input spikes.

\emph{first} and \emph{last} are then respectively starting and stopping the integration of the membrane
potential of neuron \emph{acc} at times $t_\mathrm{st}^1$ and $t_\mathrm{end}^1$ such that the value of
\emph{acc}'s membrane potential after the second input spike is, with
$\Delta T_{in}=t_\mathrm{in}^2-t_\mathrm{in}^1$ the input interspike:
\begin{equation}
  V_{sto} = \frac{w_\mathrm{acc}}{\tau_m}.(\Delta T_{in} - T_\mathrm{min})
\end{equation}
When the \emph{recall} neuron is triggered, the integration starts again until reaching $V_t$ such that
we get an output interspike $\Delta T_{out}=t_\mathrm{out}^2-t_\mathrm{out}^1$ following:
\begin{equation}
  V_t = \frac{w_\mathrm{acc}}{\tau_m}.(\Delta T_{in} - T_\mathrm{min})
         + \frac{w_\mathrm{acc}}{\tau_m}.\Delta T_{out}
\end{equation}
Considering the definition of $w_\mathrm{acc}$, we have
$V_t.\tau_m/w_\mathrm{acc}=T_\mathrm{max}$, such that
\begin{equation}
  \Delta T_{out} = T_\mathrm{max} - (\Delta T_{in} - T_\mathrm{min})
\end{equation}
We thus obtain an output spike corresponding to the maximum temporal representation of a value
($T_\mathrm{max})$ minus the actual coding time $(\Delta T_{in} - T_\mathrm{min})$ of the input value.
%This network is then able to store a value and output it, however transformed, when desired.
Chaining two of these Inverting Memory networks, can store and recall a value without modification.

We can also notice that the value is stored in the inter spike timing needed to represent the value and can be recalled
as soon as the value has been completely fed into the Inverting Memory network.

\paragraph{Memory}

\begin{figure}[h!]
  \centering
  \includegraphics[width=0.75\columnwidth]{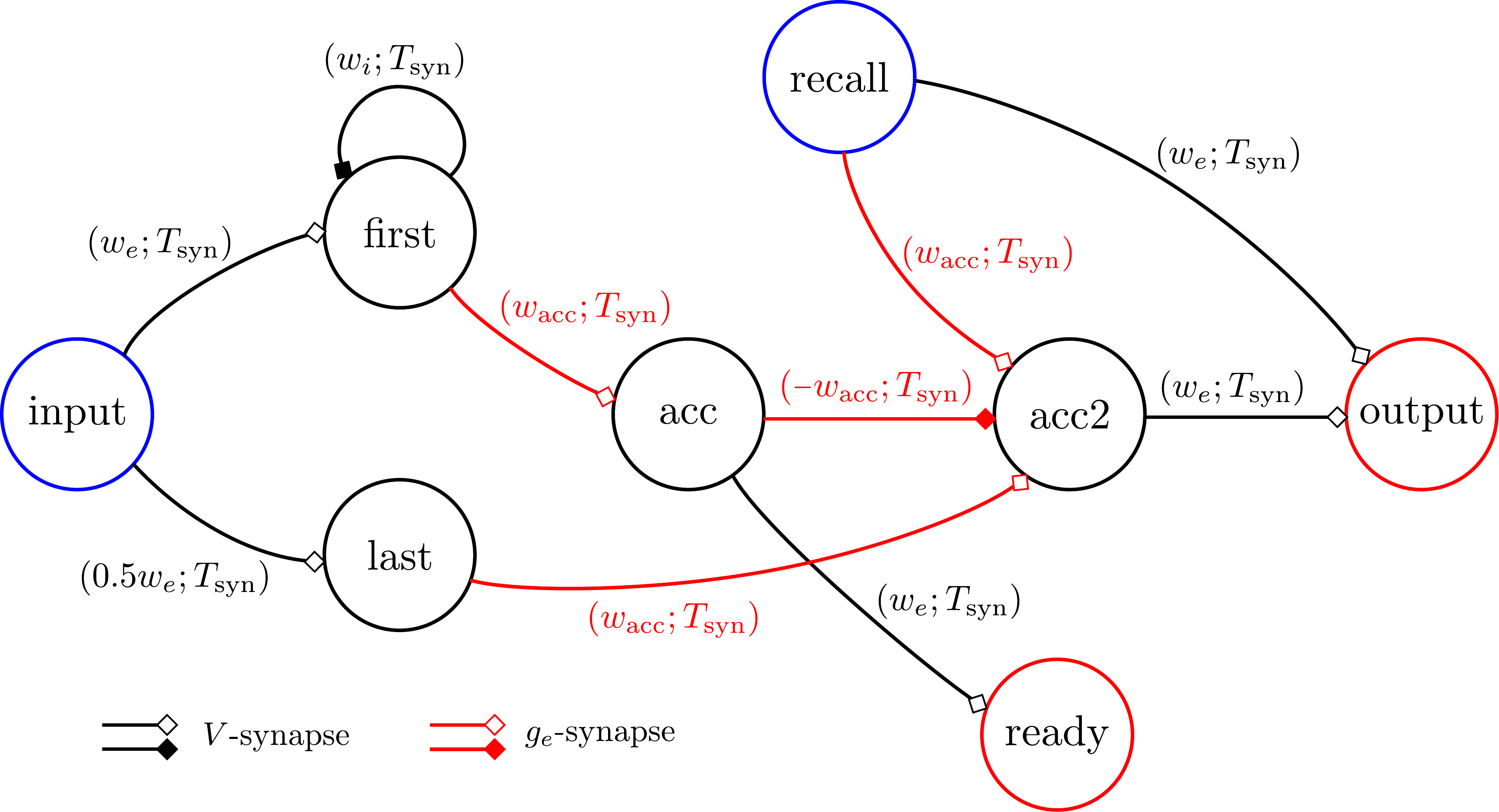}
  \caption{\textbf{Memory:} this network stores a value stored in the inter spike timing of two input spikes (to
           the \emph{input} neuron) similarly to the Inverting Memory network. Two accumulators, \emph{acc} and \emph{acc2} are used to output on the \emph{output}
           neuron the same value as previously received on the \emph{input} neuron.
           Blue, red and black neurons are, respectively, input, output and internal neurons.}
  \label{fig:net_memory}
\end{figure}

Fig.~\ref{fig:net_memory} presents a Memory network. Detailed explanations and a chronogram
of the network's operations can be found in Appendix~\ref{sec:app:memory}. This network is based on
the Inverting Memory network introduced in the previous paragraph, $2$ accumulator neurons
\emph{acc} and \emph{acc2} are added to invert the stored value twice. If we follow the same reasoning
as in the previous paragraph, \emph{acc} spikes $T_\mathrm{max}$ after the first encoding spike is
received and \emph{acc2} starts integrating when the second encoding spike is received. Because
\emph{acc} stops \emph{acc2}'s integration process, the value stored in \emph{acc2}'s membrane potential
after \emph{acc} spikes is, with $\Delta T_{in}$ the input interspike (we present here a simplifyed result
to ease the notations, the full result is available in Appendix~\ref{sec:app:memory}):
\begin{equation}
  V_{sto} = \frac{w_\mathrm{acc}}{\tau_m}.(T_\mathrm{max}-\Delta T_{in})
\end{equation}
Note that the time at which \emph{acc2} ends its integration happens after the input value has been
completely fed into the Memory network. This is the reason why we added the \emph{ready} neuron which is triggered
when the input value has been stored in the Memory network and is ready to be read out.
We then have the output interspike $\Delta T_{out}$ following:
\begin{equation}
  V_t = \frac{w_\mathrm{acc}}{\tau_m}.(T_\mathrm{max}-\Delta T_{in})
         + \frac{w_\mathrm{acc}}{\tau_m}.\Delta T_{out}
\end{equation}
such that
\begin{equation}
  \Delta T_{out} = \Delta T_{in}
\end{equation}

%% {\color{blue}Note: We could replace \emph{acc} by delays of $T_\mathrm{syn}+T_\mathrm{max}$, saving one neuron
%%   but needing a very long transmission delay between two close neurons... could be interesting to reduce
%%   the number of neurons but hard for ASIC implementation because of the added long delay...}

\paragraph{Signed Memory}

\begin{figure}[h!]
  \centering
  \includegraphics[width=0.75\columnwidth]{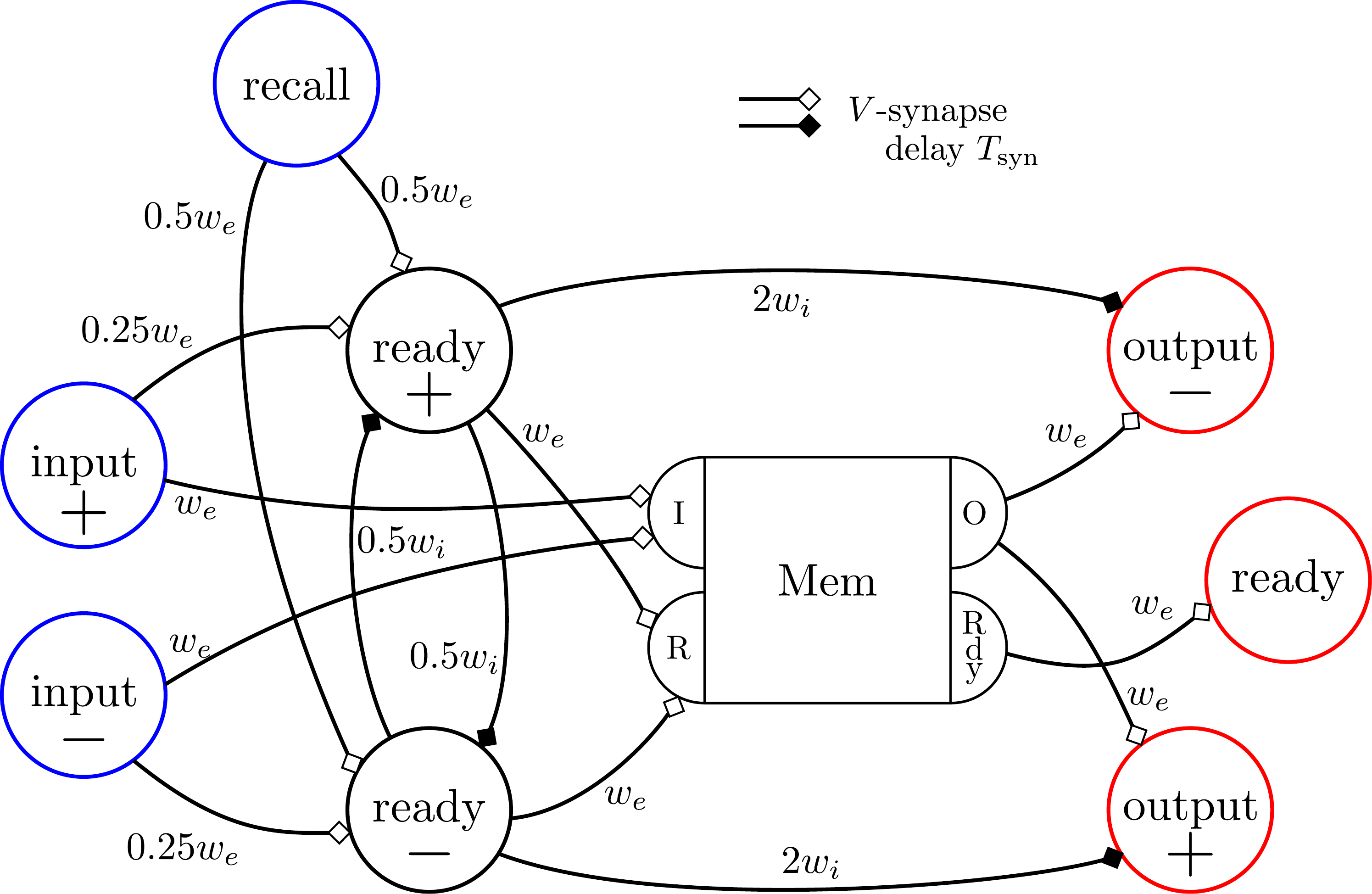}
  \caption{\textbf{Signed Memory:} this network is using a Memory network to store a signed value and its
           sign. Depending on the sign of the input (i.e. if it is received on the \emph{input~+} or
           \emph{input~-} neuron), the wrong output neuron (\emph{output~+} if the input is
           negative) is inhibited so that the output of the internal Memory network is directed to
           the output corresponding to the input sign.
           Blue, red and black neurons are, respectively, input, output and internal neurons.}
  \label{fig:net_signed_memory}
\end{figure}

Fig.~\ref{fig:net_signed_memory} presents a Signed Memory network. This network uses a Memory
network to store a value and a small state machine, implemented by neurons \emph{ready~+} and
\emph{ready~-}, to store the sign of the input. Detailed equations and the chronogram of the network's
operations can be found in Appendix~\ref{sec:app:signed_memory}. The internal Memory network is linked
in parallel to the positive and negative pathways of the Signed Memory. When an input is fed into one
of these two pathways, only the corresponding \emph{ready} neuron receives some excitation. When the
\emph{recall} neuron is triggered, only the \emph{ready} neuron corresponding to the sign of the input
spikes (because of the excitation contributed by the stored input). The \emph{ready} neuron will then
inhibit the wrong output such that only the \emph{output} neuron corresponding to the correct sign will
fire.

\paragraph{Synchronizer}

\begin{figure}[h!]
  \centering
  \includegraphics[width=0.6\columnwidth]{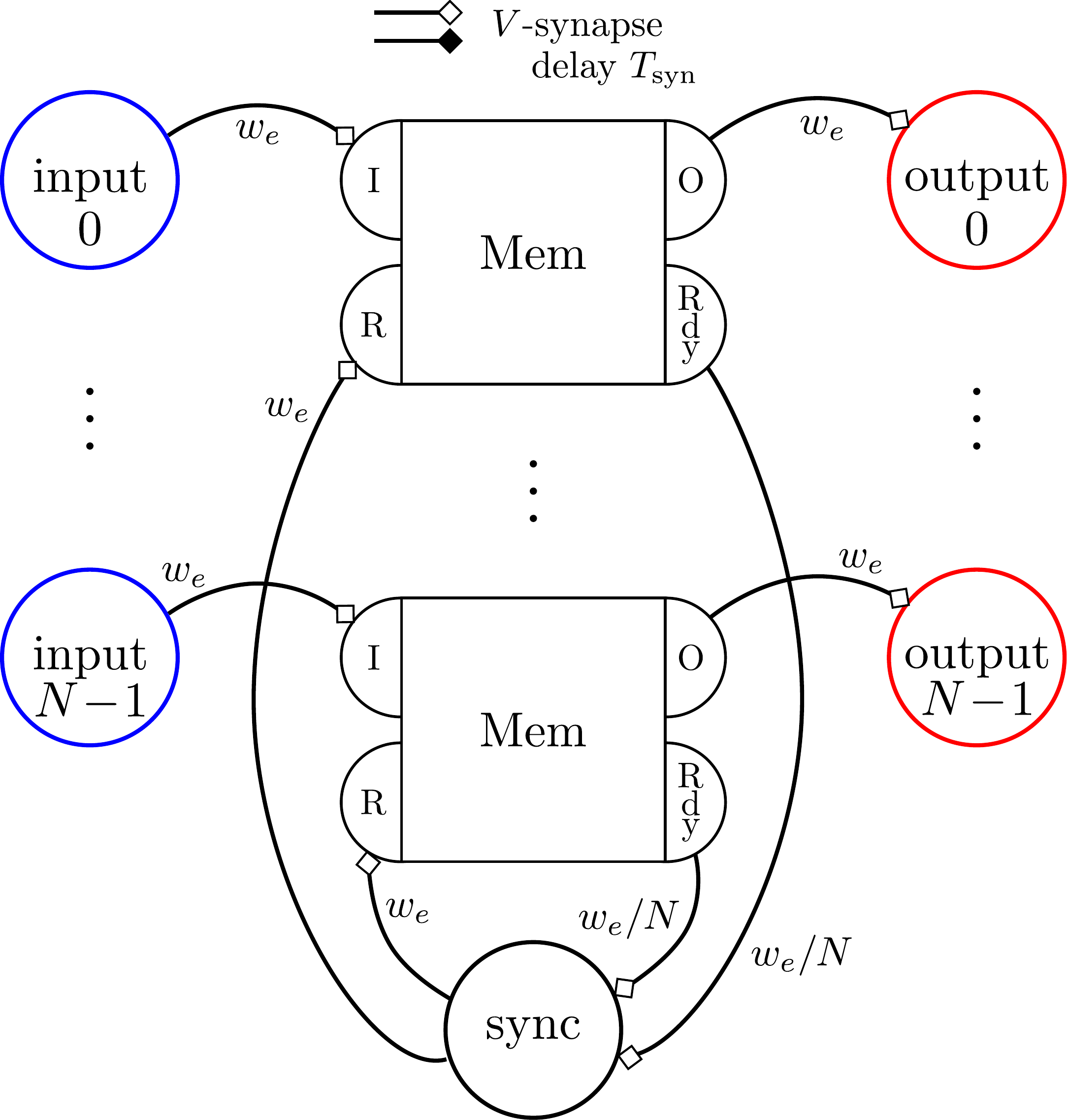}
  \caption{\textbf{Synchronizer:} this network is synchronizing a set of input values so that the first
           spikes encoding every value on its output side happen at the exact same time. It can be used
           to resynchronize values before networks requiring this condition. It uses a set of Memory
           networks to store the different input values and a \emph{sync} neuron which recalls all
           these stored values as soon as the last one of them has been stored.
           Blue, red and black neurons are, respectively, input, output and internal neurons.}
  \label{fig:net_synchronizer}
\end{figure}

Fig.~\ref{fig:net_synchronizer} presents a Synchronizer network. This network receives $N$
different inputs and synchronizes their first encoding spikes on the output end. Detailed equations and the chronogram of the network's operations can be found in Appendix~\ref{sec:app:synchronizer}. It is
implemented using $N$ Memory networks. The
\emph{sync} neuron keeps track of the number of received inputs. When all the $N$ inputs have been
received, this neuron spikes, starting the readout process of the different memories at the same time,
thus synchronizing all the outputs.

Furthermore, the same principle can be used with Signed Memory networks to obtain a Signed Synchronizer
network.

\subsection{Relational operations}

\paragraph{Minimum}

\begin{figure}[h!]
  \centering
  \includegraphics[width=0.6\columnwidth]{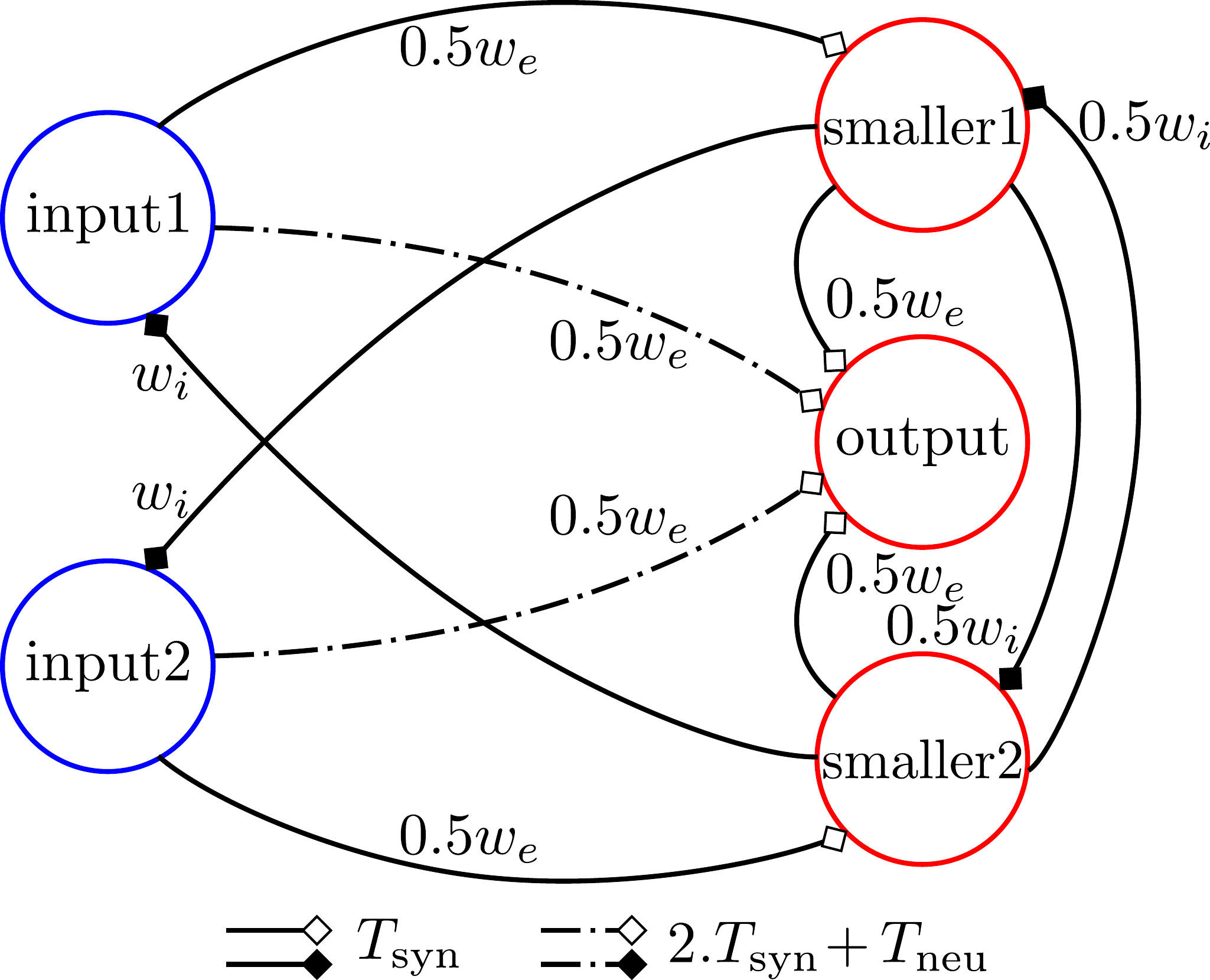}
  \caption{\textbf{Minimum:} this network outputs the smallest of its two inputs, which have to be
           synchronized, and an indicator signal on neuron \emph{smaller1} or \emph{smaller2} corresponding
           to which input has been considered as the smallest one. Because we chose an encoding function
           $f$ which output increases with its input, the smallest input corresponds to the one for which
           the second encoding spike arrives first. This is what the \emph{output} neuron is extracting.
           Blue, red and black neurons are, respectively, input, output and internal neurons.
           This network only contains $V-\mathrm{synapses}$}
  \label{fig:net_minimum}
\end{figure}

Fig.~\ref{fig:net_minimum} presents a Minimum network. This network implements the minimum
operation on 2 inputs. It outputs the smallest of its 2 inputs as well as an indicator of which input
was the smallest one. If the two inputs, \emph{input1} and \emph{input2}, are synchronized and because
our encoding function is increasing with its input value, the minimum of the 2 inputs is the one
for which the second encoding spikes arrives first. This is what the \emph{smaller1} and \emph{smaller2}
neurons are extracting. This information is also used to inhibit the excitatory contribution of the largest
input to the \emph{output} neuron in order to output only the smallest of the input values. Detailed
proof and the chronogram of operations can be found in Appendix~\ref{sec:app:minimum}.

\paragraph{Maximum}

\begin{figure}[h!]
  \centering
  \includegraphics[width=0.6\columnwidth]{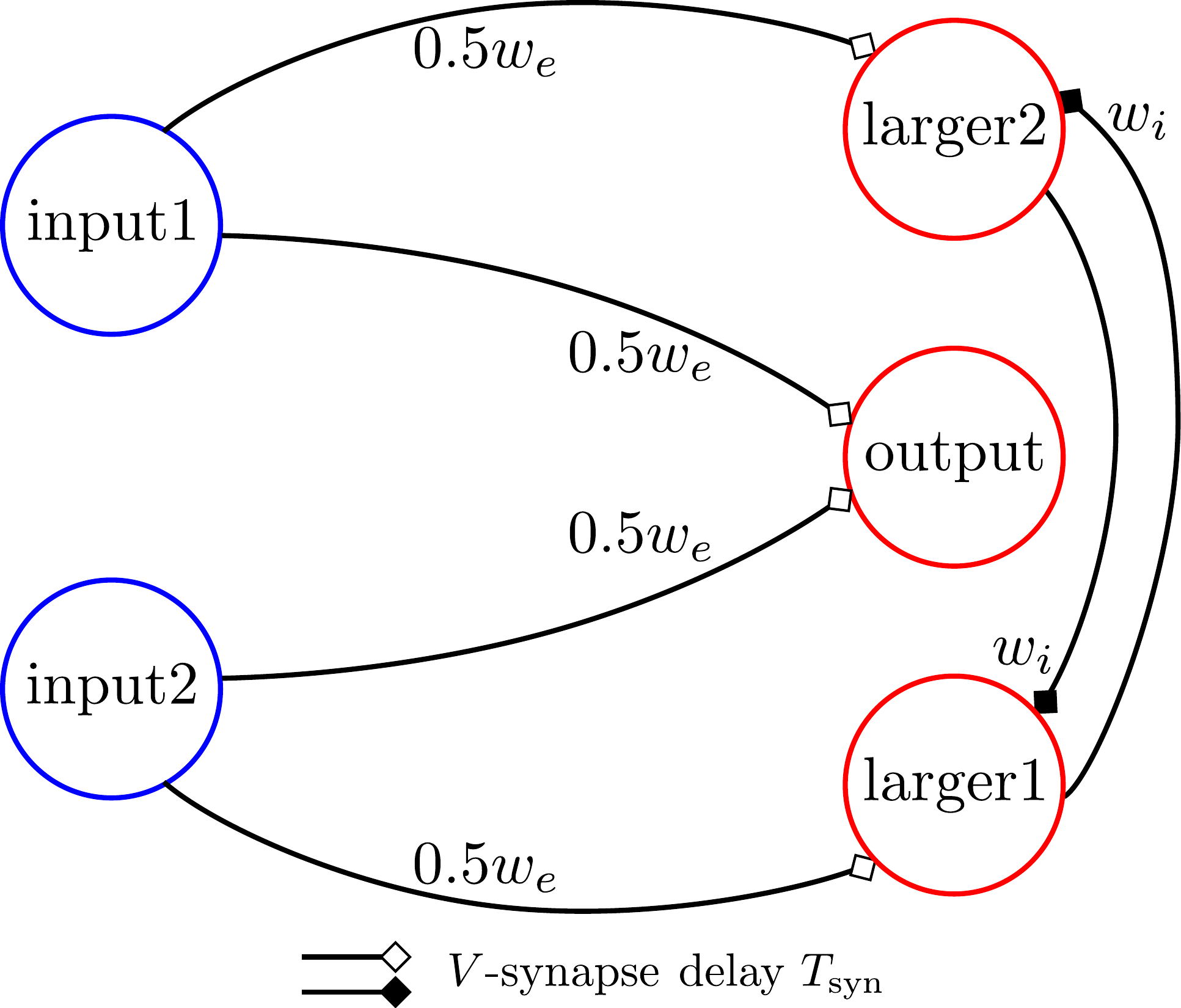}
  \caption{\textbf{Maximum:} this network outputs the largest of its two inputs, which have to be
           synchronized, and an indicator signal on neuron \emph{larger1} or \emph{larger2} corresponding
           to which input has been considered as the largest one. Because we chose an encoding function
           $f$ which output increases with its input, the largest input corresponds to the one for which
           the second encoding spike arrives last. This is what the \emph{output} neuron is extracting.
           Blue, red and black neurons are, respectively, input, output and internal neurons.}
  \label{fig:net_maximum}
\end{figure}

Fig.~\ref{fig:net_maximum} implements the maximum operation on 2 inputs. This networks follows the same principle as in the Minimum network, it differs by inverting the detection relation: when the first input is the smallest, it triggers \emph{larger2} because
the second input must then be the largest. The drive of the \emph{output} neuron is simpler in this
case as the inputs are synchronized. The maximum value corresponds to the one for which the
second encoding spike is the latest. Detailed proof and the chronogram of operations can be found
in Appendix~\ref{sec:app:maximum}.

\subsection{Linear operations}

\paragraph{Subtractor}

\begin{figure}[h!]
  \centering
  \includegraphics[width=0.75\columnwidth]{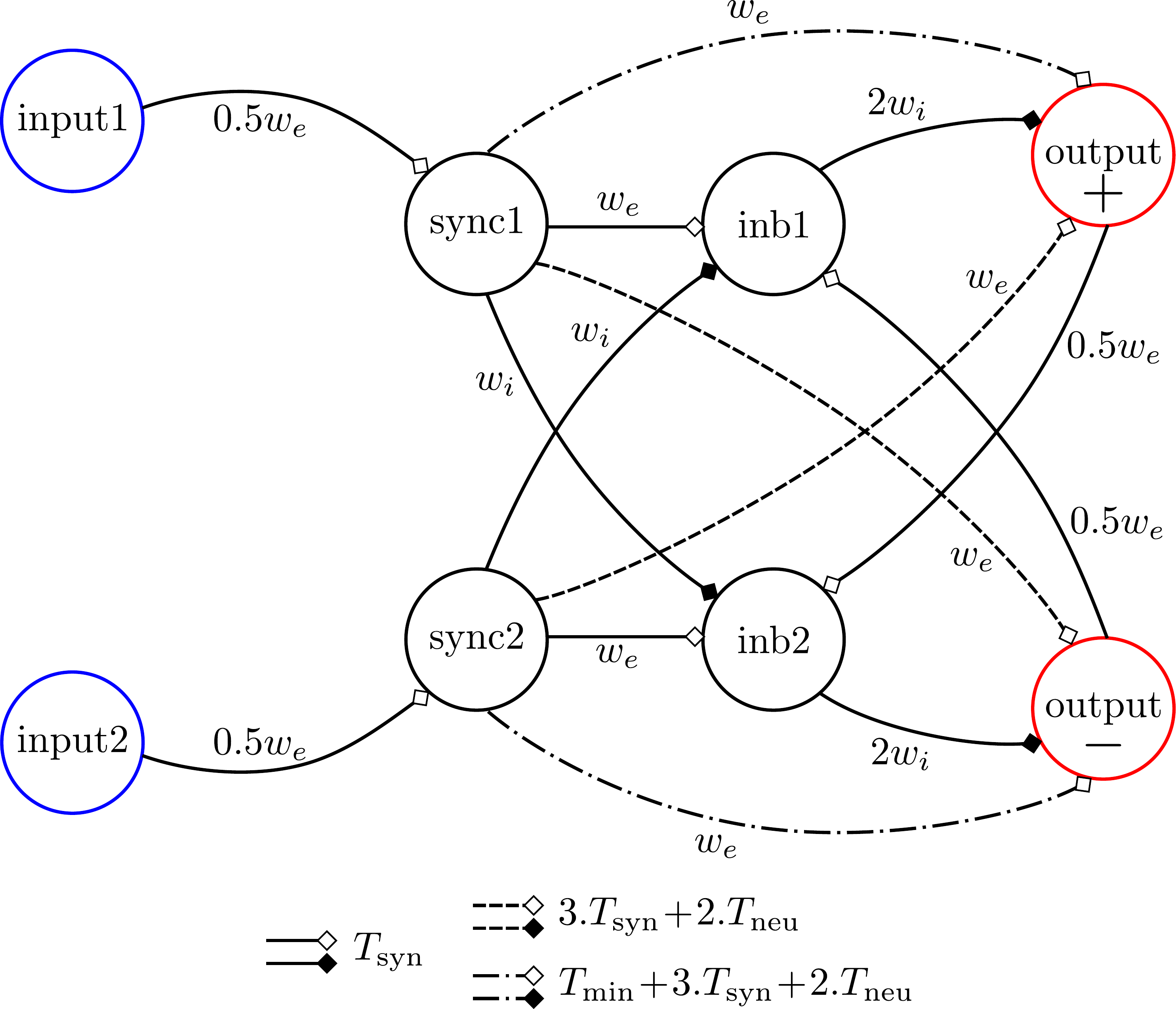}
  \caption{\textbf{Subtractor (simple):} this network substracts its two inputs, which have to be synchronized.
           The resulting value is directed to two different neurons, \emph{output~+} or \emph{output~-},
           depending on its sign. Because the two inputs are synchronized, the difference is directly
           given by the interspike between the second spikes of both inputs. The sign is determined using
           the same idea as in the Minimum network (see Fig.~\ref{fig:net_minimum}).
           Blue, red and black neurons are, respectively, input, output and internal neurons.
           This network only contains $V-\mathrm{synapses}$}
  \label{fig:net_subtractor_simple}
\end{figure}

Fig.~\ref{fig:net_subtractor_simple} presents a Subtractor network. It is here presented in its
simplest form. It will be expanded in a second stage. This network computes the difference
between \emph{input1} and \emph{input2} and directs the output depending on its sign either to the
\emph{output~+} or \emph{output~-} neuron. If the two inputs are synchronized, the difference between
the two is directly given by the interspike between the two second encoding spikes. This is the information \emph{sync1} and \emph{sync2} neurons are extracting. It also implements the same idea as in the Minimum
network (see Fig.~\ref{fig:net_minimum}) to compute the sign of the output. When the output sign is known,
\emph{sync1} or \emph{sync2} inhibits the pathway to the wrong output neuron such that the output spikes
are directed to the correct one. The detailed proof and the chronogram of operations can be found in
Appendix~\ref{sec:app:subtractor}.

\begin{figure}[h!]
  \centering
  \includegraphics[width=0.75\columnwidth]{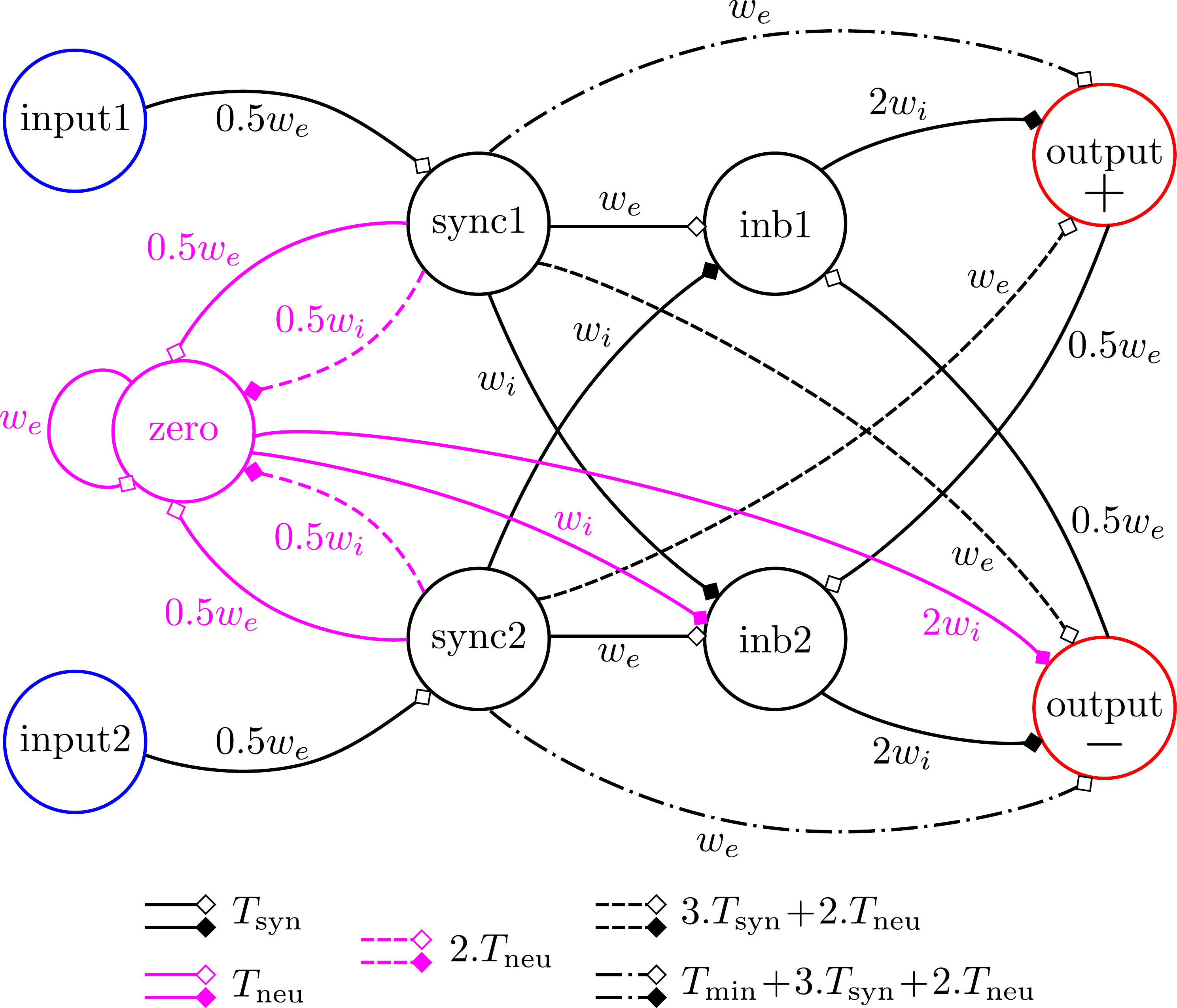}
  \caption{\textbf{Subtractor (full):} this nework implements the same operation as the one presented
           in Fig.~\ref{fig:net_subtractor_simple}. The simple version requires that the input are not
           equal so that the inhibition resulting from the sign determination has time to propagate.
           The added \emph{zero} neuron and its connections (in magenta) detects this particular case
           and drives the network to output a zero on neuron \emph{output~+} if the two inputs are
           equal.
           Blue, red and black neurons are, respectively, input, output and internal neurons.
           This network only contains $V-\mathrm{synapses}$}
  \label{fig:net_subtractor}
\end{figure}

A more robust version of the Subtractor network is presented in Fig.~\ref{fig:net_subtractor}.
The networks adds the \emph{zero} neuron and its connections (in magenta in the figure). 
In the previous 'simple' implementation shown in Fig.~\ref{fig:net_subtractor_simple},
when both inputs are equals, the two parallel pathways of the network are triggered at the same time
and the lateral inhibition has no time to select a winning pathway. In that case, the output is emitted
both on \emph{output~+} and \emph{output~-} which can be problematic for the following networks expecting
only one of the two pathways to be activated. To solve this problem, we add the \emph{zero} neuron with a set of fast
synaptic connections. They allow to detect the case of equality between the two inputs. In this case,
the \emph{output~-} pathway is quickly inhibited to produce spikes coding for the zero output only on the
\emph{output~+} neuron.

\paragraph{Linear Combination}

\begin{figure}[h!]
  \centering
  \includegraphics[width=0.8\columnwidth]{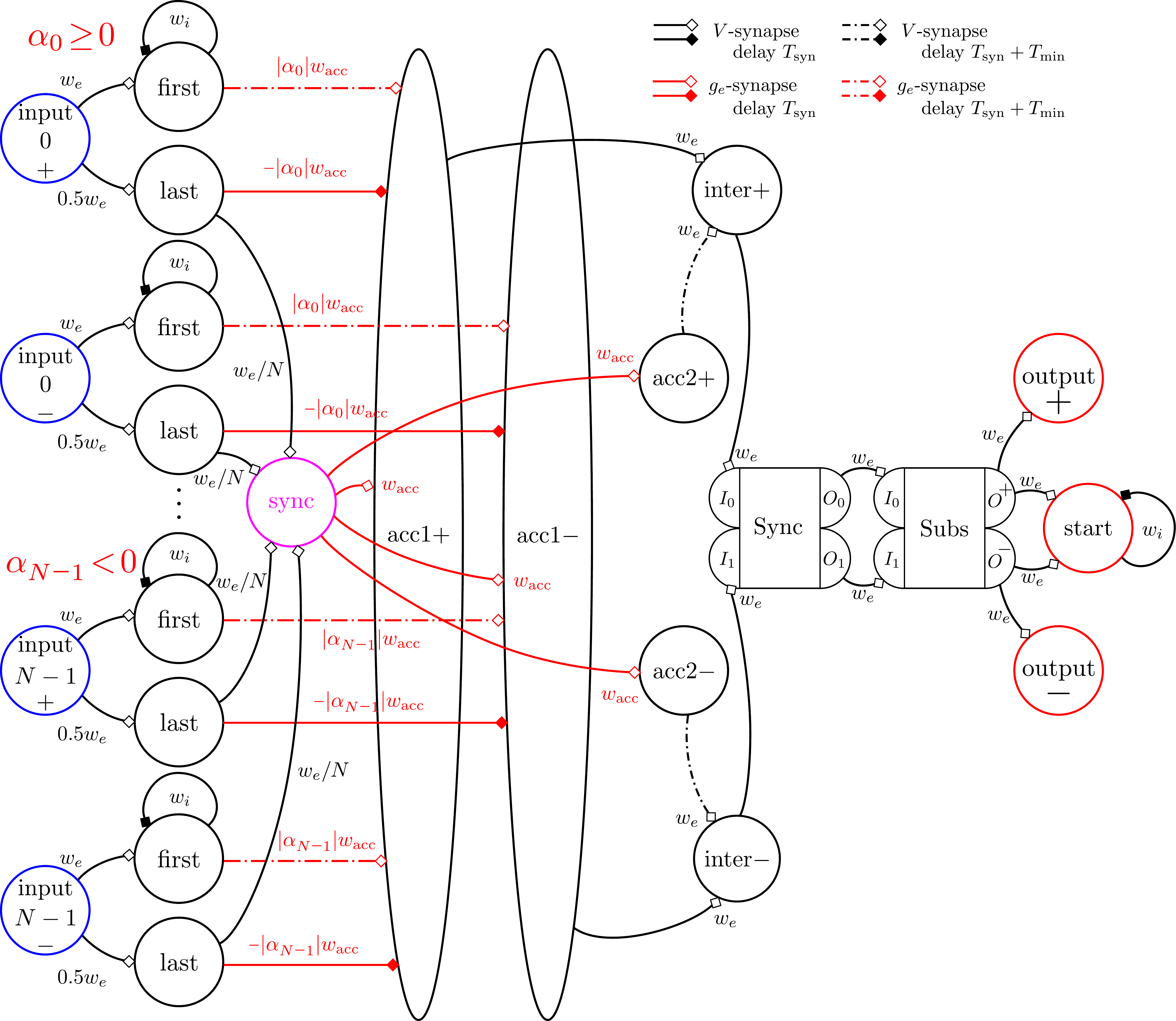}
  \caption{\textbf{Linear Combination:} this network computes the linear combination of a set of inputs
           with given coefficients $\alpha_0,...,\alpha_{N-1}$: $s=\sum_{i=0}^{N-1}\alpha_i.x_i$. The network is accumulating
           positive and negative sub-operations (i.e. $\alpha_i.x_i$) in 2 distinct accumulators.
           Synaptic connections in the network are shown for $\alpha_0 \geq 0$ and $\alpha_N < 0.$
           The two resulting values are then read out by the \emph{sync} neuron when all inputs are known,
           synchronized and substracted from one another to obtain the full, signed, result $s.$
           An indicator neuron \emph{start} is provided to notify when the result is ready.
           Blue, red and black neurons are, respectively, input, output and internal neurons.}
  \label{fig:net_linear_combination}
\end{figure}

Fig.~\ref{fig:net_linear_combination} presents a Linear Combination network. It computes the
linear combination of $N$ signed inputs with arbitrary coefficients $\alpha_0,...\alpha_{N-1}:$
\begin{equation}
  s = \sum_{i=0}^{N-1} \alpha_i.x_i,
\end{equation}
where $x_i,i\in{0,...,N-1}$ are the different inputs of the network. It uses the same principle
as in the Memory network to store values in an accumulator. To implement the coefficients of the sum, we
multiply the synaptic weight of the accumulation current by the coefficient corresponding to the input.
To handle the signs of the inputs and coefficients, we use 2 accumulators. The first one is storing
intermediate results which are positives (i.e. when the sign of the input is the same as the one of
its associated coefficient) while the second stores negative values (i.e. when the sign of the input is
different from the one of its associated coefficient). When all the inputs have been fed into the
network, the \emph{sync} network is triggered, causing the readout process of these accumulators. Their
content are inverted as for the Memory network and then synced before entering a Subtractor network.
This last network computes the difference between the positive and negative contributions of the
different inputs and produces a signed output. A \emph{start} neuron is then triggered to spike to indicate that the computation has ended. This signal can be used to trigger further networks. All details and proofs can be
found in Appendix~\ref{sec:app:linear_combination}.\\

%One can also note that another input neuron can be added, exciting the \emph{sync} neuron. If weights
%are tuned such that spikes on all inputs and this input neuron are required for the \emph{sync} neuron
%to fire, a memory can be embedded in the Linear Combination network without needing much more resources.

\subsection{Non-linear operations}

\paragraph{Natural Logarithm}

\begin{figure}[h!]
  \centering
  \includegraphics[width=0.8\columnwidth]{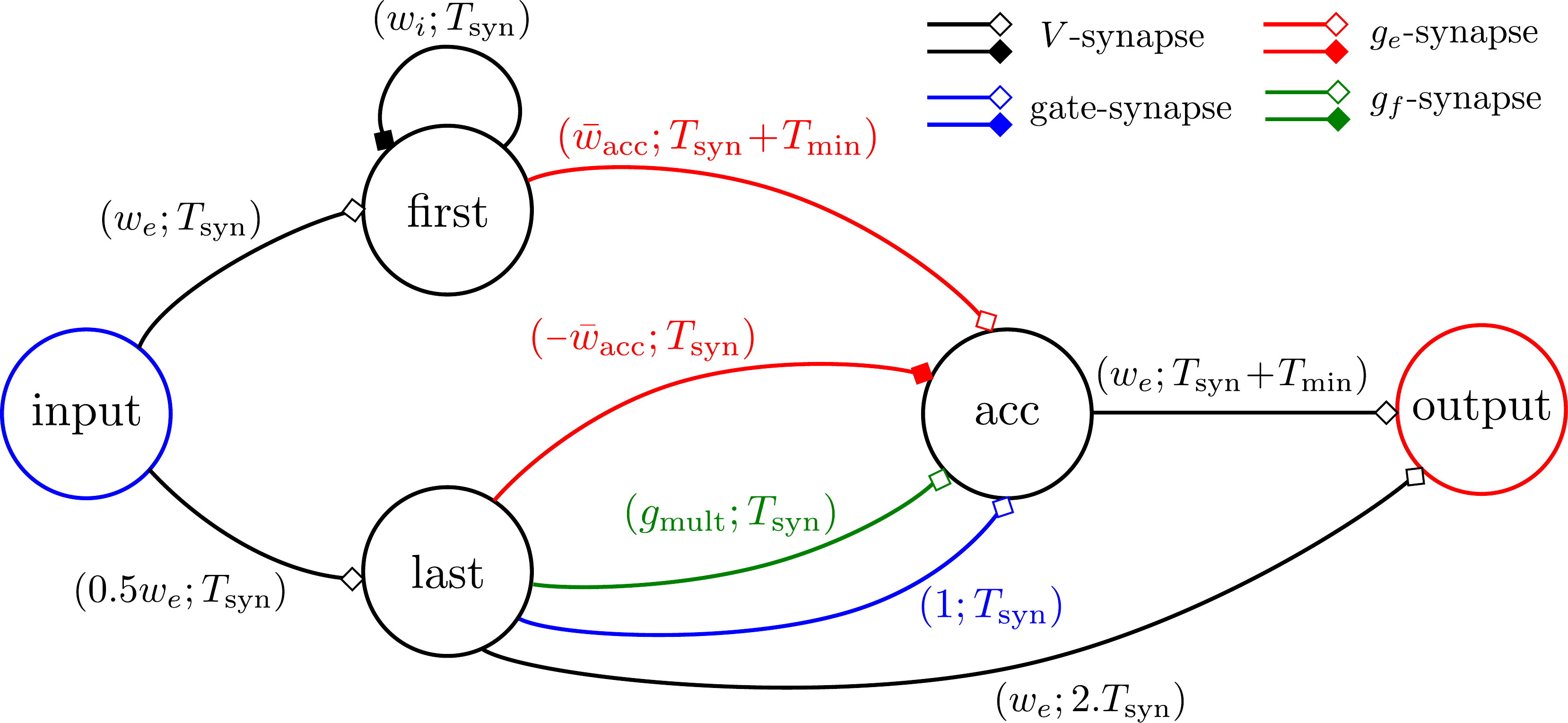}
  \caption{\textbf{Log:} this network computes the natural logarithm of its input by using
           the $g_f$ dynamics of neuron \emph{acc}. The input value is first stored in the membrane
           potential of the \emph{acc} neuron. When the second encoding spike arrives, a
           $g_f-\mathrm{synapse}$ is used to obtain a delay function of the $\log$ of the
           current value of \emph{acc}'s membrane potential.
           Blue, red and black neurons are, respectively, input, output and internal neurons.}
  \label{fig:net_log}
\end{figure}

Fig.~\ref{fig:net_log} presents a Log network capable of computing the natural logarithm
of its input value by exploiting the dynamics of a $g_f-\mathrm{synapse}.$ Detailed proof
and the chronogram of operations can be found in Appendix~\ref{sec:app:natural_log}. When an input value is
fed into the \emph{input} neuron, this value is first stored in the membrane potential of
\emph{acc} by integrating with weight $\bar{w}_\mathrm{acc}$ between the spike of \emph{first} and
\emph{last}. Because the spike from \emph{first} is delayed by an additional $T_\mathrm{min}$
compared to the one from \emph{last}, the value stored in \emph{acc}'s membrane potential is,
with $\Delta T_{in}=T_\mathrm{min}+\Delta T_{cod}$ the input interspike:
\begin{equation}
  V = \frac{\bar{w}_\mathrm{acc}}{\tau_m}(\Delta T_{in}-T_\mathrm{min})
    = \frac{\bar{w}_\mathrm{acc}}{\tau_m}.\Delta T_{cod}
    = V_t.\frac{\Delta T_{cod}}{T_\mathrm{cod}}
\end{equation}
When this integration process is stopped by \emph{last}'s spike, a synaptic event is also triggered on
a $g_f-\mathrm{synapse}$ of \emph{acc} with weight $g_\mathrm{mult}$ (another synaptic event also enables
the $g_f$ dynamics by activating the $gate$ state). \emph{acc}'s membrane potential thus follows the
evolution given by solving the differential system Eq.~(\ref{eq:neural_model}):
\begin{equation}
  V = V_t.\frac{\Delta T_{cod}}{T_\mathrm{cod}} + g_\mathrm{mult}\frac{\tau_f}{\tau_m}(1-e^{-t/\tau_f})
  \label{eq:log_gf}
\end{equation}
If we chose $g_\mathrm{mult}$ such that:
\begin{equation}
  g_\mathrm{mult} = V_t.\frac{\tau_m}{\tau_f}
  \label{eq:g_mult}
\end{equation}
We obtain
\begin{equation}
  V = V_t.\frac{\Delta T_{cod}}{T_\mathrm{cod}} + V_t.(1-e^{-t/\tau_f})
\end{equation}
Considering neuron \emph{acc} will then spike at time $t_s$ where $V=V_t$, we get:
\begin{eqnarray}
  V_t &=& V_t.\frac{\Delta T_{cod}}{T_\mathrm{cod}} + V_t.(1-e^{-t_s/\tau_f}) \\
  \frac{\Delta T_{cod}}{T_\mathrm{cod}} &=& e^{-t_s/\tau_f} \\
  t_s &=& -\tau_f.\log\left(\frac{\Delta T_{cod}}{T_\mathrm{cod}}\right)
\end{eqnarray}
which is a positive value because $\Delta T_{cod} \leq T_\mathrm{cod}$ by definition.
Adding the delays of the synaptic connections to the \emph{output} neuron, we get an output interspike
$\Delta T_{out}$:
\begin{equation}
  \Delta T_{out} = T_\mathrm{min} + \tau_f.\log\left(\frac{T_\mathrm{cod}}{\Delta T_{cod}}\right)
\end{equation}
We thus obtain a network capable of generating an output proportional to the natural logarithm of
its input.

\paragraph{Exponential}

\begin{figure}[h!]
  \centering
  \includegraphics[width=0.8\columnwidth]{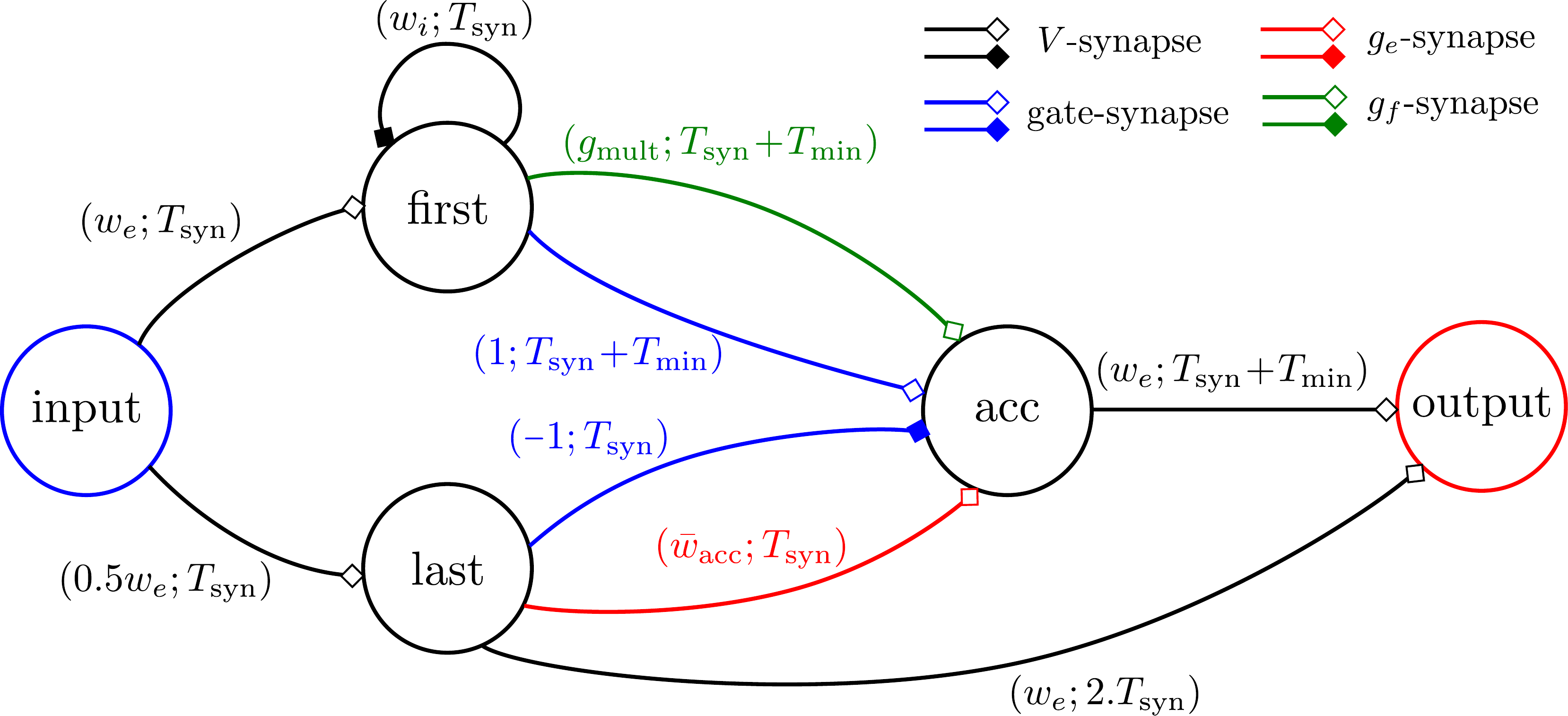}
  \caption{\textbf{Exp:} this network computes the exponential of its input by using the $g_f$
           dynamics of neuron \emph{acc}. A value function of the exponential of the input value
           is first stored in neuron \emph{acc}'s membrane potential by gating a $g_f-\mathrm{synapse}$
           to \emph{acc} on the second encoding spike of the input. \emph{acc}'s membrane potential is
           then read out to obtain the result.
           Blue, red and black neurons are, respectively, input, output and internal neurons.}
  \label{fig:net_exp}
\end{figure}

Fig.~\ref{fig:net_exp} presents the Exp network that computes the exponential of its input
value by exploiting the dynamics of a $g_f-\mathrm{synapse}.$ The detailed proof and the chronogram
of operations can be found in Appendix~\ref{sec:app:exponential}. When an input value is fed into the
\emph{input} neuron, a synaptic event is triggered on a $g_f-\mathrm{synapse}$ of the \emph{acc} neuron
with weight $g_\mathrm{mult}$ as defined in Eq.~\ref{eq:g_mult} by neuron \emph{first}. \emph{acc}'s
membrane potential thus follows the evolution given by solving the differential
system Eq.~\ref{eq:neural_model} until \emph{last} spikes:
\begin{equation}
  V = g_\mathrm{mult}\frac{\tau_f}{\tau_m}(1-e^{-t/\tau_f}) = V_t.(1-e^{-t/\tau_f})
\end{equation}
This synaptic activity is then gated when neuron \emph{last} is triggered and blocks \emph{acc}'s
membrane potential value to :
\begin{equation}
  V = V_t.(1-e^{-\Delta T_{cod}/\tau_f}),
\end{equation}
with $\Delta T_{in}=T_\mathrm{min}+\Delta T_{cod}$ the input interspike
(Because of the additional delay of $T_\mathrm{min}$ in \emph{first}'s pathway in comparison to
the one of \emph{last}).
The spiking of \emph{last} also triggers a readout of \emph{acc}'s membrane potential by initiating
a $g_e$ synaptic event with weight $\bar{w}_\mathrm{acc}$ such that \emph{acc} spikes after time $t_s$
following:
\begin{eqnarray}
  V_t &=& V_t.(1-e^{-\Delta T_{cod}/\tau_f}) + \frac{\bar{w}_\mathrm{acc}}{\tau_m}.t_s \\
  V_t &=& V_t.(1-e^{-\Delta T_{cod}/\tau_f}) + V_t.\frac{t_s}{T_\mathrm{cod}} \\
  t_s &=& T_\mathrm{cod}.e^{-\Delta T_{cod}/\tau_f}
\end{eqnarray}
Adding the delays of the synaptic connections to the \emph{output} neuron, we get an output interspike
$\Delta T_{out}$:
\begin{equation}
  \Delta T_{out} = T_\mathrm{min} + T_\mathrm{cod}.e^{-\Delta T_{cod}/\tau_f}
\end{equation}
We thus obtain a network capable of generating an output proportional to the exponential of its input.

\bigskip

Going back to the Log network from the previous paragraph, its output was :
\begin{equation}
  \Delta T_{cod}^{\log} = \Delta T_{out}^{\log}-T_\mathrm{min}
                     = \tau_f.\log\left(\frac{T_\mathrm{cod}}{\Delta T_{cod}}\right)
\end{equation}
If this output is fed into an Exp network, we obtain the output:
\begin{eqnarray}
  \Delta T_{out}^{\exp} &=& T_\mathrm{min} + T_\mathrm{cod}.e^{-\Delta T_\mathrm{cod}^{\log}/\tau_f}\\
  &=& T_\mathrm{min} + T_\mathrm{cod}.e^{\log(\Delta T_{cod}/T_\mathrm{cod})}\\
  &=& T_\mathrm{min} + \Delta T_{cod}
\end{eqnarray}

The Exp network is thus capable of inverting the Log network. 
One can take advantage of that to implement several common non-linearities by applying simple operations in between the Log and Exp networks.
For instance, summing two logarithms will allow multiplication, subtracting them will implement division,
multiplying a logarithm by a constant will compute a power function, ...

\paragraph{Multiplier}

\begin{figure}[h!]
  \centering
  \includegraphics[width=0.8\columnwidth]{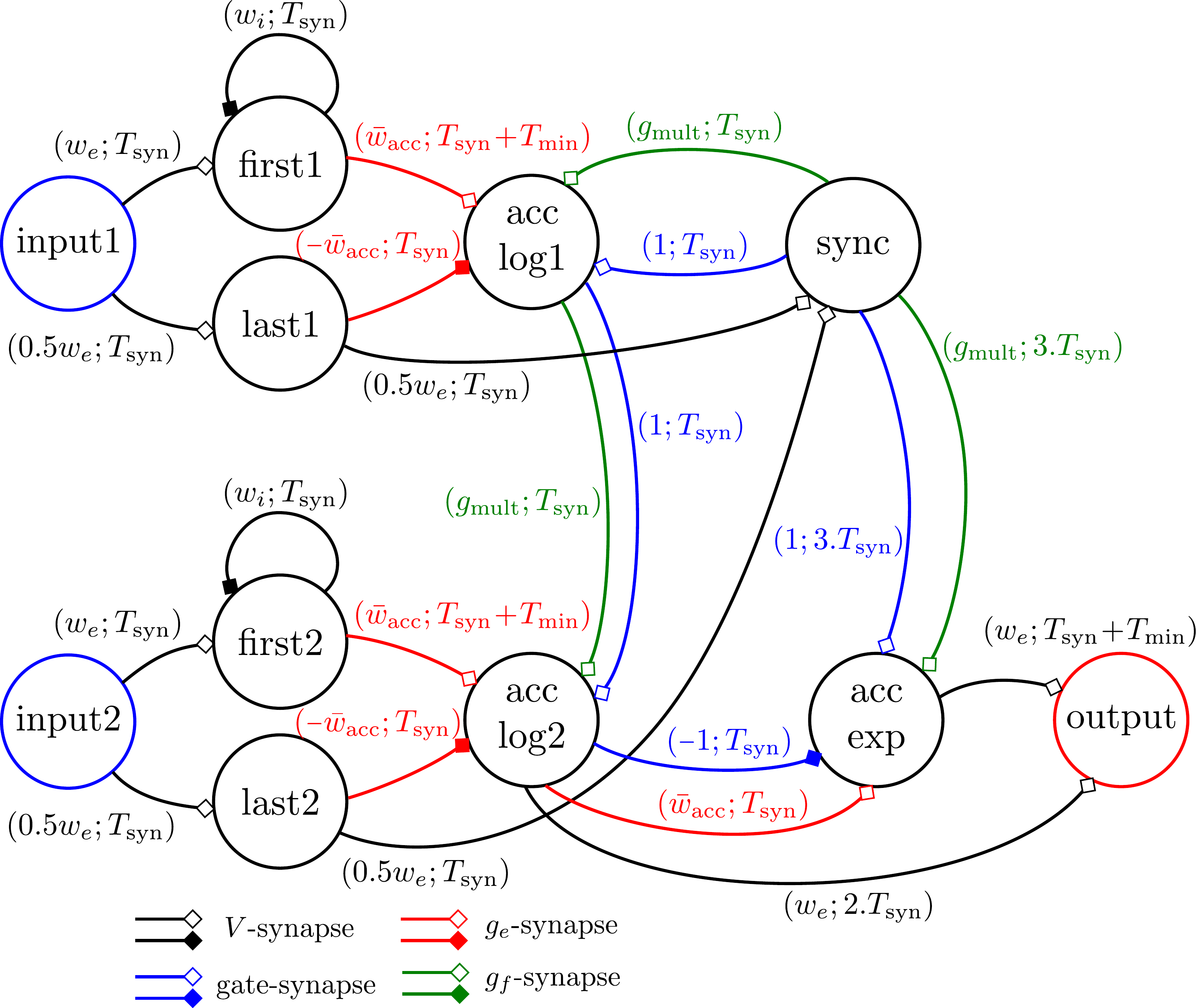}
  \caption{\textbf{Multiplier:} this network multiplies its two inputs. It computes the $\log$s of its
           inputs through neurons \emph{acc\_log1} and \emph{acc\_log2}, sum them up and gets the $\exp$
           of this value through neurons \emph{acc\_exp} to obtain the product :
           $s=x_1.x_2=\exp(\log x_1 + \log x_2).$
           Blue, red and black neurons are, respectively, input, output and internal neurons.}
  \label{fig:net_multiplier}
\end{figure}

Fig.~\ref{fig:net_multiplier} presents a network we name Multiplier. The detailed proof
and the chronogram of the network's operations can be found in Appendix~\ref{sec:app:signed_memory}.
This network is based on the principles used in the Log and Exp networks. The product $s$ of the
2 inputs $x_1$ and $x_2$ is obtained using the well known following equation:
\begin{equation}
  s = x_1.x_2 = \exp(\log x_1 + \log x_2).
\end{equation}
Neurons \emph{acc\_log1} and \emph{acc\_log2}'s membrane potentials are first loaded with the two inputs
(\emph{input1} and \emph{input2}). When the 2 inputs have been received, an exponential circuit is
triggered through \emph{acc\_exp}. To obtain the product of the input, this circuit has to be stopped
after a time corresponding to the sum of the natural logarithm of the 2 inputs. Because the absolute
value of the natural logarithm can be larger than $1$ for small inputs (i.e. it is larger than the
maximum value representable by our encoding scheme), we cannot use a Linear Combination network to sum
the logs. To overcome this problem, we sum these values by triggering the logarithm computation of the
2 inputs successively: the \emph{sync} neuron, which is detecting the end of the second input, it activates at the same time the exponential circuit and the logarithm of the first input. When the first input's
logarithm is output, it triggers the logarithm of the second input which, when computed, stops the
exponential circuit. At that point in time, the \emph{acc\_exp} neuron contains in its membrane the product of the 2 inputs. It is then read out to compute the actual output of the network.

\paragraph{Signed Multiplier}

\begin{figure}[h!]
  \centering
  \includegraphics[width=0.7\columnwidth]{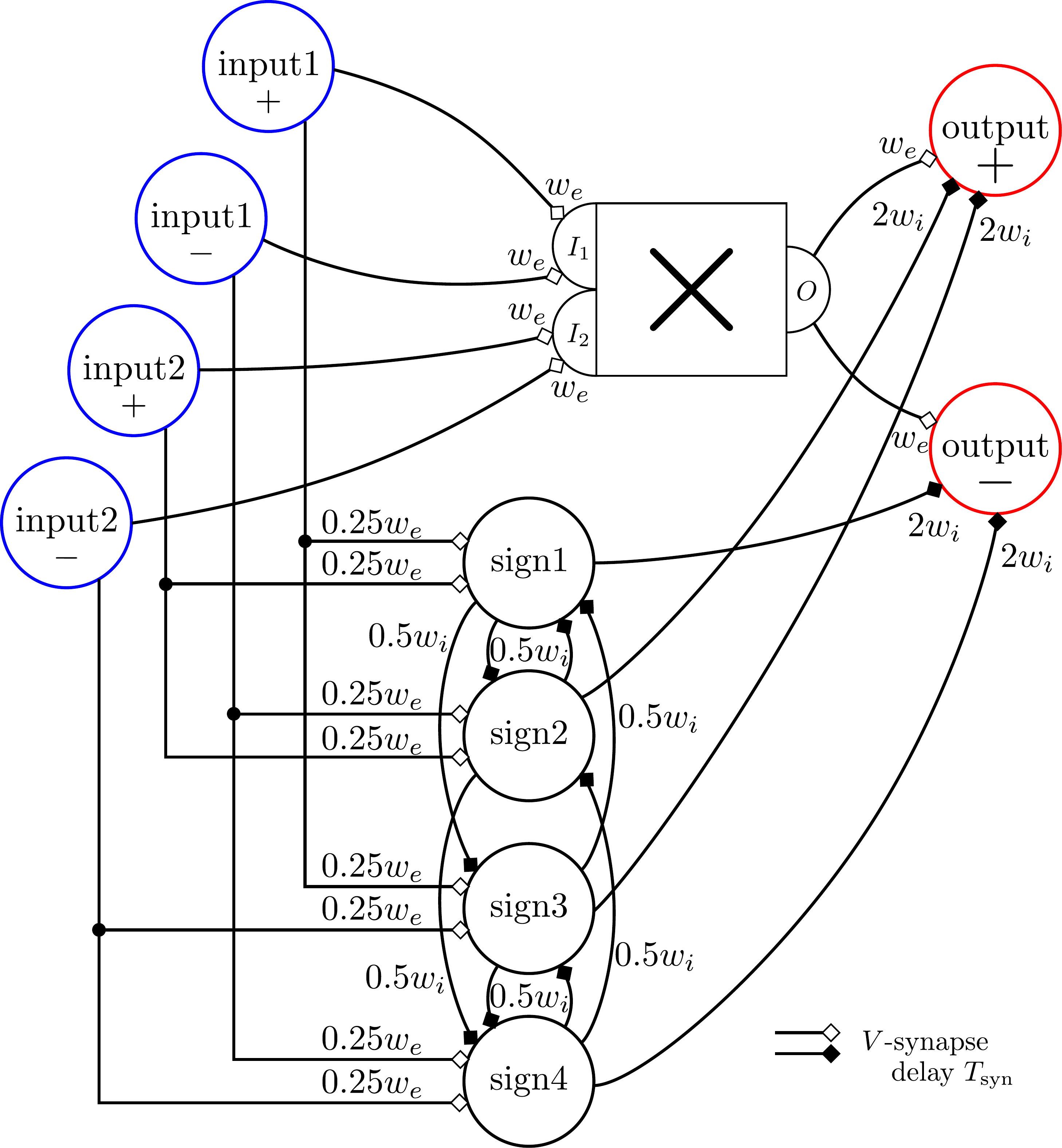}
  \caption{\textbf{Signed Multiplier:} this network computes the result and the sign of the multiplication
           of its two inputs. A Multiplier network is used to compute the absolute value of the result.
           This value is then directed to the correct output (\emph{output~+} or \emph{output~-}) by
           a small truth-table implemented in neurons \emph{sign1}, \emph{sign2}, \emph{sign3} and
           \emph{sign4} which are determining the sign of the output from the signs of the two inputs.
           Blue, red and black neurons are, respectively, input, output and internal neurons.}
  \label{fig:net_signed_multiplier}
\end{figure}

Fig.~\ref{fig:net_signed_multiplier} presents a Signed Multiplier network. It computes the product of two inputs independently of their sign. In parallel, a set of
neurons: \emph{sign1}, \emph{sign2}, \emph{sign3} and \emph{sign4} are used as a truth table to
determine the sign of the output from the sign of the inputs. When the output sign is known, the wrong
output pathways (either \emph{output~+} or \emph{output~-}) is inhibited to direct the output of
the Multiplier network to the right output neuron.
The different \emph{sign} neurons implement a truth table with the excitatory connections they receive
from the input neurons. Input connections and weights are designed such that only one \emph{sign}
neurons spikes when an input is fed into the circuit. This ``winning'' neuron can then be associated to
an output sign. Lateral inhibition between the \emph{sign} neurons is present to suppress residual
activations by the input of non-winning \emph{sign} neurons (this allows all the \emph{sign} neurons to
go back to their reset state once the output sign is computed).

\subsection{Differential equations}

\paragraph{Integrator}

\begin{figure}[h!]
  \centering
  \includegraphics[width=0.7\columnwidth]{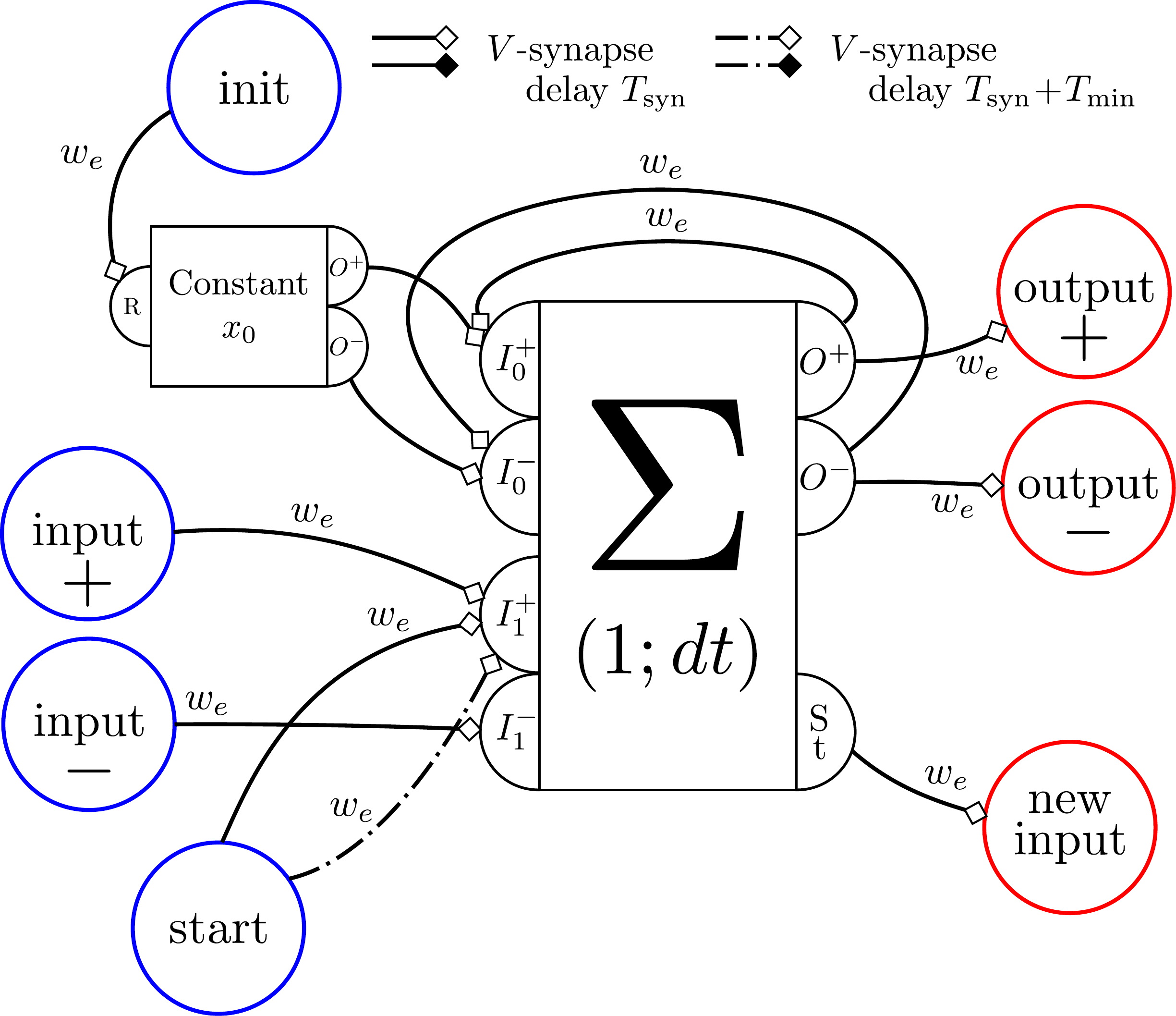}
  \caption{\textbf{Integrator:} this network integrates an input over time. The \emph{init} and
           \emph{start} neurons are used to initialize the inner state of the integrator.
           The network is then driven by its input. Everytime an input is received, it is integrated
           and added to the latest value of the output. When a new value is output, the
           \emph{new\_input} neuron spikes, requesting for a new input.
           Blue, red and black neurons are, respectively, input, output and internal neurons.}
  \label{fig:net_integrator}
\end{figure}

Fig.~\ref{fig:net_integrator} presents a Integrator network that allows to reconstruct a signal
from its derivative fed into its input. It is using a multiplier and an accumulation network with a Linear Combination network. The output of this accumulator network is looped into its first input
with a unit gain. The input, composed of \emph{input~+} and \emph{input~-}, is fed into this accumulator
with a gain $dt$ corresponding to the chosen integration timestep. Each time an output is produced on
\emph{output~+} and \emph{output~-}, the indicator neuron \emph{new\_input} is triggered to notify that the integrator is ready to receive its next input. 
This system is thus driven by its input: every time an input is provided, the corresponding output is computed.
Two auxiliary input neurons are also provided. The \emph{init} neuron loads the integrator with its
initial value. This allows the internal state of the integrator (through the Linear Combination network)
to be set to an initialization value. The \emph{start} neuron
feeds a zero into the input of the integrator, thus computing its first output. 
%It can be required to start the integrator loop in a simulation for instance.

%% {\color{blue} Note:
%%   now that we have multiplication, we could use an accumulator to integrate a constant input
%%   over time and get into a neuron the time between two inputs are fed into the integrator. Theoretically,
%%   I think it would be relatively easy to build a version of the integrator which is actually integrating
%%   an input over time. The latest input fed into the circuit could be stored in a memory and recalled
%%   everytime an integration step occurs (on an overflow of the time counting circuit for instance, which
%%   would time-drive the integrator). Then we could imagine updating the integration circuit every time
%%   a new input is fed or when a certain time has passed with no input. We would then get an integration
%%   circuit with a dynamic $dt$ which depends on the update frequency of its input... if this input comes
%%   from an event-based sensor, we would get an integrator which automatically adapt its timestep to the
%%   dynamics of the input.}

\paragraph{System design}

All the networks presented in this section can be assembled to achieve more complex computational tasks.
Multiplier and Linear Combination networks can be associated to compute arbitrary functions on some
state variables. Integrator networks can then be used to solve systems of differential equations.
Examples of such network will be demonstrated in the next section.

%% file: results.tex
\section{Results}
\label{sec:results}

%% {\color{blue} I did not put details about validation of every circuit... Firstly it would be strange
%%   to have everything again after the \emph{Methods} section, secondly it may be more noise than results
%%   in the paper (it would consist in showing rasters of spikes hard to read and interpret)...}

We implement in this section different computational tasks. We start by implementing linear differential equations with a first order and
a second order system. In a second stage, we implement a more complex set of non-linear differential equations
from Edward Lorenz.

\subsection{Linear differential equations}

\paragraph{First order system}

\begin{figure}[h!]
  \centering
  \includegraphics[width=\columnwidth]{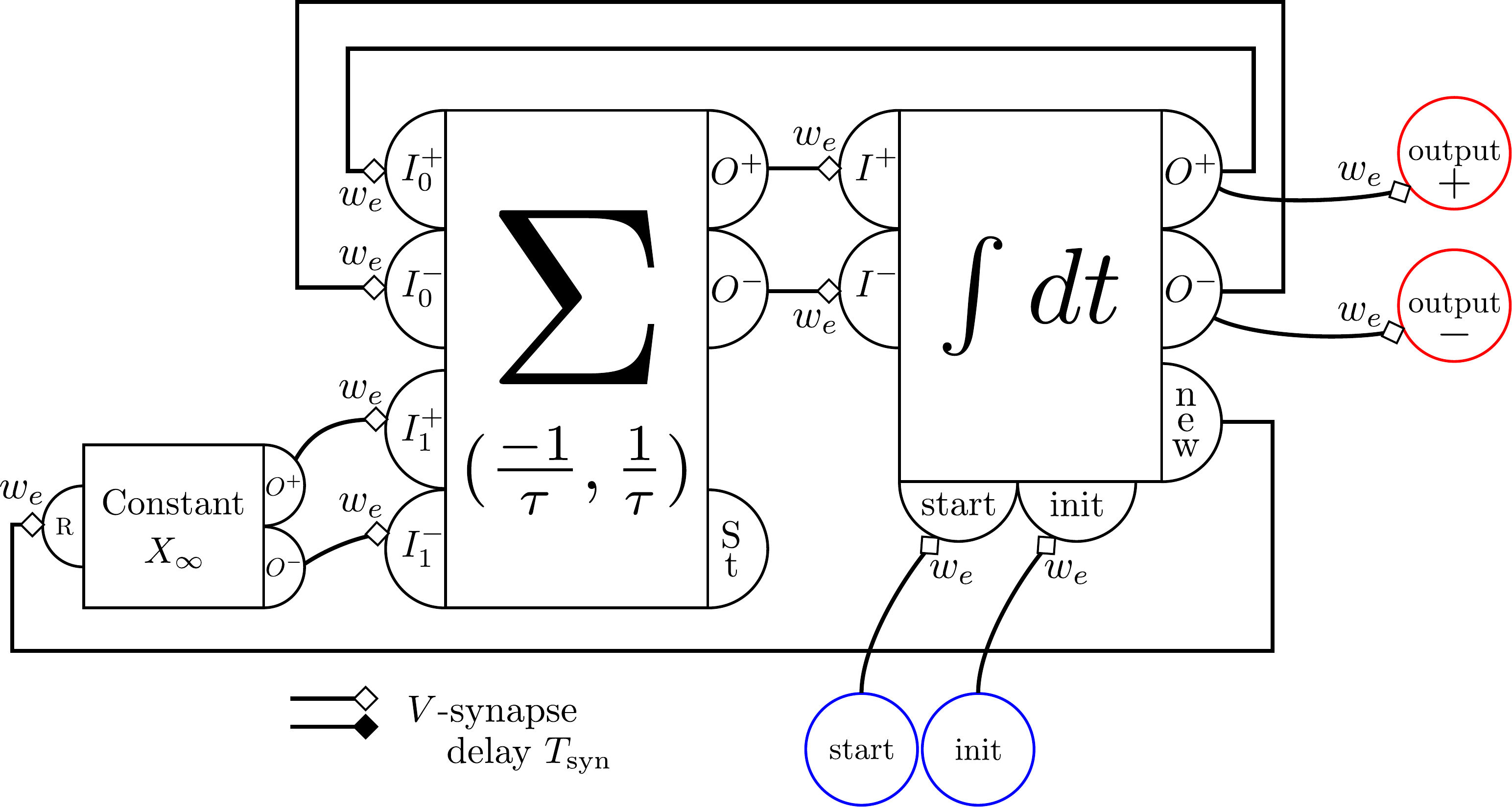}
  \caption{Neural network implementing a first order differential equation. It is composed of
           a Constant network providing the input, a Linear Combination network computing
           the derivative of $X$ and an Integrator network computing $X$ from $\mathrm{d}X/dt.$}
  \label{fig:first_order_network}
\end{figure}

\begin{figure}[h!]
  \centering
  \subfigure[Different values of {$\tau$}]%
            {\includegraphics[width=0.49\columnwidth]{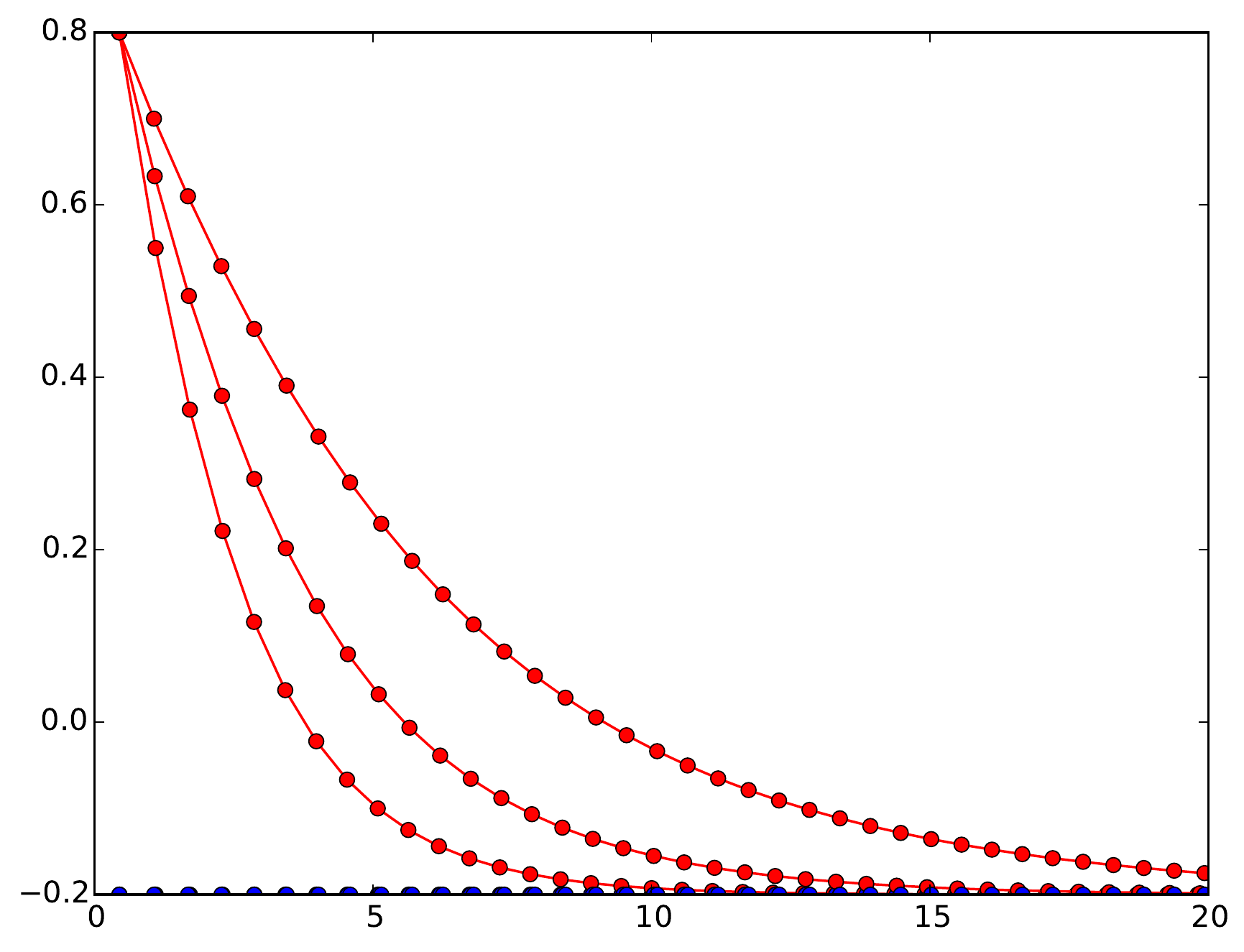}\label{fig:first_order_tau}}
  \subfigure[Different values of {$X_\infty$}]%
            {\includegraphics[width=0.49\columnwidth]{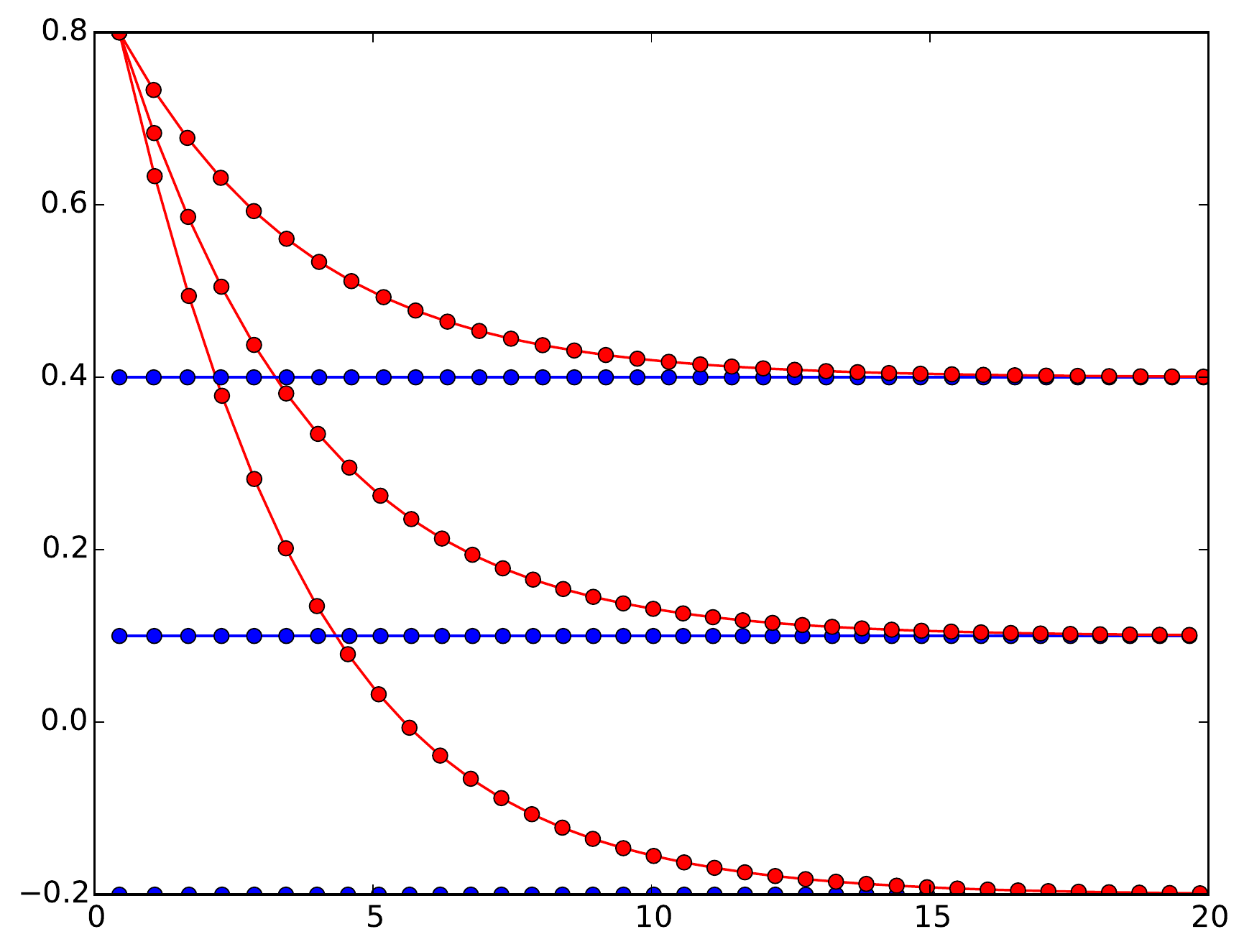}\label{fig:first_order_inf}}
  \caption{Neural implementation of a first order filter. Blue curves show the input values to the
           filter, red curves show the output values of the filter for different parameters.
           (Horizontal axis is time, every circle correspond to an actual output encoded by spikes
           of the network.)}
  \label{fig:first_order_res}
\end{figure}

We first implement a first order differential system. This system solves the following
equation:
\begin{equation}
  \tau.\frac{\mathrm{d}X}{dt} + X(t) = X_\infty
\end{equation}
We implement this network as shown in Fig.~\ref{fig:first_order_network} using 3 of the networks
described in the previous section:
\begin{itemize}
  \item a Constant network is providing the input $X_\infty$ to the system,
  \item a Linear Combination network is computing $\mathrm{d}X/dt$,
  \item an Integrator network is computing $X$ from its derivative.
\end{itemize}
The \emph{init} and \emph{start} neurons enable to initialize and start the integration process.
\emph{init} has to be triggered before the integration process can take place to load the initial
value of the Integrator network. \emph{start} has to be triggered to output the first value from
the Integrator network. When an output is provided by the Integrator network, the Constant network
is activated using the \emph{new\_input} neuron of the Integrator. Hence feeding two values into
the Linear Combination computing the new derivative of the output. This derivative is then integrated
by the Integrator to obtain a new output.
With the implementations presented in the previous section, this network requires 118 neurons.
Results of its simulation with different set of parameters for $\tau$ and $X_\infty$ and for
$dt=0.5$ are presented Fig.~\ref{fig:first_order_res}.
%We can observe that the system is behaving as expected.

\paragraph{Second order system}

\begin{figure}[h!]
  \centering
  \includegraphics[width=\columnwidth]{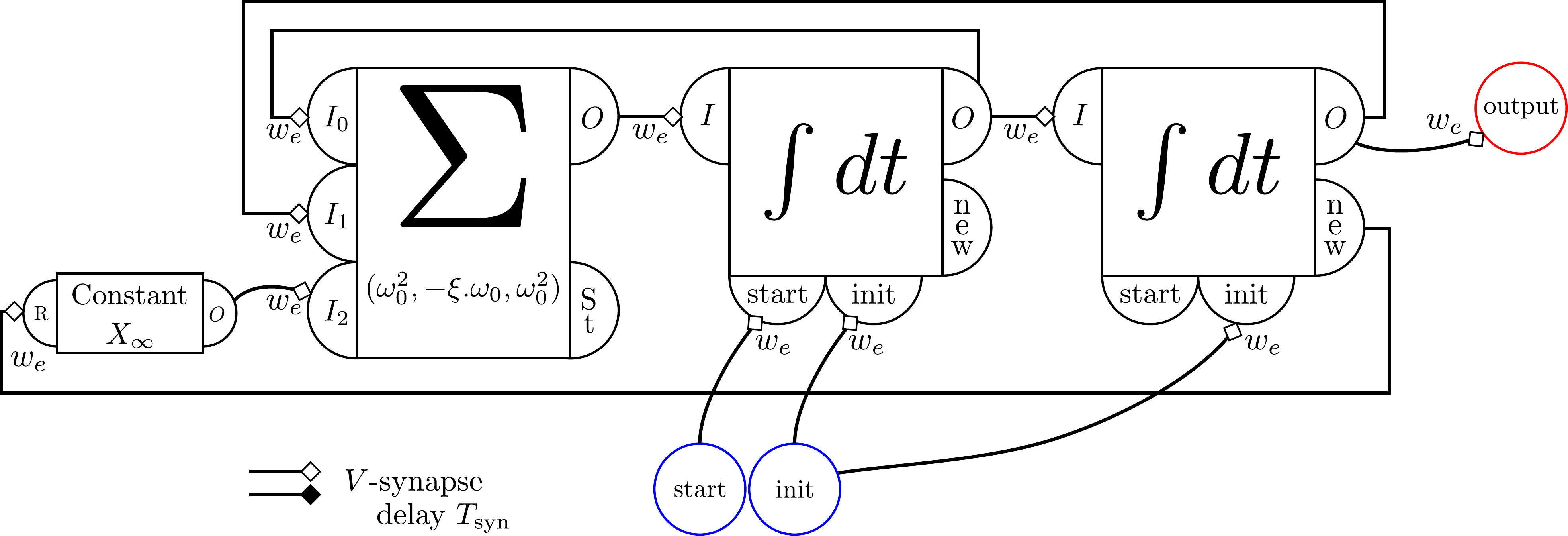}
  \caption{Neural network implementing a second order differential equation. It is composed of
           a Constant network providing the input, a Linear Combination network computing
           the second derivative of $X$ and two Integrator networks computing $\mathrm{d}X/dt$ and
           $X$ from $\mathrm{d}^2X/dt^2.$ (Signs of the different signals have been omitted to
           increase readability.)}
  \label{fig:second_order_network}
\end{figure}

\begin{figure}[h!]
  \centering
  \subfigure[Different values of {$\omega_0$}]%
            {\includegraphics[width=0.49\columnwidth]{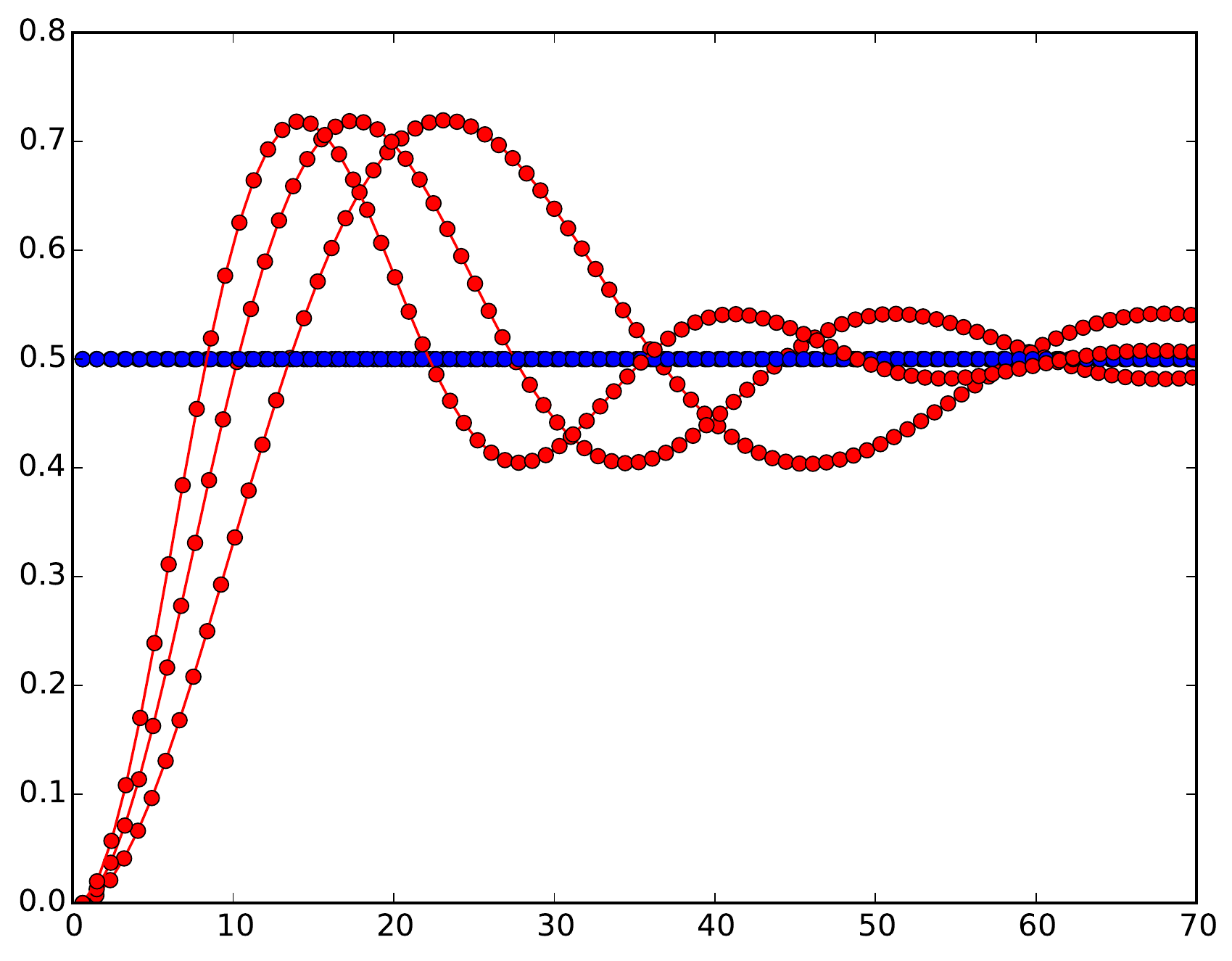}\label{fig:second_order_omega}}
  \subfigure[Different values of {$\xi$}]%
            {\includegraphics[width=0.49\columnwidth]{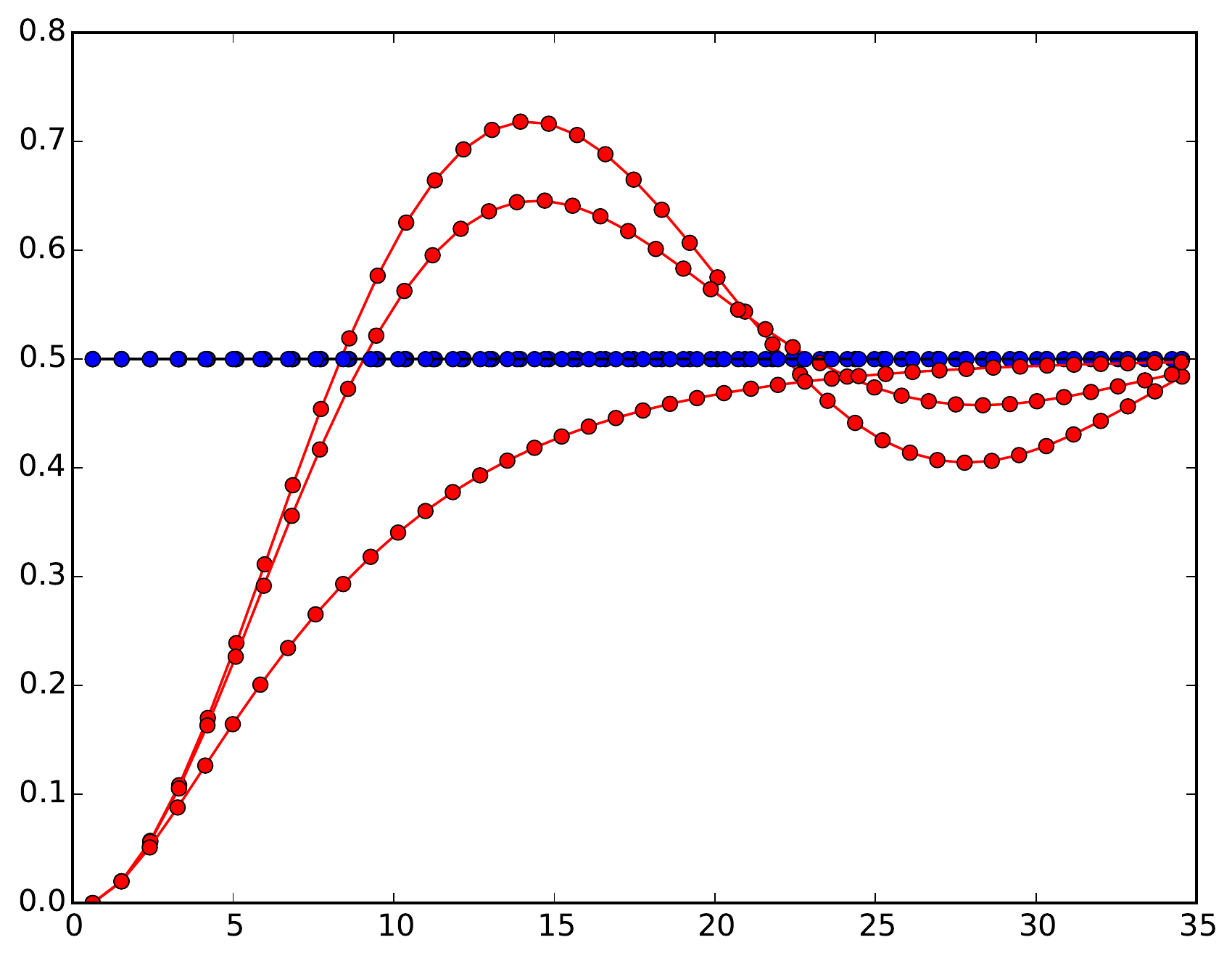}\label{fig:second_order_ksi}}
  \caption{Neural implementation of a second order filter. Blue curves show the input values to the
           filter, red curves show the output values of the filter for different parameters.
           (Horizontal axis is time, every circle correspond to an actual output encoded by spikes
           of the network.)}
  \label{fig:second_order_res}
\end{figure}

To add complexity, we now implement a second order differential system. This system solves the following
equation:
\begin{equation}
  \frac{1}{\omega_0^2}.\frac{\mathrm{d}^2X}{dt^2}
  + \frac{\xi}{\omega_0}.\frac{\mathrm{d}X}{dt}  + X(t) = X_\infty
\end{equation}
We implement this network as shown in Fig.~\ref{fig:second_order_network} using 4 of the networks
described in the previous section:
\begin{itemize}
  \item a Constant network is providing the input $X_\infty$ to the system,
  \item a Linear Combination network is computing $\frac{\mathrm{d}^2X}{dt^2}$,
  \item a first Integrator network is computing $\mathrm{d}X/dt$ from the second derivative of $X$,
  \item a second Integrator network is computing $X$ from its first derivative.
\end{itemize}
With the implementations presented in the previous section, this network requires 187 neurons.
Results of its simulation with different set of parameters for $\xi$ and $\omega_0$ and for
$dt=0.2$ are presented Fig.~\ref{fig:second_order_res}.
%We can observe that the system is behaving as expected.

\subsection{Lorenz attractor}

\begin{figure}[h!]
  \centering
  \includegraphics[width=\columnwidth]{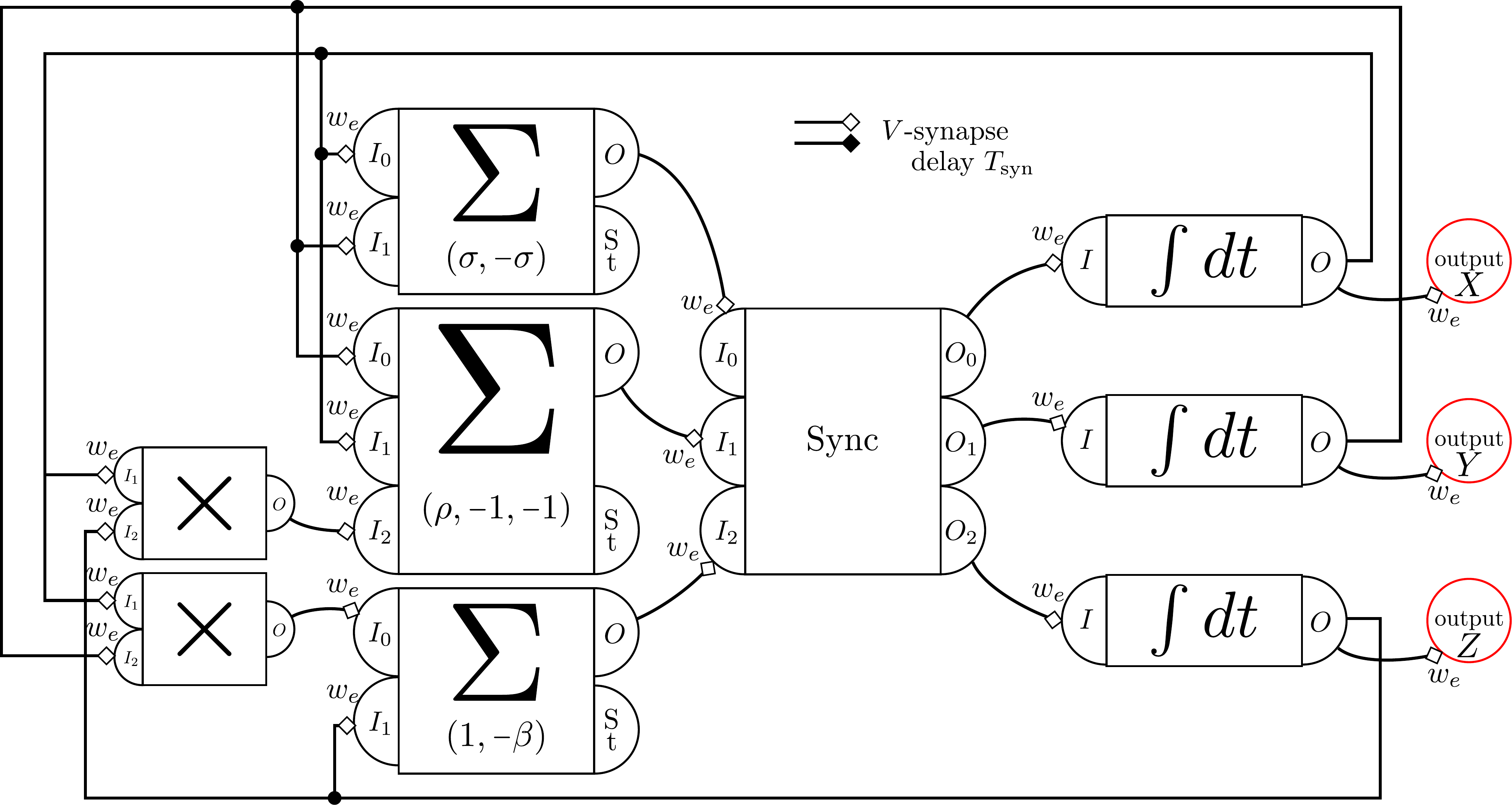}
  \caption{Neural network implementing Edward Lorenz's non-linear differential equation system.
           (Signs of the different signals have been omitted to increase readability.)}
  \label{fig:lorenz_network}
\end{figure}

\begin{figure}[h!]
  \centering
  \includegraphics[width=\columnwidth]{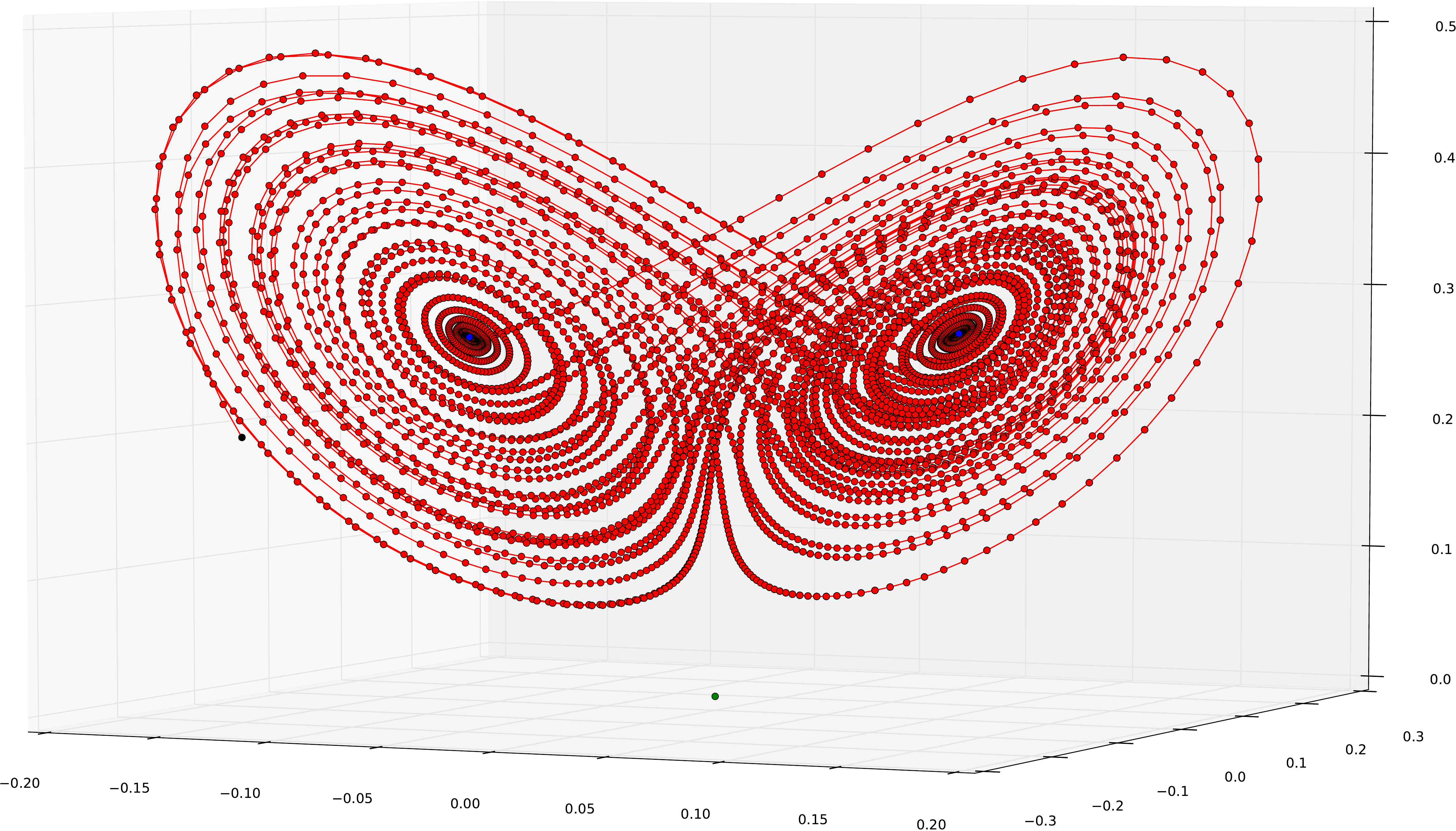}
  \caption{Neural implementation of a set of non-linear differential equations. These equations
           converge to the Lorenz attractor. The plot shows the evolution of the system in the phase
           space, every axis being one of the state variables $X$, $Y$, $Z.$ Fixed points of
           the system are shown in blue, the origin point is shown in green. Every circle
           corresponds to an actual output encoded by spikes of the network.}
  \label{fig:lorenz_res}
\end{figure}

%To show the capabilities of the implementation proposed in this work to solve complex tasks,
We now implement the set of non-linear differential equation proposed by \cite{lorenz1963}:
\begin{eqnarray}
  \frac{\mathrm{d}X}{dt} &=& \sigma(Y(t)-X(t))\\
  \frac{\mathrm{d}Y}{dt} &=& \rho X(t) - Y(t) - X(t).Z(t)\\
  \frac{\mathrm{d}Z}{dt} &=& X(t).Y(t) - \beta Z(t)
\end{eqnarray}
using $\sigma=10,\,\beta=8/3$ and $\rho=28$ to ensure chaotic behavior of the system. We also use
a variable substitution to obtain state variables $X,\,Y$ and $Z$ evolving in $[0,1]$ so that they
can be represented by our framework. The initial state of the system is set to
$X=-0.15,$ $Y=-0.20$ and $Z=0.20.$ We use an integral step of $dt=0.01.$

We implement this network as shown Fig.~\ref{fig:lorenz_network} by using 9 of the networks
described in the previous section:
\begin{itemize}
  \item 2 Signed Multiplier networks compute the non-linarities contained in $Y$ and $Z$'s derivatives,
  \item 3 Linear Combination networks compute the derivatives of $X$, $Y$ and $Z$,
  \item a Signed Synchronizer network allows to wait for the three derivatives to be computed before
        evolving the system's state,
  \item 3 Integrator networks compute the new state from the derivatives of $X$, $Y$ and $Z$.
\end{itemize}
With the implementations presented in the previous section, this network requires 549 neurons.
Results of its simulation are presented Fig.~\ref{fig:lorenz_res}.
We can observe that the system is behaving as expected, following the strange attractor described
by Lorenz.

%% file: discussion.tex
\section{Discussion}
\label{fig:discussion}

In this work, we choose a linear encoding function to map values into inter spikes. In this case, it
results in a direct trade-off between the time needed to represent a value and the time precision
of the system. A finer time precision leads to a larger number of possible different values
in a given maximum representation time.
In the paper we chose to set time scales that are compatible with neuroscience evidence.
However, current hardware allows much faster time scales up to nanoseconds.
In that case not only can we obtain higher precision but also faster computation times.\\

%Future work will explore the advantages and drawbacks of using logarithmic
%representations of values. This would allow dynamic precision of the encoded values depending on
%their absolute value (larger values have less precision).
%This solution gives more flexibility to the scaling of input values but is changing the way we have
%to tackle the problem of operations. While simplifying multiplication of signals, linear operations
%become harder.

The number of neurons used in the current implementation of the examples given in this paper can be reduced in size. They have been designed to ease comprehension and to be easy to plug into each other. In almost all the networks, the first and second encoding
spikes for the output are generated by different neurons of the network. All the networks have a pair of \emph{first} and \emph{last} neurons which task is
to separate the two incoming spikes. By directly routing these two spikes independently, the
whole implementation would be much more efficient and would require much less neurons.
Less neurons would also imply less spikes and therefore less energy requirements and less latency in signal propagation.\\

For feedforward architectures, the different layers of computation could also be pipelined. This means
that if a task is composed of a series of operations which can be considered as layers, the full
operation would require data to go through all the layers, with all layers being active only a fraction
of the time. But at every point in time, one layer could also be computing its operation on a different
input, such that all of the layers are always active. In the end, one can get one output per pipeline
period, even if the whole computation takes several pipeline
period to be completed (as many pipeline period as there are steps in the computation, the pipeline
period being the longest time needed by one stage to output its result).
This means better throughput for feedforward computation, but imposes a minimum delay on feedback
(because the result to be fed back is not available before a certain number of pipeline periods).\\

The developed framework can be used to compute any algorithm, it is intersting to notice that it is naturally compatible with all type of time oriented AER (Address Event Representation \citep{Boahen2000}) data and all kind of AER sensors \citep{Delbruck2010}.
The use of interspike makes the framework particularly adapted to process luminance time encoded events data from the neuromorphic camera ATIS ("Asynchronous Time-based Image Sensor") \citep{Posch2011,Posch2008}. Thus every event-based machine vision algorithm developed so far could be systematically implemented on neural boards.\\

A question outside the scope of this paper relates to the architecture of the platform that should implement this computation paradigm.
Several solutions could be possible. A pure analog chip could be used. Analog design is known to be a difficult endeavour, moreover we would need to robustify the framework to overcome mismatch. Several options are possible, the most straightforward is to use regularization techniques such as the one introduced in \citep{DBLP:journals/tnn/BenosmanCLIB14} or simply using calibration techniques to provide a precisely timed output from analog chip similar to what has been introduced in \citep{Sheik_etal12,Sheik_etal11}.\\
A mixed-signal integrated circuit mixing both analog circuits and digital circuits could also be considered. Exponential decays could be generated using analog circuits while the remaining operations could use digital circuits. Finally the last solution could be a pure digital architecture preserving the principles introduced in the paper such as inter spike encoding, local computations that allow to overcome the Von Neumann bottelneck. In this case, we could use local binary computation units to perform conventional computation rather than using membrane potentials, synapses, delay,\ldots. Everything being local to the units, the architecture would still be efficient. This by far is the less elegant solution, but it could be a rapid and an intermediate step toward a full true neuromorphic implementation.

%% file: conclusion.tex
\section*{Conclusion}
\label{fig:conclusion}

This work introduced a new clockless framework to build a multipurpose neuromorphic computer.
Instead of representing values as a set of bits in a register or a part of some central memory, values are coded in the precise timing
between events happening in the system. This dataflow framework we called STICK (Spike Time Interval Computational Kernel) offers a new
method to design computing platforms where memory and computation are intertwined. By removing
the numerous accesses to a central memory, inherent to standard computers, we free ourselves from
the \emph{Von Neumann} bottleneck.
The systems scales naturally, the more neurons are available the more computation can be performed.
Building a large machine consisting of several small elementary computational units allows to build natively massively parallel machines which rely on neuromorphic engineering principles to reduce their power consumption: energy is needed to produce events
which only happens when information is produced.

Moreover, the STICK framework offers a way to design large neural networks that accomplish complex tasks
by design of their architecture instead of only exploiting the randomness of a given connectivity topology.
It also defines all the weights necessary to obtain a functional solution without the need for a costly
(in terms of energy, time or computational power) learning process. A hardware fabric consisting of
a large number of the proposed computational units with dynamic connection could then be used and adapted
to successively solve different tasks.

%% file: appendix.tex
\section*{Appendix}

This appendix presents the detailed proofs of all the networks described in Section~\ref{sec:methods}

\section{Storing data: analog memories}
\subsection{Inverting Memory}
\label{sec:app:inverting_memory}

\begin{figure}[h!]
  \centering
  \includegraphics[width=0.6\columnwidth]{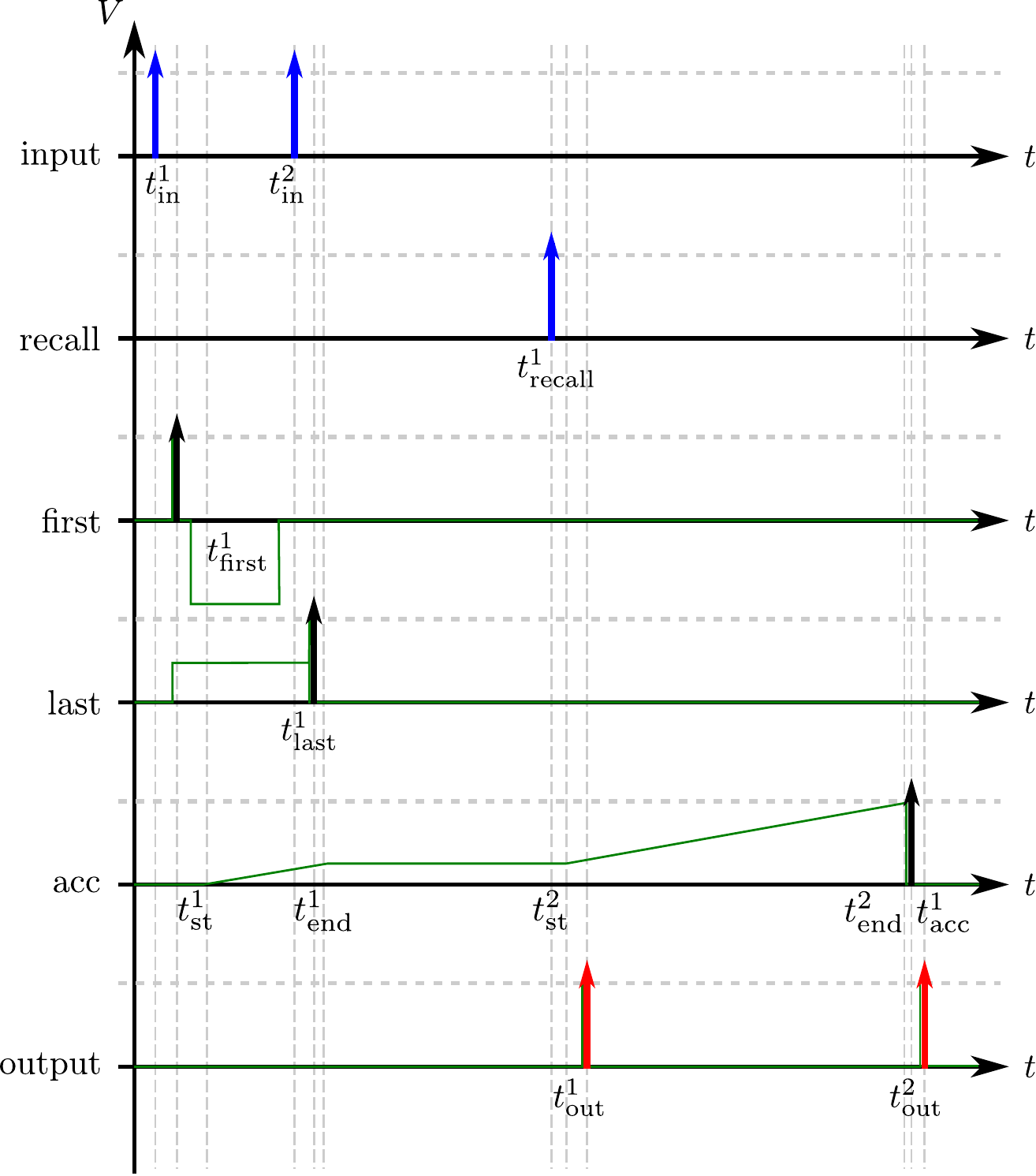}
  \caption{\textbf{Inverting Memory:} chronogram of the network for an input at times
           $t_\mathrm{in}^1$ and $t_\mathrm{in}^2$ and a recall at time $t_\mathrm{recall}^1.$
           (Input spikes are drawn in blue, output spikes in red. Green plots show the membrane
           potential of interesting neurons, recall of Fig.~\ref{fig:chrono_inverting_memory})}
  \label{fig:app:chrono_inverting_memory}
\end{figure}

The Inverting Memory network (see Fig.~\ref{fig:net_inverting_memory}) receives 2 spikes on the
\emph{input} neuron at times $t_\mathrm{in}^1$
and $t_\mathrm{in}^2$ such that $\Delta T_{in}=t_\mathrm{in}^2-t_\mathrm{in}^1$ encodes its input value.
When \emph{input} spikes at $t_\mathrm{in}^1$, synaptic connections are activated towards neurons
\emph{first} and \emph{last}. Because of the synaptic delays and weights, \emph{last}'s membrane potential
is set to $V_t/2$ and \emph{first} spikes at time
$t_\mathrm{first}^1=t_\mathrm{in}^1+T_\mathrm{syn}+T_\mathrm{neu}$, $T_\mathrm{neu}$ being the time needed by
\emph{first} to produce a spike.
When \emph{first} spikes, an inhibitory connection to itself sets its potential to $-V_t$ while a second
connection triggers the integration of neuron \emph{acc} after a delay $T_\mathrm{syn}+T_\mathrm{min}$ with
weight $w_\mathrm{acc}$. We then have:
\begin{eqnarray}
  t_\mathrm{st}^1 &=& t_\mathrm{first}^1+T_\mathrm{syn}+T_\mathrm{min} \\
                &=& t_\mathrm{in}^1+2.T_\mathrm{syn}+T_\mathrm{neu}+T_\mathrm{min}.
\end{eqnarray}
When \emph{input} spikes for the second time at time $t_\mathrm{in}^2$, \emph{first}'s membrane potential
gets back to its reset value while \emph{last} reaches its threshold. This produces a spike from
neuron \emph{last} at time $t_\mathrm{last}^1=t_\mathrm{in}^2+T_\mathrm{syn}+T_\mathrm{neu}.$ The connection
to \emph{acc} with delay $T_\mathrm{syn}$ and weight $-w_\mathrm{acc}$ stops \emph{acc}'s integration at
time:
\begin{eqnarray}
  t_\mathrm{end}^1 &=& t_\mathrm{last}^1+T_\mathrm{syn} \\
                 &=& t_\mathrm{in}^2+2.T_\mathrm{syn}+T_\mathrm{neu}.
\end{eqnarray}
During this integration window we had, for neuron \emph{acc}, $g_e=w_\mathrm{acc}$ such that, the membrane
potential of \emph{acc} at time $t_\mathrm{end}^1$ is:
\begin{eqnarray}
  V_{sto} &=& \frac{w_\mathrm{acc}}{\tau_m}.(t_\mathrm{end}^1-t_\mathrm{st}^1) \\
         &=& \frac{w_\mathrm{acc}}{\tau_m}.(t_\mathrm{in}^2-(t_\mathrm{in}^1+T_\mathrm{min})) \\
         &=& \frac{w_\mathrm{acc}}{\tau_m}.(\Delta T_{in} - T_\mathrm{min}) \label{eq:imem_vsto}
\end{eqnarray}

When the \emph{recall} neuron receives an input at time $t_\mathrm{recall}^1$, \emph{acc}'s integration
starts again at time $t_\mathrm{st}^2 = t_\mathrm{recall}^1+T_\mathrm{syn}.$ At the same time, its second
connection triggers a spike of the \emph{output} neuron at time:
\begin{eqnarray}
  t_\mathrm{out}^1 &=& t_\mathrm{recall}^1+(2.T_\mathrm{syn}+T_\mathrm{neu})+T_\mathrm{neu} \\
                 &=& t_\mathrm{recall}^1+2.T_\mathrm{syn}+2.T_\mathrm{neu}
\end{eqnarray}
The integration stops again when \emph{acc} reaches its threshold at time $t_\mathrm{end}^2$, giving
the following equation:
\begin{eqnarray}
  V_t &=& \frac{w_\mathrm{acc}}{\tau_m}.(t_\mathrm{end}^2-t_\mathrm{st}^2) + V_{sto} \\
  t_\mathrm{end}^2 &=& (t_\mathrm{recall}^1+T_\mathrm{syn}) - (\Delta T_{in} - T_\mathrm{min}) + V_t.\frac{\tau_m}{w_\mathrm{acc}},
\end{eqnarray}
By definition of $w_\mathrm{acc}$, we have $V_t.\tau_m/w_\mathrm{acc}=T_\mathrm{max}, so:$
\begin{equation}
  t_\mathrm{end}^2 = t_\mathrm{recall}^1 + T_\mathrm{syn} + T_\mathrm{max} - (\Delta T_{in} - T_\mathrm{min}),
\end{equation}
Because \emph{acc1} then needs the time $T_\mathrm{neu}$ to produce a spike, we get:
\begin{equation}
  t_\mathrm{acc}^1 = t_\mathrm{end}^2+T_\mathrm{neu}
\end{equation}
We thus get the second spike of \emph{output} at time:
\begin{eqnarray}
  t_\mathrm{out}^2 &=& t_\mathrm{acc1}^1 + T_\mathrm{syn} + T_\mathrm{neu} \\
  t_\mathrm{out}^2 &=& t_\mathrm{recall}^1 + 2.T_\mathrm{syn} + 2.T_\mathrm{neu} + T_\mathrm{max} - (\Delta T_{in} - T_\mathrm{min})
\end{eqnarray}
such that:
\begin{eqnarray}
  \Delta T_{out} &=& t_\mathrm{out}^2 - t_\mathrm{out}^1 \\
                &=& T_\mathrm{max} - (\Delta T_{in} - T_\mathrm{min}).
\end{eqnarray}

\subsection{Memory}
\label{sec:app:memory}

\begin{figure}[h!]
  \centering
  \includegraphics[width=0.6\columnwidth]{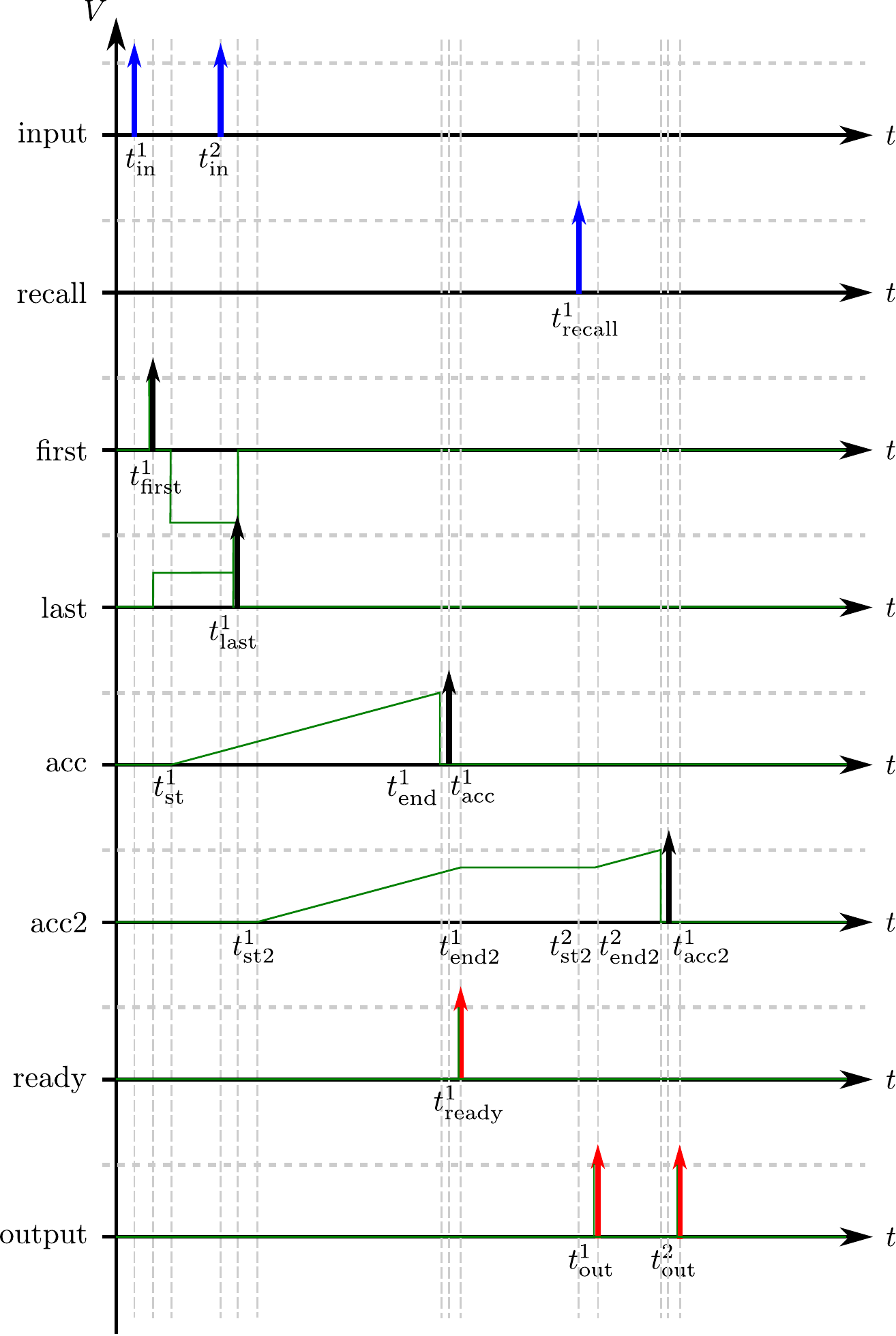}
  \caption{\textbf{Memory:} chronogram of the network for an input at times
           $t_\mathrm{in}^1$ and $t_\mathrm{in}^2$ and a recall at time $t_\mathrm{recall}^1.$
           (Input spikes are drawn in blue, output spikes in red. Green plots show the membrane
           potential of interesting neuron.)}
  \label{fig:app:chrono_memory}
\end{figure}

The Memory network (see Fig.~\ref{fig:net_memory}) receives 2 spikes on the \emph{input} neuron at
times $t_\mathrm{in}^1$ and $t_\mathrm{in}^2$ such that $\Delta T_{in}=t_\mathrm{in}^2-t_\mathrm{in}^1$
encodes its input value. With the reasoning used in the previous subsection, we get:
\begin{eqnarray}
  t_\mathrm{first}^1 &=& t_\mathrm{in}^1+T_\mathrm{syn}+T_\mathrm{neu} \\
  t_\mathrm{last}^1 &=& t_\mathrm{in}^2+T_\mathrm{syn}+T_\mathrm{neu}.
\end{eqnarray}
When \emph{first} spikes, it triggers integration in \emph{acc}'s membrane potential. Because
of the synaptic delay, we get:
\begin{equation}
  t_\mathrm{st}^1 = t_\mathrm{first}^1 + T_\mathrm{syn}
                = t_\mathrm{in}^1+ 2.T_\mathrm{syn} + T_\mathrm{neu}
\end{equation}
Neuron \emph{acc} continues to integrate its $w_\mathrm{acc}$ input until reaching its threshold.
This gives us the following equation:
\begin{equation}
  V_t = \frac{w_\mathrm{acc}}{\tau_m}.(t_\mathrm{end}^1-t_\mathrm{st}^1) \\
\end{equation}
by definition of $w_\mathrm{acc},$ we thus get:
\begin{eqnarray}
  t_\mathrm{end}^1 &=& t_\mathrm{st}^1 + T_\mathrm{max} \\
                 &=& t_\mathrm{in}^1 + T_\mathrm{max} + 2.T_\mathrm{syn} + T_\mathrm{neu} \\
  t_\mathrm{acc}^1 &=& t_\mathrm{end}^1 + T_\mathrm{neu} \\
                 &=& t_\mathrm{in}^1 + T_\mathrm{max} + 2.T_\mathrm{syn} + 2.T_\mathrm{neu} \\
  t_\mathrm{end2}^1 &=& t_\mathrm{acc}^1 + T_\mathrm{syn} \\
                 &=& t_\mathrm{in}^1 + T_\mathrm{max} + 3.T_\mathrm{syn} + 2.T_\mathrm{neu}
\end{eqnarray}
At the same time, integration starts in \emph{acc2}'s membrane potential after \emph{last} spikes,
which corresponds to:
\begin{equation}
  t_\mathrm{st2}^1 = t_\mathrm{last}^1 + T_\mathrm{syn} = t_\mathrm{in}^2+2.T_\mathrm{syn}+T_\mathrm{neu}
\end{equation}
The membrane potential of \emph{acc2} after the end of the first integration phase is thus:
\begin{eqnarray}
  V_{sto} &=& \frac{w_\mathrm{acc}}{\tau_m}.(t_\mathrm{end2}^1-t_\mathrm{st2}^1) \\
         &=& \frac{w_\mathrm{acc}}{\tau_m}.(T_\mathrm{max} + t_\mathrm{in}^1 - t_\mathrm{in}^2 + T_\mathrm{syn} + T_\mathrm{neu}) \\
         &=& \frac{w_\mathrm{acc}}{\tau_m}.(T_\mathrm{max} - \Delta T_{in} + T_\mathrm{syn} + T_\mathrm{neu})
\end{eqnarray}
When the \emph{recall} neuron is triggered, it starts \emph{acc2}'s integration again at time:
\begin{equation}
  t_\mathrm{st2}^2 = t_\mathrm{recall}^1 + T_\mathrm{syn}
\end{equation}
Which then produces the first output spike at time:
\begin{equation}
  t_\mathrm{out}^1 = t_\mathrm{recall}^1+T_\mathrm{syn}+T_\mathrm{neu}.
\end{equation}
This second integration phase of \emph{acc2} finishes when its threshold is reached:
\begin{eqnarray}
  V_t &=& \frac{w_\mathrm{acc}}{\tau_m}.(t_\mathrm{end2}^2-t_\mathrm{st2}^2) + V_{sto} \\
  T_\mathrm{max} &=& t_\mathrm{end2} - t_\mathrm{recall}^1 - T_\mathrm{syn} + T_\mathrm{max} - \Delta T_{in} + T_\mathrm{syn} + T_\mathrm{neu} \\
  t_\mathrm{end}^2 &=& t_\mathrm{recall}^1 + \Delta T_{in} - T_\mathrm{neu}
\end{eqnarray}
We thus get a spike from \emph{acc2} at time:
\begin{equation}
  t_\mathrm{acc2}^1 = t_\mathrm{end2}^2 + T_\mathrm{neu} = t_\mathrm{recall}^1 + \Delta T_{in}
\end{equation}
leading to the second output spike at time:
\begin{equation}
  t_\mathrm{out}^2 = t_\mathrm{acc2}+T_\mathrm{syn}+T_\mathrm{neu} = t_\mathrm{recall}^1 + \Delta T_{in} + T_\mathrm{syn} + T_\mathrm{neu}
\end{equation}
such that:
\begin{eqnarray}
  \Delta T_{out} &=& t_\mathrm{out}^2-t_\mathrm{out}^1 \\
                &=& \Delta T_{in}
\end{eqnarray}

\subsection{Signed Memory}
\label{sec:app:signed_memory}

\begin{figure}[h!]
  \centering
  \includegraphics[width=\columnwidth]{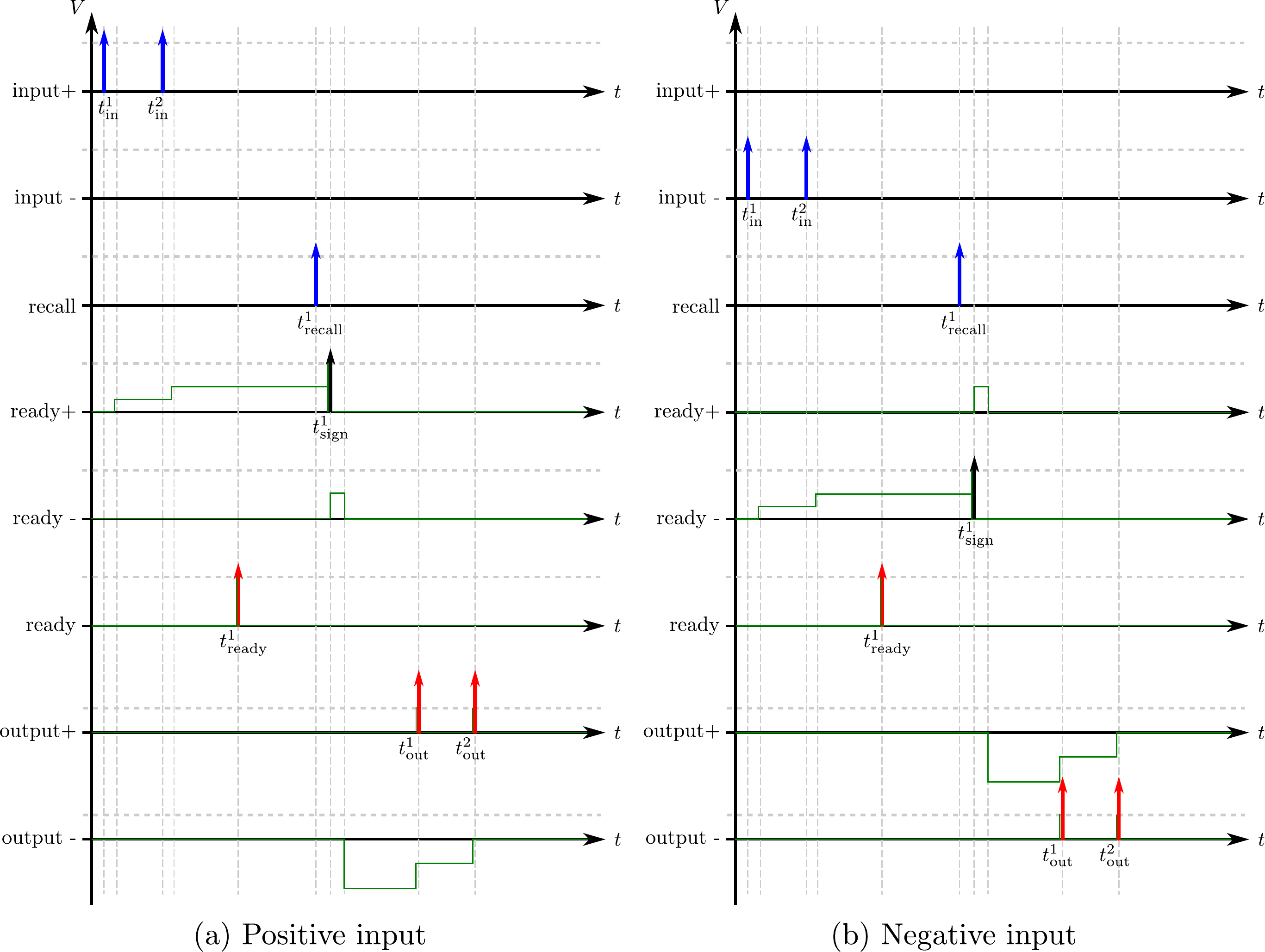}
  \caption{\textbf{Signed Memory:} chronogram of the network for an input at times
           $t_\mathrm{in}^1$ and $t_\mathrm{in}^2$ and a recall at time $t_\mathrm{recall}^1.$
           (Input spikes are drawn in blue, output spikes in red. Green plots show the membrane
           potential of interesting neurons). (a) Depicts the case of a positive input whereas
           (b) depicts the case of a negative input.}
  \label{fig:app:chrono_signed_memory}
\end{figure}

The Signed Memory network (see Fig.~\ref{fig:net_signed_memory}) receives 2 spikes on the
\emph{input} neuron at times $t_\mathrm{in}^1$ and $t_\mathrm{in}^2$ such that
$\Delta T_{in}=t_\mathrm{in}^2-t_\mathrm{in}^1$ encodes its input value and the receiving neuron
encodes the sign of the input (\emph{input~+} for positive inputs, \emph{input~-} for negative
inputs). Let's consider, without loss of generality, the case where the input is positive
(Fig.~\ref{fig:app:chrono_signed_memory}(a)). For each of the 2 input spikes, the
\emph{ready~+} neuron receives a synaptic contribution of $0.25w_e$. When the input has been
completely fed into the network, \emph{ready~+}'s membrane potential is thus resting at a value
of $V_t/2$ while \emph{ready~-}'s is still resting at its reset potential. At the same time,
the input spikes are fed into the central Memory network (see previous subsection) independently
of its sign. When the value is stored in the Memory network, it outputs a spike on its \emph{Rdy}
output, which is propagated to the \emph{ready} neuron.

When the \emph{recall} neuron is triggered, its connections contribute to \emph{ready~+} and
\emph{ready~-}'s membrane potentials with a weight of $0.5w_e$. This leads to a spike of the
\emph{ready~+} neuron at time $t_\mathrm{sign}^1$ while \emph{ready~-}'s membrane potential moves
to $V_t/2.$ The lateral inhibition between the 2 \emph{ready} neurons then sets \emph{ready~-}
back to its reset potential. When \emph{ready~+} spikes, it triggers the recall of the Memory
network and at the same time inhibits the output neuron corresponding to a negative value,
\emph{output~-}, setting its potential to $-2.V_t.$ When the Memory network outputs its stored
value, spikes are transmitted to both output. Because of their respective potential at this
moment, only the positive output \emph{output~+} spikes, while \emph{output~-}'s membrane potential
is set back to its reset potential by the 2 output spikes of the Memory network.

Fig.~\ref{fig:app:chrono_signed_memory}(b) shows the same principle applied to a negative input.
One can notice that the \emph{ready~+} and \emph{ready~-} neurons are implementing a small state machine
routing the spikes produced by the central Memory network to different output neurons depending on
the input neurons.

\subsection{Synchronizer}
\label{sec:app:synchronizer}

\begin{figure}[h!]
  \centering
  \includegraphics[width=0.6\columnwidth]{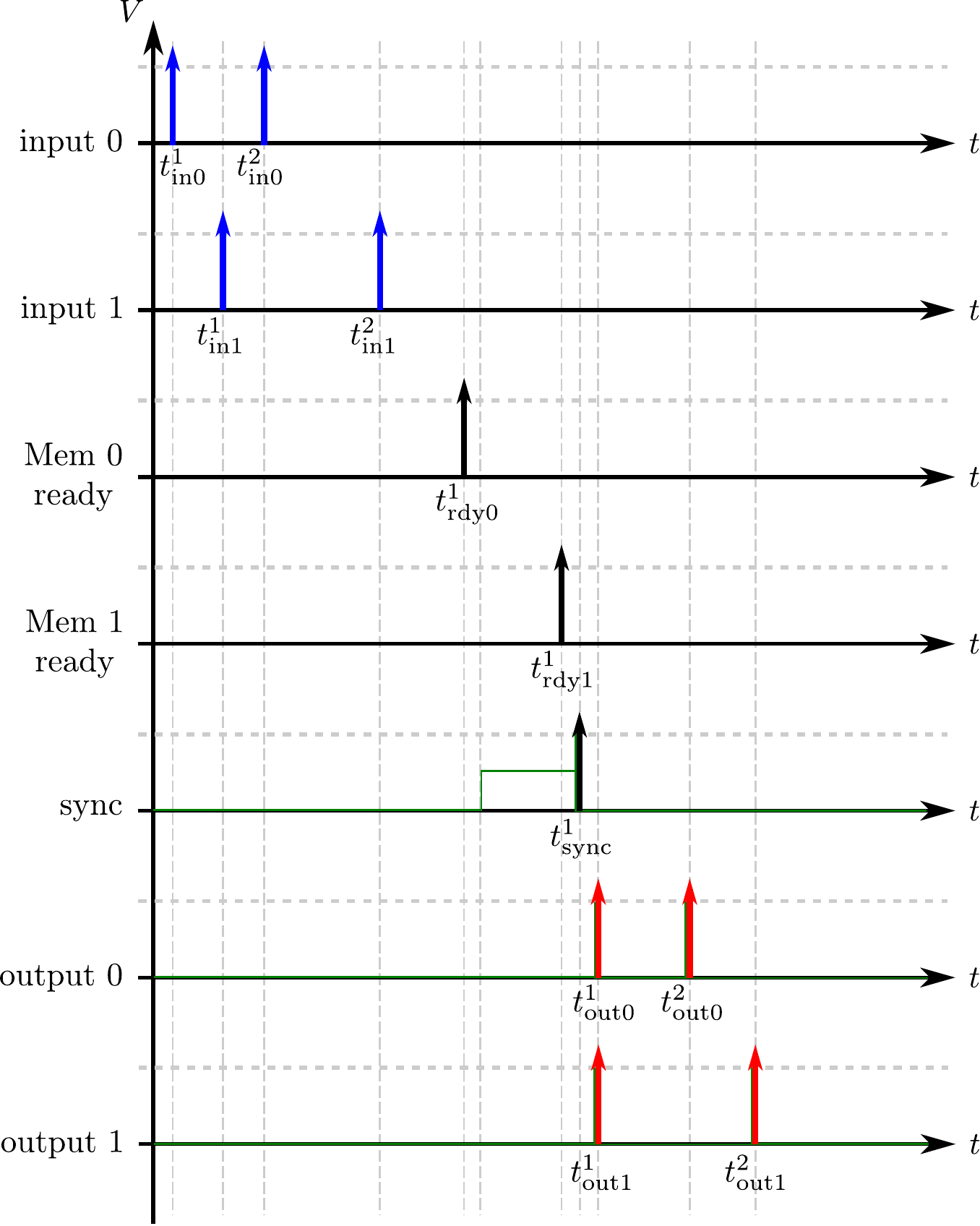}
  \caption{\textbf{Synchronizer:} chronogram of the network for inputs at times
           $t_\mathrm{in0}^1$, $t_\mathrm{in0}^2$ and $t_\mathrm{in1}^1$, $t_\mathrm{in1}^2.$
           (Input spikes are drawn in blue, output spikes in red. Green plots show the membrane
           potential of interesting neuron.)}
  \label{fig:app:chrono_synchronizer}
\end{figure}

The Synchronizer network (see Fig.~\ref{fig:net_synchronizer}) for $N$ inputs uses $N$
Memory networks. Fig.~\ref{fig:app:chrono_synchronizer} presents the chronogram of this network
for $N=2.$ Every time an internal memory has finished storing an input, its \emph{Rdy} output activates
the \emph{sync} neuron with a weight $w_e/N.$ Thus, after $i$ inputs have been presented, \emph{sync}'s
membrane potential $V_\emph{sync}^i$ is:
\begin{equation}
  V_\emph{sync}^i = \frac{i}{N}.V_t.
\end{equation}
The \emph{sync} neuron thus spikes after the $N^\mathrm{th}$ and last memory is ready. It then recalls
the values stored in all Memory networks at the same time, effectively synchronizing the first encoding
spikes of all its outputs.

\section{Relational operations}

\subsection{Minimum}
\label{sec:app:minimum}

\begin{figure}[h!]
  \centering
  \includegraphics[width=0.6\columnwidth]{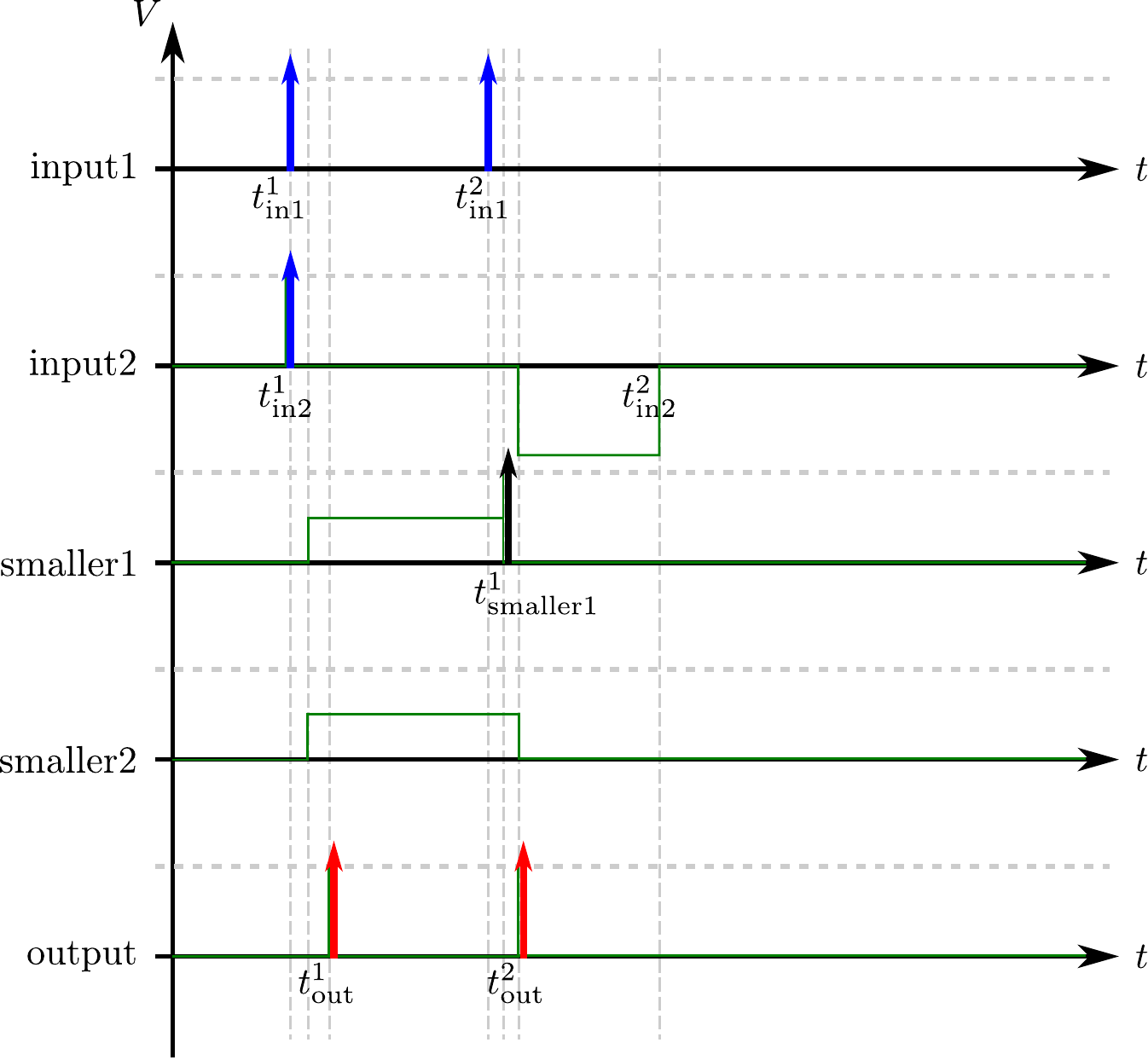}
  \caption{\textbf{Minimum:} chronogram of the network for inputs at times
           $t_\mathrm{in1}^1$, $t_\mathrm{in1}^2$ and $t_\mathrm{in2}^1$, $t_\mathrm{in2}^2.$
           (Input spikes are drawn in blue, output spikes in red. Green plots show the membrane
           potential of interesting neuron.)}
  \label{fig:app:chrono_minimum}
\end{figure}

The Minimum network (see Fig.~\ref{fig:net_minimum}) receives 2 different inputs (a pair of spikes) from each input neurons  \emph{input1} ($t_\mathrm{in1}^1$ and $t_\mathrm{in1}^2$) and
\emph{input2} ($t_\mathrm{in2}^1$ and $t_\mathrm{in2}^2$) such that
$\Delta T_{in1}=t_\mathrm{in1}^2-t_\mathrm{in1}^1$ and $\Delta T_{in2}=t_\mathrm{in2}^2-t_\mathrm{in2}^1$ encode
its 2 inputs. Let us consider, without loss of generality, the case where the first input
(\emph{input1}) is smaller than the second one as shown in Fig.~\ref{fig:app:chrono_minimum}.
Assuming the inputs to be synchronized, we have:
\begin{equation}
  t_\mathrm{in1}^1 = t_\mathrm{in2}^1,
\end{equation}
When \emph{input1} and \emph{input2} are triggered by the first encoding spike of each input,
\emph{smaller1} and \emph{smaller2}'s membrane potentials are set to $V_t/2$ and the \emph{output}
neuron spikes after a delay due to the connection from \emph{input1} and \emph{input2} such that:
\begin{equation}
  t_\mathrm{out1} = t_\mathrm{in1}^1 + 2.T_\mathrm{syn} + 2.T_\mathrm{neu}.
\end{equation}
When the smallest input (in this case \emph{input1}) receives its second encoding spike at time
$t_\mathrm{in1}^2,$ the \emph{smaller1} neuron reaches its threshold and emits a spike at time:
\begin{equation}
  t_\mathrm{smaller1}^1 = t_\mathrm{in1}^2 + T_\mathrm{syn} + T_\mathrm{neu}.
\end{equation}
The spike from \emph{smaller1} inhibits the other input (here \emph{input2}) such that it will not
be triggered by its second encoding spike (as can be seen in the chronogram). It also inhibits the
\emph{smaller2} neuron such that its membrane potential goes back to its reset potential.
The second encoding spike from \emph{input1} and the spike from \emph{smaller1} are, in addition,
enough contribution to trigger a second spike in the \emph{output} neuron at time:
\begin{eqnarray}
  t_\mathrm{out}^2 &=& \min \left\{ t_\mathrm{in1}^2+2.T_\mathrm{syn}+T_\mathrm{neu}, t_\mathrm{smaller1} + T_\mathrm{syn} \right\} + T_\mathrm{neu} \\
  &=& \min \left\{ t_\mathrm{in1}^2+2.T_\mathrm{syn}+T_\mathrm{neu}, t_\mathrm{in1}^2 + 2.T_\mathrm{syn} + T_\mathrm{neu} \right\} + T_\mathrm{neu} \\
  &=& t_\mathrm{in1}^2+2.T_\mathrm{syn}+2.T_\mathrm{neu}.
\end{eqnarray}
We thus get an output pair of spikes spaced in time such that:
\begin{equation}
  \Delta T_{out} = t_\mathrm{out}^2 - t_\mathrm{out}^1 = t_\mathrm{in1}^2-t_\mathrm{in1}^1 = \Delta T_{in1}.
\end{equation}
The output is thus corresponding to the smallest of the 2 inputs of the network while the indicator
provides which of the two inputs is the smallest (\emph{smaller1}).
The same reasonning can be applied to the case where the second input (\emph{input2}) is the smallest.

\subsection{Maximum}
\label{sec:app:maximum}

\begin{figure}[h!]
  \centering
  \includegraphics[width=0.6\columnwidth]{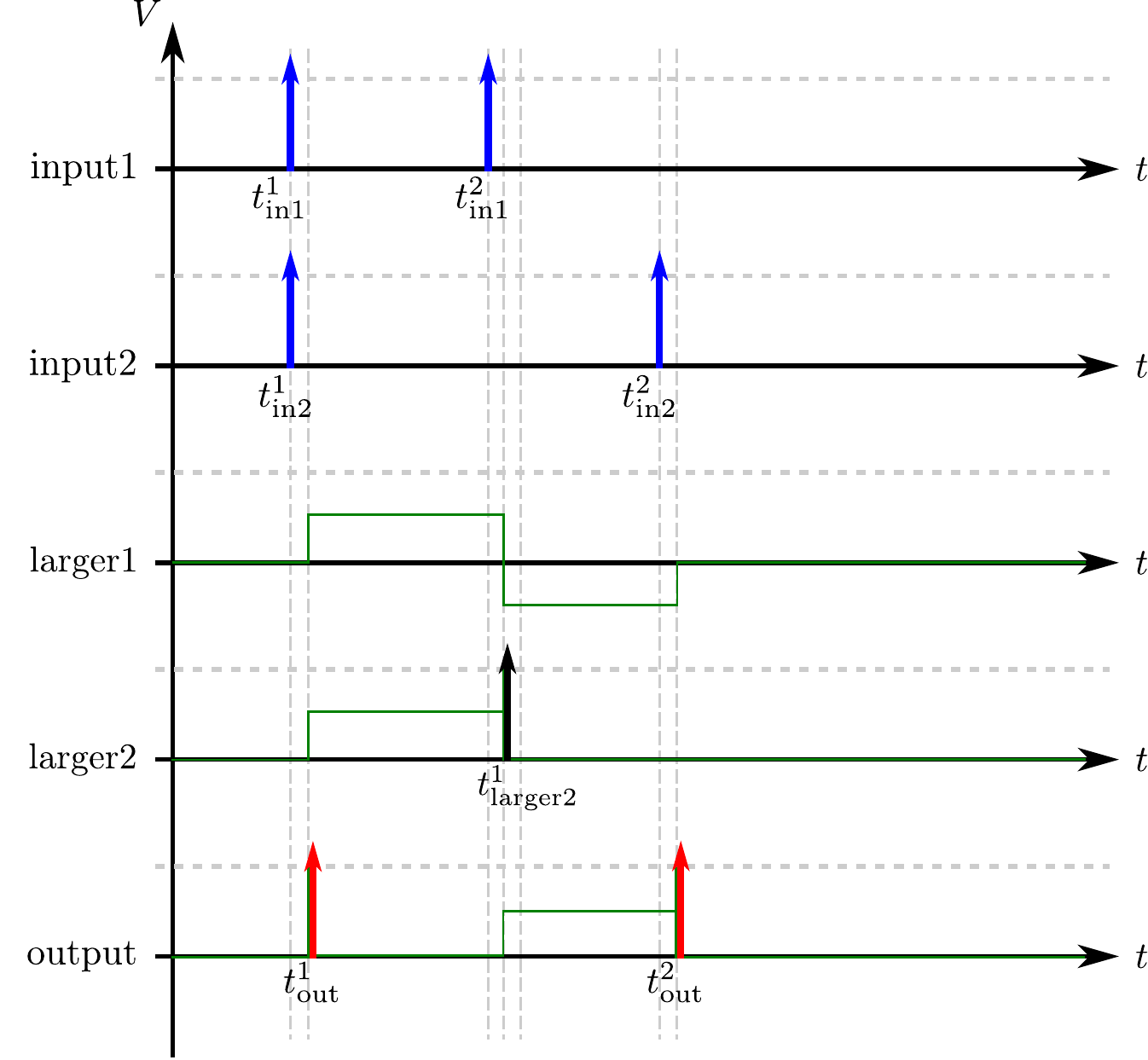}
  \caption{\textbf{Maximum:} chronogram of the network for inputs at times
           $t_\mathrm{in1}^1$, $t_\mathrm{in1}^2$ and $t_\mathrm{in2}^1$, $t_\mathrm{in2}^2.$
           (Input spikes are drawn in blue, output spikes in red. Green plots show the membrane
           potential of interesting neuron.)}
  \label{fig:app:chrono_maximum}
\end{figure}

The Maximum network (see Fig.~\ref{fig:net_maximum}) receives 2 different inputs (a pair of spikes) from each input neurons \emph{input1} ($t_\mathrm{in1}^1$ and $t_\mathrm{in1}^2$) and
\emph{input2} ($t_\mathrm{in2}^1$ and $t_\mathrm{in2}^2$) such that
$\Delta T_{in1}=t_\mathrm{in1}^2-t_\mathrm{in1}^1$ and $\Delta T_{in2}=t_\mathrm{in2}^2-t_\mathrm{in2}^1$ encode
its 2 inputs. Let's consider, without loss of generality, the case where the second input
(\emph{input2}) is larger than the first one as depicted Fig.~\ref{fig:app:chrono_maximum}.
The 2 inputs being synchronized as a prerequisite of the network, we have:
\begin{equation}
  t_\mathrm{in1}^1 = t_\mathrm{in2}^1,
\end{equation}
When \emph{input1} and \emph{input2} are triggered by the first encoding spike of each input,
\emph{larger1} and \emph{larger2}'s membrane potentials are set to $V_t/2$ and the \emph{output}
neuron spikes after a delay due to the connection from \emph{input1} and \emph{input2} such that:
\begin{equation}
  t_\mathrm{out1} = t_\mathrm{in2}^1 + T_\mathrm{syn} + T_\mathrm{neu}.
\end{equation}
When the smallest input (in this case \emph{input1}) receives its second encoding spike at time
$t_\mathrm{in1}^2,$ we know that the other input has to be the larger one. Hence, the \emph{larger2}
neuron reaches its threshold and emits a spike at time:
\begin{equation}
  t_\mathrm{larger2}^1 = t_\mathrm{in1}^2 + T_\mathrm{syn} + T_\mathrm{neu}.
\end{equation}
The spike from \emph{larger2} inhibits the \emph{larger1} neuron such that its membrane potential
goes down to $-V_t/2,$ while the connection \emph{input1} to \emph{output} raises \emph{output}'s
membrane potential to $V_t/2.$ 
When the second encoding spike from \emph{input2} is received,
the connection from \emph{input2} to \emph{larger1} moves back \emph{larger1}'s membrane potential
to its reset potential while its connection to \emph{output} triggers a spike at time:
\begin{equation}
  t_\mathrm{out}^2 = t_\mathrm{in2}^2 + T_\mathrm{syn} + T_\mathrm{neu}.
\end{equation}
We thus get an output pair of spikes spaced in time such that:
\begin{equation}
  \Delta T_{out} = t_\mathrm{out}^2 - t_\mathrm{out}^1 = t_\mathrm{in2}^2-t_\mathrm{in2}^1 = \Delta T_{in2}.
\end{equation}
The output is thus corresponding to the largest of the 2 inputs of the network.while the indicator
provides which of the two inputs is the largest (\emph{larger1}).
The same reasonning can be applied to the case where the first input (\emph{input1}) is the largest.

\section{Linear operations}

\subsection{Subtractor}
\label{sec:app:subtractor}

\begin{figure}[h!]
  \centering
  \includegraphics[width=0.5\columnwidth]{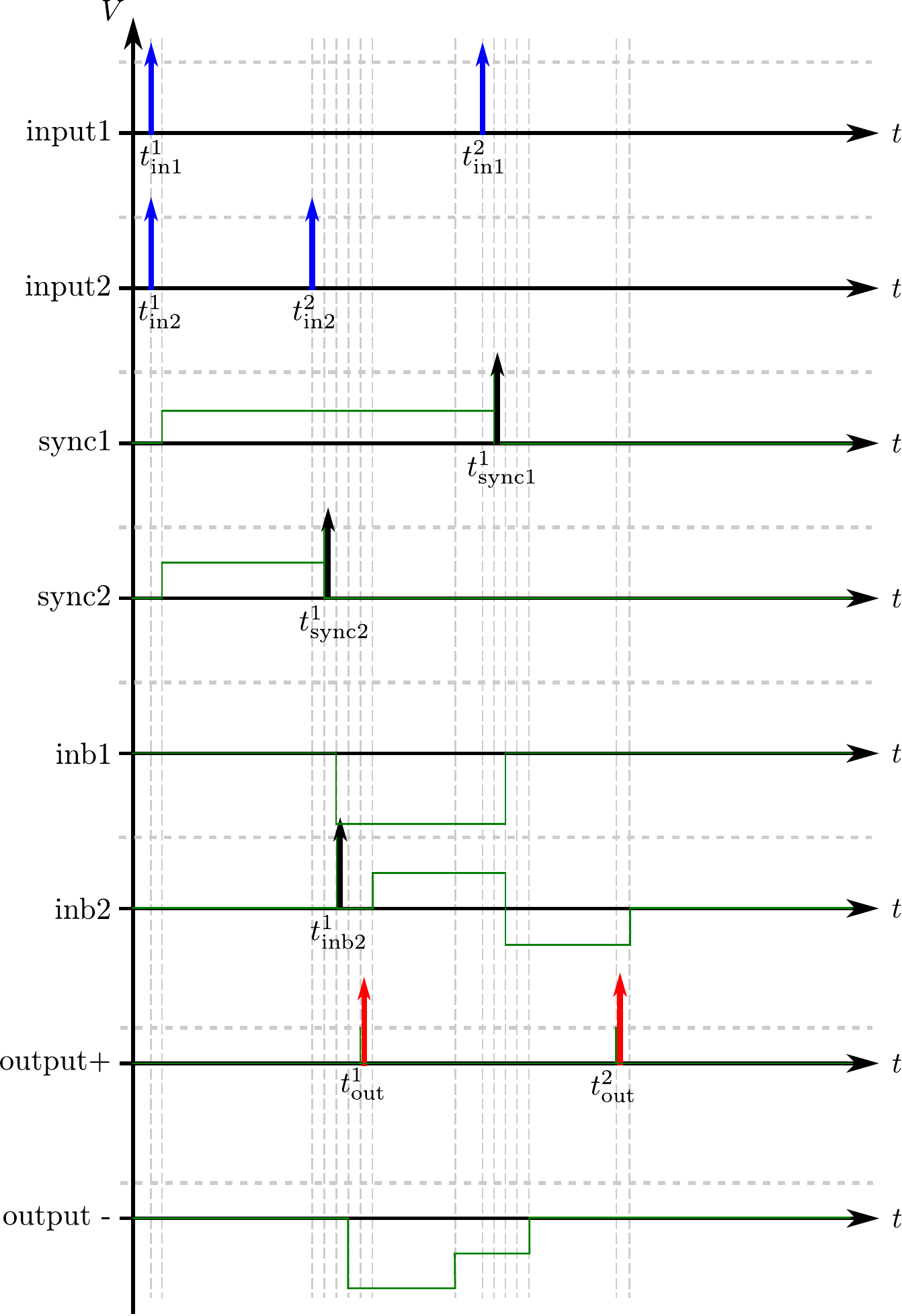}
  \caption{\textbf{Subtractor:} chronogram of the network for inputs at times
           $t_\mathrm{in1}^1$, $t_\mathrm{in1}^2$ and $t_\mathrm{in2}^1$, $t_\mathrm{in2}^2.$
           (Input spikes are drawn in blue, output spikes in red. Green plots show the membrane
           potential of interesting neuron.)}
  \label{fig:app:chrono_subtractor}
\end{figure}

The Subtractor network (see Fig.~\ref{fig:net_subtractor_simple}) receives 2 different inputs (a pair of spikes) from each input neurons \emph{input1} ($t_\mathrm{in1}^1$ and $t_\mathrm{in1}^2$) and
\emph{input2} ($t_\mathrm{in2}^1$ and $t_\mathrm{in2}^2$) such that
$\Delta T_{in1}=t_\mathrm{in1}^2-t_\mathrm{in1}^1$ and $\Delta T_{in2}=t_\mathrm{in2}^2-t_\mathrm{in2}^1$ encode
its 2 inputs. Let us consider, without loss of generality, the case where the first input
(\emph{input1}) is larger than the second one as depicted Fig.~\ref{fig:app:chrono_subtractor}.
Assuming the inputs to be synchronized, we have:
\begin{equation}
  t_\mathrm{in1}^1 = t_\mathrm{in2}^1.
\end{equation}
When \emph{input1} and \emph{input2} are triggered by the first encoding spikes of each input, they
activate the \emph{sync1} and \emph{sync2} neurons such that their membrane potentials are now set to
$V_t/2.$ When the second encoding spike of the smallest input (here, \emph{input2}) is received, the
activation from \emph{input2} to \emph{sync2} is sufficient to trigger a spike at time:
\begin{equation}
  t_\mathrm{sync2}^1 = t_\mathrm{in2}^2 + T_\mathrm{syn} + T_\mathrm{neu}.
\end{equation}
This spike inhibits the \emph{inb1} neuron after a time $T_\mathrm{syn},$ moving its membrane potential to
$-V_t$ and, because the sign of the output is now known, it triggers and output spikes on \emph{output~+}
at time:
\begin{equation}
  t_\mathrm{out}^1 = T_\mathrm{sync2}^1+3.T_\mathrm{syn}+2.T_\mathrm{neu}+T_\mathrm{neu}
                 = T_\mathrm{in2}^2 + 4.T_\mathrm{syn} + 4.T_\mathrm{neu}.
\end{equation}
It also contributes to \emph{output~-}'s membrane potential with an activation of $w_e$ at time
$t_\mathrm{sync2}^1+T_\mathrm{min}+3.T_\mathrm{syn}+2.T_\mathrm{neu}.$
But before this contribution reaches \emph{output~-}, the \emph{inb2} neuron is triggered and
produces a spike at time:
\begin{equation}
  t_\mathrm{inb2}^1 = t_\mathrm{sync} + T_\mathrm{syn} + T_\mathrm{neu}
\end{equation}
which inhibits \emph{output~-} with weight $2w_i$ at time
$t_\mathrm{inb2}^1+T_\mathrm{syn} = t_\mathrm{sync2}^1 + T_\mathrm{syn}+T_\mathrm{neu}.$ This inhibition
thus happens before the direct excitation from \emph{sync2}, which then leads to \emph{output~-} not
emitting a spike and its membrane potential to be set to $-V_t$ after receiving these 2 spikes.

When the second encoding spike of the largest input is received, the spike from \emph{input1} triggers
a spike from \emph{sync1} at time:
\begin{equation}
  t_\mathrm{sync1}^1 = t_\mathrm{in1}^2+T_\mathrm{syn}+T_\mathrm{neu}.
\end{equation}
This spike leads to the inhibition of \emph{inb2} to a membrane potential of $-V_t$ and an excitation of
\emph{inb1} back to its reset potential. It also activates \emph{output~+} back to its reset
potential and triggers an output spike at time:
\begin{equation}
  t_\mathrm{out}^2 = t_\mathrm{sync1}+T_\mathrm{min}+3.T_\mathrm{syn}+2.T_\mathrm{neu}+T_\mathrm{neu}
                 = t_\mathrm{in1}^2+T_\mathrm{min}+4.T_\mathrm{syn}+4.T_\mathrm{neu}
\end{equation}
We thus get a positive output as expected and an output value:
\begin{eqnarray}
  \Delta T_{out} &=& t_\mathrm{out}^2 - t_\mathrm{out}^1\\
  &=& T_\mathrm{min} + t_\mathrm{in1}^2 - t_\mathrm{in2}^2 \\
  &=& T_\mathrm{min} + (t_\mathrm{in1}^2 - t_\mathrm{in1}^1) - (t_\mathrm{in2}^2 - t_\mathrm{in2}^1) \\
  &=& T_\mathrm{min} + (\Delta T_{in1} - \Delta T_{in2}) \label{eq:subs_res}\\
  &=& T_\mathrm{min} + (\Delta T_{in1}-T_\mathrm{min}) - (\Delta T_{in2}-T_\mathrm{min}).
\end{eqnarray}
Knowing that for each value $x$, we encode it as a time interval
$f(x) = T_\mathrm{min} + x.T_\mathrm{cod},$ we have:
\begin{eqnarray}
  \Delta T_{out} &=& T_\mathrm{min} + x_{in1}.T_\mathrm{cod} - x_{in2}.T_\mathrm{cod} \\
                &=& T_\mathrm{min} + (x_{in1} - x_{in2}).T_\mathrm{cod}
\end{eqnarray}
which corresponds to the encoding of the result $input1 - input2.$

%% With this simple architecture, when the 2 inputs are equal, the 2 branches of the network (i.e.
%% the \emph{input1} branch and the \emph{input2} branches) are activated at the same time when the
%% second encoding spikes of both inputs arrive. This lead to an indetermination of the ouput sign
%% because lateral inhibition between the branches have no time to propagate. To solve this problem,
%% we introduced the full Subtractor architecture in Fig.~\ref{fig:net_subtractor}. A \emph{zero} neuron
%% is added to detect the particular case of equal inputs. This is implemented by the synaptic connections
%% from \emph{sync1} and \emph{sync2} to \emph{zero} which allows the \emph{zero} neuron to act as a
%% coincidence detector between spikes from \emph{sync1} and \emph{sync2}. Because these 2 neurons spike
%% only at the time of the second encoding spike of their associated input, detecting a time coincidence
%% between spikes produced by \emph{sync1} and \emph{sync2} is equivalent to detecting the equality of
%% the 2 inputs (if they are correctly synchronized). In this case, the \emph{zero} neuron uses a fast
%% inhibitory pathway to inhibit neurons leading to a negative output (\emph{inb2} and \emph{output~-}).
%% This allows the network to output a correct result on its positive output (\emph{output~+}).

\subsection{Linear Combination}
\label{sec:app:linear_combination}

The first part of the Linear Combination presented Fig.~\ref{fig:net_linear_combination} can be
decomposed in a series of simpler circuits to what has been presented before for the
inverting memory. For each of the $N$ inputs, one branch is managing positive inputs while the other
one is managing negative inputs (only one of these 2 can be activated in any computation). The
architecture of each of these branches can be seen as an inverting memory
(see Fig.~\ref{fig:net_inverting_memory}) storing a value either in \emph{acc1+} or \emph{acc1-}.
The targeted accumulator is chosen depending on the sign of the input and the sign of its associated
coefficient $\alpha_i$ to represent the sign of the input's contribution to the result.
\emph{acc1+} is accumulating all positive contributions (positive inputs with positive coefficients and
negative inputs with negative inputs) and \emph{acc1-} is accumulating all negative contributions
(negative inputs with positive coefficients and positive inputs with negative coefficients).
With the same reasoning leading to Eq.~(\ref{eq:imem_vsto}) we can get the contribution of an input $i$
to its associated accumulator, if $\Delta T_{in}^i$ is the input interspike associated to the considered
input:
\begin{equation}
  V_{sto}^i = |\alpha_i|.\frac{w_\mathrm{acc}}{\tau_m}.(\Delta T_{in}^i - T_\mathrm{min}),
\end{equation}
The integration in the accumulators being linear, we obtain the membrane potential stored in the 2
accumulators after all inputs have been fed into the network:
\begin{eqnarray}
  V_{sto}^{acc1+} &=& \sum_{i\in\mathcal{I}^+} V_{sto}^i \\
               &=& \sum_{i\in\mathcal{I}^+} |\alpha_i|.\frac{w_\mathrm{acc}}{\tau_m}.(\Delta T_{in}^i - T_\mathrm{min}) \\
  V_{sto}^{acc1-} &=& \sum_{i\in\mathcal{I}^-} V_{sto}^i \\
               &=& \sum_{i\in\mathcal{I}^-} |\alpha_i|.\frac{w_\mathrm{acc}}{\tau_m}.(\Delta T_{in}^i - T_\mathrm{min})
\end{eqnarray}
where $\mathcal{I}^+$ is the set of inputs contributing positively to the output and
$\mathcal{I}^-$ is the set of inputs contributing negatively to the output.
When the $N$ inputs have been fed into the network, the \emph{sync} neuron finally receives enough
excitation to produce a spike at time $t_\mathrm{sync}^1.$ This spike triggers the readout process of
\emph{acc1+} and \emph{acc1-} and, at the same time, starts integrating in neurons \emph{acc2+}
and \emph{acc2-}. This process is similar to the one used in the Memory network (see
Appendix~\ref{sec:app:memory}). If we consider the positive accumulator, we obtain spikes from
\emph{acc1+} and \emph{acc2+} at time $t_\mathrm{acc1+}^1$ and $t_\mathrm{acc2+}^1$ respectively with the
following conditions, where $t_\mathrm{st}^1=t_\mathrm{sync}^1+T_\mathrm{syn}$ is the time at
which the integration begins:
\begin{equation}
  \left\{
  \begin{matrix}
    V_t &=& \frac{w_\mathrm{acc}}{\tau_m}.(t_\mathrm{acc1+}-t_\mathrm{st}^1) + V_{sto}^{acc1+} \\
    V_t &=& \frac{w_\mathrm{acc}}{\tau_m}.(t_\mathrm{acc2+}-t_\mathrm{st}^1)
  \end{matrix}
  \right.
\end{equation}
By definition of $w_\mathrm{acc}$, we thus get:
\begin{equation}
  t_\mathrm{acc2+} = t_\mathrm{st}^1+T_\mathrm{max}
\end{equation}
and
\begin{eqnarray}
  T_\mathrm{max} &=& t_\mathrm{acc1+} - t_\mathrm{st}^1 + \sum_{i\in\mathcal{I}^+} |\alpha_i|.(\Delta T_{in}^i - T_\mathrm{min}) \\
  t_\mathrm{acc1+} &=& t_\mathrm{st}^1 + T_\mathrm{max} - \sum_{i\in\mathcal{I}^+} |\alpha_i|.(\Delta T_{in}^i - T_\mathrm{min})
\end{eqnarray}
Neuron \emph{inter+} is thus producing 2 spikes with an interspike $\Delta T_{inter}^+$ such that:
\begin{eqnarray}
  \Delta T_{inter}^+ &=& t_\mathrm{acc2+} + T_\mathrm{min} + T_\mathrm{syn} + T_\mathrm{neu} - (t_\mathrm{acc1+} + T_\mathrm{syn} + T_\mathrm{neu}) \\
  &=& \sum_{i\in\mathcal{I}^+} |\alpha_i|.(\Delta T_{in}^i - T_\mathrm{min}) + T_\mathrm{min}
\end{eqnarray}
The same reasoning on \emph{acc1-} and \emph{acc2-} leads to a pair of spikes on neuron \emph{inter-}
with an interspike $\Delta T_{inter}^-$:
\begin{equation}
  \Delta T_{inter}^- = \sum_{i\in\mathcal{I}^-} |\alpha_i|.(\Delta T_{in}^i - T_\mathrm{min}) + T_\mathrm{min}
\end{equation}
This two values are then synchronized by a Synchronizer network described previously and subtracted from
one another such that the output is, according to Eq.~(\ref{eq:subs_res}):
\begin{eqnarray}
  \Delta T_{out} &=& \Delta T_{inter}^+ - \Delta T_{inter}^- + T_\mathrm{min} \\
  &=& \sum_{i\in\mathcal{I}^+} |\alpha_i|.(\Delta T_{in}^i - T_\mathrm{min}) - \sum_{i\in\mathcal{I}^-} |\alpha_i|.(\Delta T_{in}^i - T_\mathrm{min}) + T_\mathrm{min} \\
  &=& \sum_{i=0}^{N-1} \epsilon_i.\alpha_i.(\Delta T_{in}^i - T_\mathrm{min}) + T_\mathrm{min}
\end{eqnarray}
where $\epsilon_i$ is $+1$ if input $i$ is positive and $-1$ otherwise. This is the expected
result of the linear combination.

\section{Non-linear operations}

\subsection{Natural Logarithm}
\label{sec:app:natural_log}

\begin{figure}[h!]
  \centering
  \includegraphics[width=0.6\columnwidth]{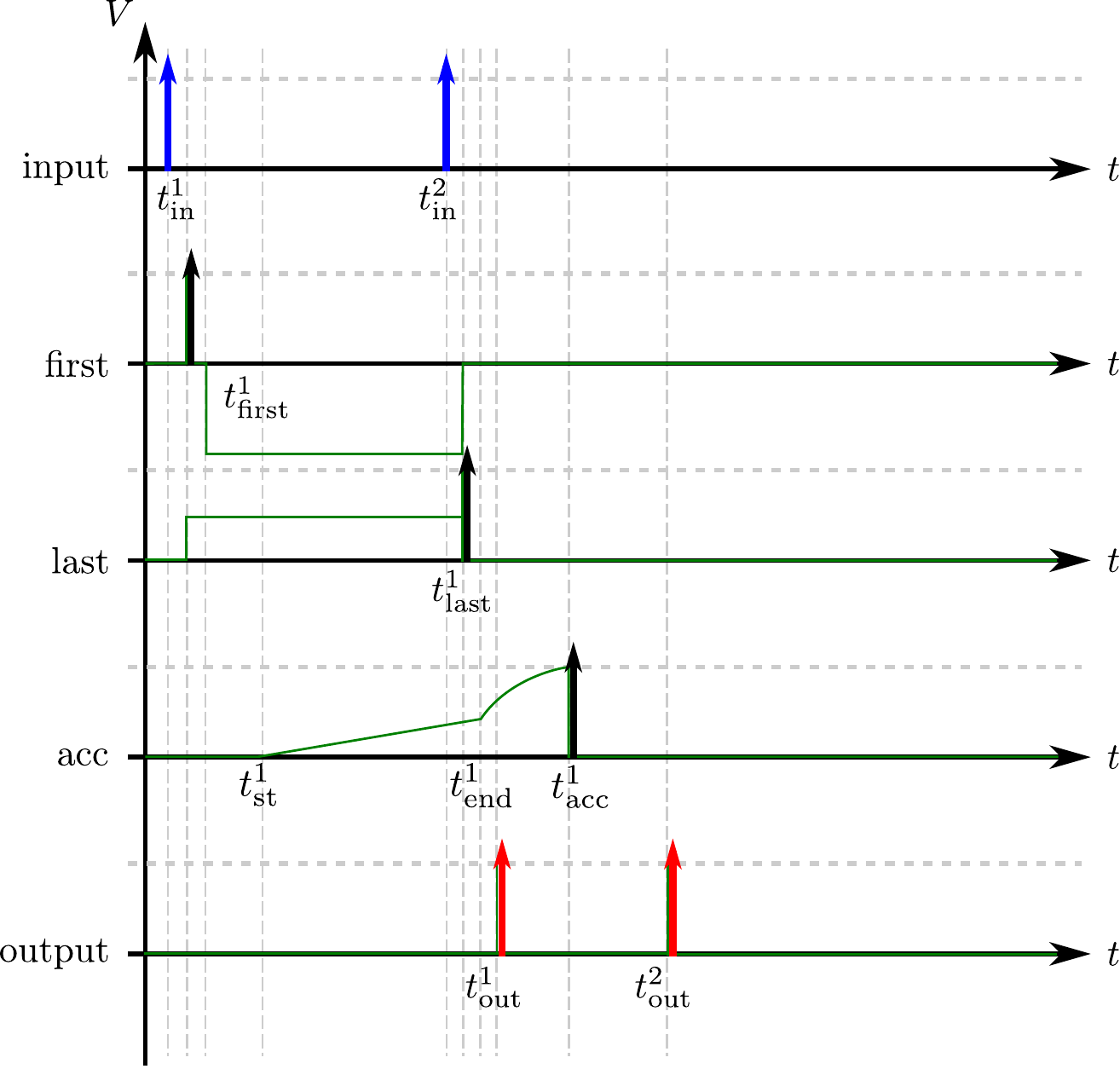}
  \caption{\textbf{Log:} chronogram of the network for an input at times
           $t_\mathrm{in}^1$ and $t_\mathrm{in}^2.$
           (Input spikes are drawn in blue, output spikes in red. Green plots show the membrane
           potential of interesting neurons).}
  \label{fig:app:chrono_log}
\end{figure}

The Log network (see Fig.~\ref{fig:net_log}) receives 2 spikes on the
\emph{input} neuron at times $t_\mathrm{in}^1$ and $t_\mathrm{in}^2$ such that
$\Delta T_{in}=t_\mathrm{in}^2-t_\mathrm{in}^1$ encodes its input value. The same reasoning leads
us to obtain the spike times of the \emph{first} and \emph{last} neurons at:
\begin{eqnarray}
  t_\mathrm{first}^1 &=& t_\mathrm{in}^1 + T_\mathrm{syn} + T_\mathrm{neu} \\
  t_\mathrm{last}^1 &=& t_\mathrm{in}^2 + T_\mathrm{syn} + T_\mathrm{neu}
\end{eqnarray}
The \emph{acc} neuron is thus integrating from time:
\begin{eqnarray}
  t_\mathrm{st}^1 &=& t_\mathrm{first}^1 + T_\mathrm{syn} + T_\mathrm{min} \\
                &=& t_\mathrm{in}^1 + 2.T_\mathrm{syn} + T_\mathrm{min} + T_\mathrm{neu}
\end{eqnarray}
to time:
\begin{eqnarray}
  t_\mathrm{end}^1 &=& t_\mathrm{last}^1 + T_\mathrm{syn} \\
                 &=& t_\mathrm{in}^2 + 2.T_\mathrm{syn} + T_\mathrm{neu}
\end{eqnarray}
The membrane potential of neuron \emph{acc} at the end of this integration phase is thus:
\begin{eqnarray}
  V_{sto} &=& \frac{\bar{w}_\mathrm{acc}}{\tau_m}(t_\mathrm{end}^1-t_\mathrm{st}^1) \\
  &=& \frac{\bar{w}_\mathrm{acc}}{\tau_m}(t_\mathrm{in}^2-t_\mathrm{in}^1-T_\mathrm{min}) \\
  &=& \frac{\bar{w}_\mathrm{acc}}{\tau_m}(\Delta T_{in} - T_\mathrm{min})
\end{eqnarray}
If we name $\Delta T_{cod} = \Delta T_{in} - T_\mathrm{min}$ and considering the definition of
$\bar{w}_\mathrm{acc}$, we have:
\begin{equation}
  V_{sto} = V_t.\frac{\Delta T_{cod}}{T_\mathrm{cod}}.
\end{equation}
The other synaptic connections from \emph{last} to \emph{acc} also activate the $g_f$ dynamics
of the \emph{acc} neuron at time $t_\mathrm{end}^1$. When the $g_f-\mathrm{synapse}$ gets activated,
\emph{acc}'s membrane potential follows the following evolution obtained by solving the differential
system Eq.~(\ref{eq:neural_model}):
\begin{equation}
  V = V_{sto} + g_\mathrm{mult}\frac{\tau_f}{\tau_m}(1-e^{-(t-t_\mathrm{end}^1)/\tau_f})
\end{equation}
for $t \geq t_\mathrm{end}^1.$
According to Eq.~(\ref{eq:g_mult}), we chose $g_\mathrm{mult} = V_t.\frac{\tau_m}{\tau_f}$ so that:
\begin{equation}
  V = V_t.\frac{\Delta T_{cod}}{T_\mathrm{cod}} + V_t.(1-e^{-(t-t_\mathrm{end}^1)/\tau_f})
\end{equation}
The \emph{acc} neuron will then spike at time $t_\mathrm{acc}^1$ when the condition $V=V_t$ is met.
This gives us:
\begin{eqnarray}
  V_t &=& V_t.\frac{\Delta T_{cod}}{T_\mathrm{cod}} + V_t.(1-e^{-(t_\mathrm{acc}^1-t_\mathrm{end}^1)/\tau_f}) \\
  \frac{\Delta T_{cod}}{T_\mathrm{cod}} &=& e^{-(t_\mathrm{acc}^1-t_\mathrm{end}^1)/\tau_f} \\
  t_\mathrm{acc}^1 &=& -\tau_f.\log \left(\frac{\Delta T_{cod}}{T_\mathrm{cod}}\right) + t_\mathrm{end}^1
\end{eqnarray}
The first output spike is generated by the connection from \emph{last} to \emph{output}, thus:
\begin{equation}
  t_\mathrm{out}^1 = t_\mathrm{last}^1 + T_\mathrm{neu} + 2.T_\mathrm{syn}
\end{equation}
While the second output spike is generated by the connection from \emph{acc} to \emph{output}, thus:
\begin{eqnarray}
  t_\mathrm{out}^2 &=& t_\mathrm{acc}^1 + T_\mathrm{neu} + T_\mathrm{syn} + T_\mathrm{min} \\
  &=& -\tau_f.\log \left(\frac{\Delta T_{cod}}{T_\mathrm{cod}}\right)+ t_\mathrm{end}^1 + T_\mathrm{neu} + T_\mathrm{syn} + T_\mathrm{min}\\
  &=& -\tau_f.\log \left(\frac{\Delta T_{cod}}{T_\mathrm{cod}}\right) + t_\mathrm{last}^1 + T_\mathrm{neu} + 2.T_\mathrm{syn} + T_\mathrm{min}
\end{eqnarray}
This gives us the output $\Delta T_{out}$:
\begin{eqnarray}
  \Delta T_{out} &=& t_\mathrm{out}^2 - t_\mathrm{out}^1 \\
  &=& -\tau_f.\log \left(\frac{\Delta T_{cod}}{T_\mathrm{cod}}\right) + T_\mathrm{min} \\
  &=& T_\mathrm{min} + \tau_f.\log \left(\frac{T_\mathrm{cod}}{\Delta T_{cod}}\right)
\end{eqnarray}

\subsection{Exponential}
\label{sec:app:exponential}

\begin{figure}[h!]
  \centering
  \includegraphics[width=0.6\columnwidth]{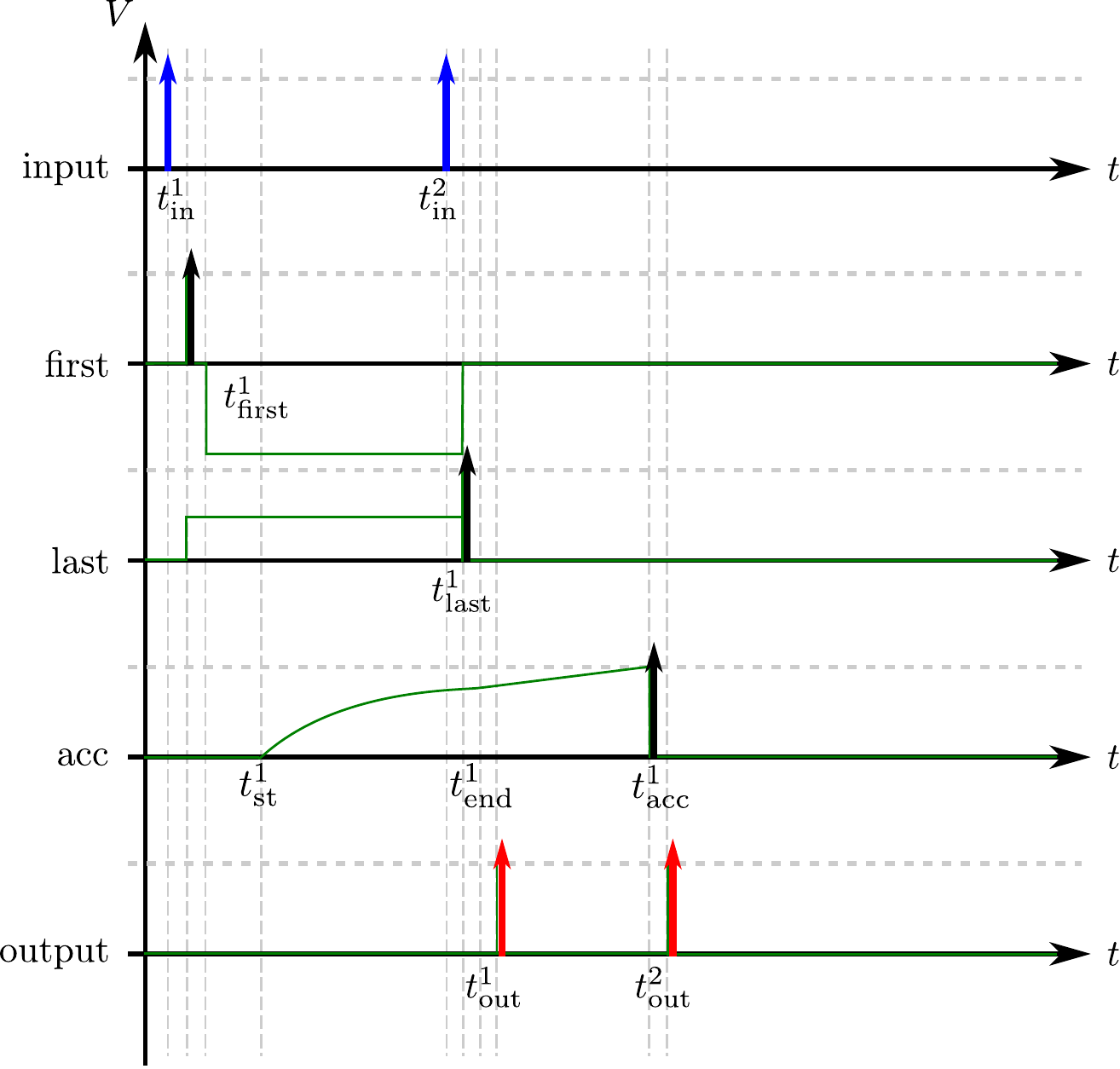}
  \caption{\textbf{Exp:} chronogram of the network for an input at times
           $t_\mathrm{in}^1$ and $t_\mathrm{in}^2.$
           (Input spikes are drawn in blue, output spikes in red. Green plots show the membrane
           potential of interesting neurons).}
  \label{fig:app:chrono_exp}
\end{figure}

The Exp network (see Fig.~\ref{fig:net_exp}) receives 2 spikes on the
\emph{input} neuron at times $t_\mathrm{in}^1$ and $t_\mathrm{in}^2$ such that
$\Delta T_{in}=t_\mathrm{in}^2-t_\mathrm{in}^1$ encodes its input value. The same reasoning leads
us to obtain the spike times of the \emph{first} and \emph{last} neurons at:
\begin{eqnarray}
  t_\mathrm{first}^1 &=& t_\mathrm{in}^1 + T_\mathrm{syn} + T_\mathrm{neu} \\
  t_\mathrm{last}^1 &=& t_\mathrm{in}^2 + T_\mathrm{syn} + T_\mathrm{neu}
\end{eqnarray}
The \emph{first} neuron then triggers the $g_f$ dynamics of neuron \emph{acc} at time:
\begin{equation}
  t_\mathrm{st}^1 = t_\mathrm{first}^1+T_\mathrm{syn}+T_\mathrm{min}
  = t_\mathrm{in}^1+T_\mathrm{neu}+2.T_\mathrm{syn}+T_\mathrm{min}
\end{equation}
Solving the differential system Eq.~(\ref{eq:neural_model}), \emph{acc}'s membrane potential is
following the evolution:
\begin{eqnarray}
  V &=& g_\mathrm{mult}\frac{\tau_f}{\tau_m}(1-e^{-(t-t_\mathrm{st}^1)/\tau_f}) \\
  &=& V_t.(1-e^{-(t-t_\mathrm{st}^1)/\tau_f})
\end{eqnarray}
This evolution is stopped at $t_\mathrm{end}^1$ by the connection from \emph{last} to \emph{acc} through its
action on the $gate$ signal:
\begin{equation}
  t_\mathrm{end}^1 = t_\mathrm{last}^1 + T_\mathrm{syn} = t_\mathrm{in}^2 + T_\mathrm{neu} + 2.T_\mathrm{syn}
\end{equation}
At the end of this phase, \emph{acc}'s membrane potential is thus equal to:
\begin{eqnarray}
  V_{sto} &=& V_t.(1-e^{-(t_\mathrm{end}^1-t_\mathrm{st}^1)/\tau_f}) \\
  &=& V_t.(1-e^{-(t_\mathrm{in}^2-t_\mathrm{in}^1-T_\mathrm{min})/\tau_f}) \\
  &=& V_t.(1-e^{-(\Delta T_{in}-T_\mathrm{min})/\tau_f}) \\
  &=& V_t.(1-e^{-\Delta T_{cod}/\tau_f})
\end{eqnarray}
with $\Delta T_{cod} = \Delta T_{in} - T_\mathrm{min}.$ At the same time, \emph{last} is starting a second
integration process of a $g_e-\mathrm{synapse}$. \emph{acc}'s membrane potential is then following the
evolution:
\begin{equation}
  V = V_{sto} + \frac{\bar{w}_\mathrm{acc}}{\tau_m}.(t-t_\mathrm{end}^1)
\end{equation}
This behavior leads to a spike at time $t_\mathrm{acc}^1$ when the condition $V=V_t$ is met:
\begin{eqnarray}
  V_t &=& V_{sto} + \frac{\bar{w}_\mathrm{acc}}{\tau_m}.(t_\mathrm{acc}^1-t_\mathrm{end}^1) \\
  V_t &=& V_t.(1-e^{-\Delta T_{cod}/\tau_f}) + \frac{V_t}{T_\mathrm{cod}}.(t_\mathrm{acc}^1-t_\mathrm{end}^1) \\
  t_\mathrm{acc}^1 &=& t_\mathrm{end}^1 + T_\mathrm{cod}.e^{-\Delta T_{cod}/\tau_f}.
\end{eqnarray}
The first output spike is produced by the connection from \emph{last} to \emph{output}, thus:
\begin{equation}
  t_\mathrm{out}^1 = t_\mathrm{last}^1 + T_\mathrm{neu} + 2.T_\mathrm{syn}
\end{equation}
The second output spike is produced by the connection from \emph{acc1} to \emph{output}, thus:
\begin{eqnarray}
  t_\mathrm{out}^2 &=& t_\mathrm{acc}^1 + T_\mathrm{neu} + T_\mathrm{syn} + T_\mathrm{min} \\
  &=& T_\mathrm{cod}.e^{-\Delta T_{cod}/\tau_f} + t_\mathrm{end}^1 + T_\mathrm{neu} + T_\mathrm{syn} + T_\mathrm{min} \\
  &=& T_\mathrm{cod}.e^{-\Delta T_{cod}/\tau_f} + t_\mathrm{last}^1 + T_\mathrm{neu} + 2.T_\mathrm{syn} + T_\mathrm{min}
\end{eqnarray}
This gives us the output $\Delta T_{out}$:
\begin{eqnarray}
  \Delta T_{out} &=& t_\mathrm{out}^2 - t_\mathrm{out}^1 \\
  &=& T_\mathrm{cod}.e^{-\Delta T_{cod}/\tau_f} + T_\mathrm{min}
\end{eqnarray}

\subsection{Multiplier}
\label{sec:app:multiplier}

\begin{figure}[h!]
  \centering
  \includegraphics[width=0.6\columnwidth]{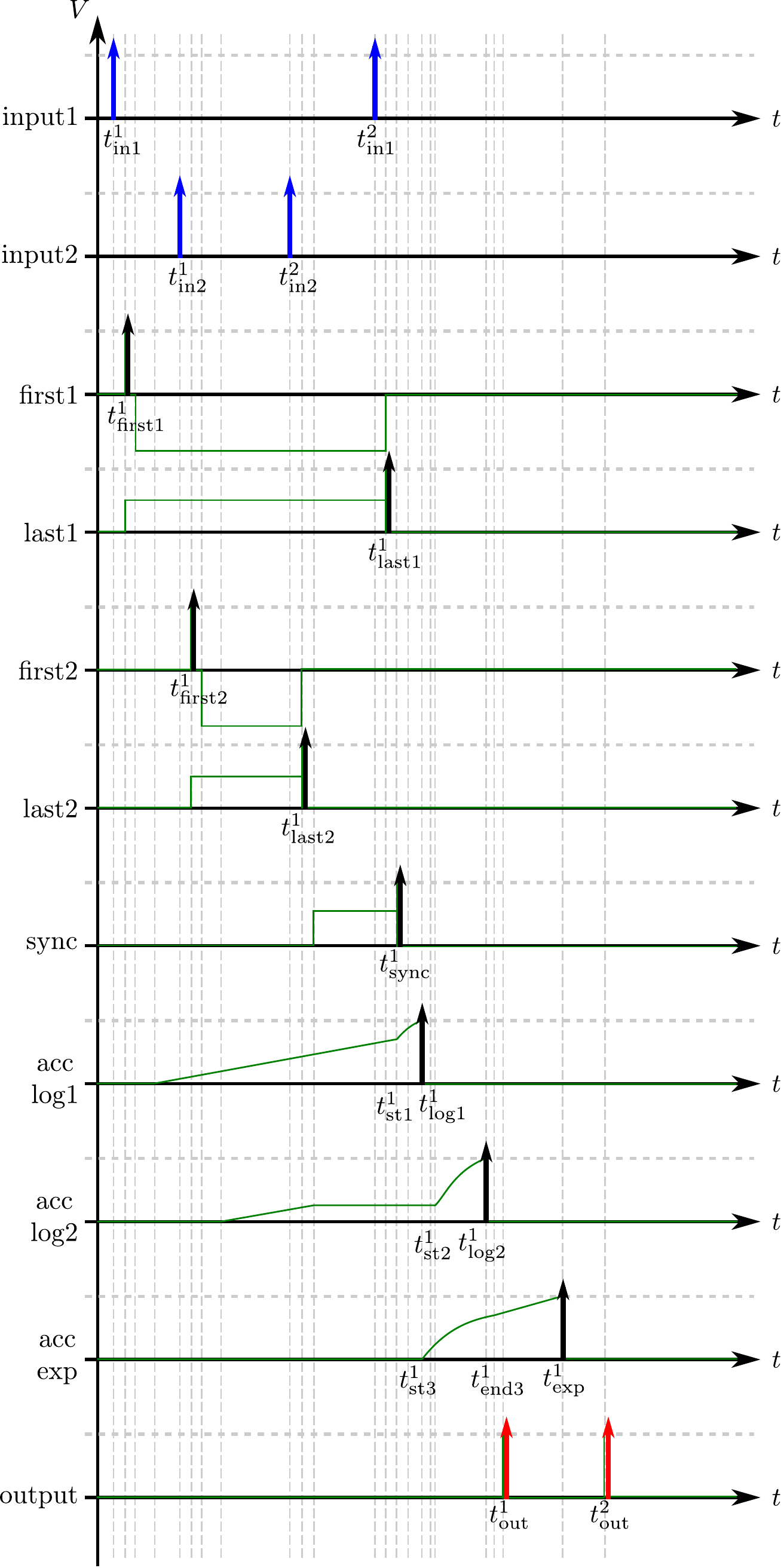}
  \caption{\textbf{Multiplier:} chronogram of the network for inputs at times
           $t_\mathrm{in1}^1,$ $t_\mathrm{in1}^2$ and $t_\mathrm{in2}^1,$ $t_\mathrm{in2}^2.$
           (Input spikes are drawn in blue, output spikes in red. Green plots show the membrane
           potential of interesting neurons).}
  \label{fig:app:chrono_multiplier}
\end{figure}

The Multiplier network (see Fig.~\ref{fig:net_multiplier}) receives 2 different inputs (a pair of spikes) from each input neurons \emph{input1} ($t_\mathrm{in1}^1$ and $t_\mathrm{in1}^2$) and
\emph{input2} ($t_\mathrm{in2}^1$ and $t_\mathrm{in2}^2$) such that
$\Delta T_{in1}=t_\mathrm{in1}^2-t_\mathrm{in1}^1$ and $\Delta T_{in2}=t_\mathrm{in2}^2-t_\mathrm{in2}^1$ encode
its 2 inputs.
The first layers of the network, composed of the neurons \emph{input}, \emph{first}, \emph{last} and
\emph{acc\_log} for each inputs are similar to the Logarithm network. Considering the results from
Appendix~\ref{sec:app:natural_log}, when the 2 inputs have been fed into the network, we get 2 potentials
stored in \emph{acc\_log1} and \emph{acc\_log2} (respectively $V_{sto1}$ and $V_{sto2}$):
\begin{equation}
  \left\{\begin{matrix}
  V_{sto1} = V_t.\frac{\Delta T_{cod1}}{T_\mathrm{cod}} \\
  V_{sto2} = V_t.\frac{\Delta T_{cod2}}{T_\mathrm{cod}} \\
  \end{matrix}\right.
\end{equation}
When the 2 inputs have been fed into the network, spikes from \emph{last1} and \emph{last2} activate
the \emph{sync} neuron which spikes at time $t_\mathrm{sync}^1.$ This triggers the readout of the log
of input1 in the \emph{acc\_log1} neuron at time:
\begin{equation}
  t_\mathrm{st1}^1 = t_\mathrm{sync1}^1 + T_\mathrm{syn}.
\end{equation}
Results from Appendix~\ref{sec:app:natural_log} tell us that this neuron will thus spike at time:
\begin{equation}
  t_\mathrm{log1}^1 = -\tau_f.\log \left(\frac{\Delta T_{cod1}}{T_\mathrm{cod}}\right) + t_\mathrm{st}^1.
\end{equation}
A spike from the \emph{acc\_log1} neuron will then trigger the readout of the log value of input2
in the \emph{acc\_log2} neuron at time:
\begin{equation}
  t_\mathrm{st2}^1 = t_\mathrm{log1}^1 + T_\mathrm{syn}.
\end{equation}
\emph{acc\_log2} will then produce a spike at time:
\begin{equation}
  t_\mathrm{log2}^1 = -\tau_f.\log \left(\frac{\Delta T_{cod2}}{T_\mathrm{cod}}\right) + t_\mathrm{st}^2.
\end{equation}
Which will trigger the first output spike at time:
\begin{equation}
  t_\mathrm{out}^1 = t_\mathrm{log2}^1 + 2.T_\mathrm{syn}.
\end{equation}
At the same time, the \emph{sync} neuron also started the $g_f$ dynamics of neuron \emph{acc\_exp} at
time:
\begin{equation}
  t_\mathrm{st3}^1 = t_\mathrm{sync}^1 + 3.T_\mathrm{syn}.
\end{equation}
This process is stopped by the spike from neuron \emph{acc\_log2} at time:
\begin{equation}
  t_\mathrm{end3}^1 = t_\mathrm{log2}^1 + T_\mathrm{syn}
\end{equation}
Results from Appendix~\ref{sec:app:exponential} tell us that the potential stored in \emph{acc\_exp} at
that time is thus:
\begin{equation}
  V_{sto3} = V_t.(1-e^{-(t_\mathrm{end3}^1-t_\mathrm{st3}^1)/\tau_f})
\end{equation}
and that the integration process started by the second connection from \emph{acc\_log2} to
\emph{acc\_exp} (acting on $g_e$) will result in a spike at time:
\begin{equation}
  t_\mathrm{exp}^1 =  t_\mathrm{end3}^1 + T_\mathrm{cod}.e^{-(t_\mathrm{end3}^1-t_\mathrm{st3}^1)/\tau_f}.
\end{equation}
This will result in the second output spike at time:
\begin{equation}
  t_\mathrm{out}^2 = t_\mathrm{exp}^1 + T_\mathrm{syn} + T_\mathrm{min}.
\end{equation}
The output of the network is thus the interspike $\Delta T_{out}$ such that:
\begin{eqnarray}
  \Delta T_{out} &=& t_\mathrm{out}^2 - t_\mathrm{out}^1 \\
  &=& T_\mathrm{min} + t_\mathrm{exp}^1 - t_\mathrm{log2}^1 - T_\mathrm{syn} \\
  &=& T_\mathrm{min} + T_\mathrm{cod}.e^{-(t_\mathrm{end3}^1-t_\mathrm{st3}^1)/\tau_f} + t_\mathrm{end3}^1 - T_\mathrm{syn} \\
  &=& T_\mathrm{min} + T_\mathrm{cod}.e^{-(t_\mathrm{end3}^1-t_\mathrm{st3}^1)/\tau_f}
\end{eqnarray}
From previous equations, we get:
\begin{eqnarray}
  t_\mathrm{end3}^1-t_\mathrm{st3}^1 &=& t_\mathrm{log2} - t_\mathrm{sync}^1 - 2.T_\mathrm{syn} \\
  &=& -\tau_f.\log \left(\frac{\Delta T_{cod2}}{T_\mathrm{cod}}\right) + t_\mathrm{st}^2 - t_\mathrm{sync}^1 - 2.T_\mathrm{syn} \\
  &=& -\tau_f.\log \left(\frac{\Delta T_{cod2}}{T_\mathrm{cod}}\right) + t_\mathrm{log1}^1 - t_\mathrm{sync}^1 - T_\mathrm{syn} \\
  &=& -\tau_f.\log \left(\frac{\Delta T_{cod2}}{T_\mathrm{cod}}\right) - \tau_f.\log \left(\frac{\Delta T_{cod1}}{T_\mathrm{cod}}\right) \\
  &=& -\tau_f.\log \left(\frac{\Delta T_{cod1}.\Delta T_{cod2}}{T_\mathrm{cod}}\right).
\end{eqnarray}
Which then gives us:
\begin{equation}
  \Delta T_{out} = T_\mathrm{min} + \Delta T_{cod1}.\Delta T_{cod2}
\end{equation}
which corresponds to the encoded value of the produce of the value encoded by the 2 inputs.